\documentclass[iicol]{sn-jnl}

\usepackage[switch]{lineno}
\usepackage{epsfig}
\usepackage{natbib}
\usepackage{graphicx}
\usepackage{amsmath}
\usepackage{amssymb}
\usepackage{algorithm}
\usepackage{algpseudocode}
\usepackage{float}
\usepackage{color}
\usepackage{colortbl}
\usepackage{breqn}
\usepackage{rotating}
\usepackage{makecell}
\usepackage{tabularx}
\usepackage{xcolor}
\usepackage{hyperref}
\usepackage{multirow}
\usepackage{listings}%
\usepackage{hhline}
\usepackage{amsfonts}
\usepackage{empheq}
\usepackage{framed, color}
\usepackage{graphics}
\usepackage{graphicx}
\usepackage{filecontents}
\usepackage{subfigure}
\usepackage{verbatim}
\usepackage{lineno}
\usepackage{setspace}
\usepackage{textcomp,booktabs}
\usepackage{caption}
\usepackage{stmaryrd}
\usepackage{manyfoot}%
\usepackage{amsthm}%
\usepackage{mathrsfs}%
\usepackage[mathscr]{eucal}
\usepackage{adjustbox}
\usepackage{marvosym}  
\usepackage{enumitem}
\usepackage{booktabs}
\usepackage{subcaption}

\def\0{{\bf 0}}
\def\1{{\bf 1}}

\definecolor{hyperlinkTextType}{rgb}{0.8, 0.0, 0.0}

\definecolor{purple}{rgb}{0.56,0.27,0.68}
\definecolor{red}{rgb}{0.95,0.4,0.4}
\definecolor{purered}{rgb}{1,0,0}
\definecolor{blue}{rgb}{0.4,0.4,0.95}
\definecolor{darkblue}{rgb}{0,0,0.8}
\definecolor{grey}{rgb}{0.6,0.6,0.6}
\definecolor{col1}{RGB}{232, 161, 148}
\definecolor{col11}{RGB}{255, 228, 228}
\definecolor{col2}{RGB}{148, 187, 232}
\definecolor{col33}{RGB}{206, 239, 255}
\definecolor{yellow}{RGB}{255, 255, 224}
\definecolor{col3}{RGB}{233, 255, 245}
\definecolor{lightgreen}{RGB}{233, 255, 245}
\definecolor{lightgrey}{rgb}{0.85,0.85,0.85}
\definecolor{lightlightgrey}{rgb}{0.9,0.9,0.9}
\definecolor{verylightBG}{rgb}{0.9,0.99,0.99}
\definecolor{darkgreen}{rgb}{0., 0.85, 0.5}

\newcommand\darkgreen[1]{\textcolor{darkgreen}{#1}}

\graphicspath{{./figures/}}

\begin{document}

\title[Article Title]{Long-Tailed 3D Detection via Multi-Modal Late-Fusion}

\author[1]{\fnm{Yechi} \sur{Ma}}
\equalcont{These authors contributed equally to this work.}

\author[2]{\fnm{Neehar} \sur{Peri}}
\equalcont{These authors contributed equally to this work. \href{https://github.com/cc50121/lt3d-lf}{Project page} is available.}

\author[3]{\fnm{Achal} \sur{Dave}}

\author[1]{\fnm{Wei} \sur{Hua}}

\author[4]{\fnm{Deva} \sur{Ramanan}}

\author*[5]{\fnm{Shu} \sur{Kong}}
\email{skong@um.edu.mo}

\affil[1]{\orgdiv{Department of Computer Science}, \orgname{Zhejiang University}} 

\affil[2]{\orgdiv{Computing \& Mathematical Sciences}, \orgname{Caltech}}

\affil[3]{
\orgname{Toyota Research Institute}}

\affil[4]{\orgdiv{Robotics Institute}, \orgname{Carnegie Mellon University}} 

\affil[5]{\orgdiv{Faculty of Information Science and Computing},  \orgname{University of Macau}}


\abstract{
Contemporary autonomous vehicle (AV) benchmarks have significantly advanced multimodal (LiDAR+RGB) 3D detection. 
However, despite the naturally long-tailed distribution of object classes, existing benchmarking protocols primarily focus on a small set of frequent categories (e.g., {\tt pedestrian} and {\tt car}),
largely overlooking rare but safety-critical classes such as {\tt stroller} and {\tt emergency vehicle}.
In practice, 
reliable detection of both common and rare classes is essential for safe autonomous driving.
We formalize this problem as \emph{Long-Tailed 3D Detection} (LT3D), where evaluation encompasses {\em all} annotated classes, including rare ones.
To address LT3D, we introduce hierarchical losses that promote feature sharing across classes, 
diagnostic metrics that assign partial credit to semantically reasonable mistakes with respect to the semantic hierarchy (e.g., confusing a {\tt child} with an {\tt adult}),
and a multimodal late-fusion (MMLF) framework to fuse detections.
In particular, we show that {\tt rare}-class accuracy benefits substantially from MMLF of independently trained \emph{uni-modal} LiDAR and RGB detectors.
Because of the modular design,
unlike prevailing end-to-end trained multi-modal detectors that require paired LiDAR-RGB data,
MMLF enables the use of advanced unimodal detectors that are trained on large-scale uni-modal datasets with sufficient data for rare classes.
Lastly, 
we  examine three fundamental design choices in MMLF, including the RGB detector representation (2D vs. 3D), cross-modal association (3D vs. image plane), and fusion strategy.
We find that 2D RGB detectors recognize rare classes more reliably than 3D RGB detectors, image-plane association is more robust to depth estimation errors, and probabilistic score-calibrated fusion consistently yields the best performance.
Extensive experiments on nuScenes and Argoverse2 demonstrate substantial improvements of MMLF, establishing a new state of the art.
\vspace{-1mm}
}


\maketitle

\section{Introduction} 
\label{sec:intro}

\begin{figure*}[h]
\centering
\small
\hspace{-90mm} {\tt Few} \hspace{10mm} {\tt Medium} \hspace{10mm} {\tt Many}\\
\includegraphics[width=.45\linewidth, clip, trim={0cm 0cm 0cm 0cm}]{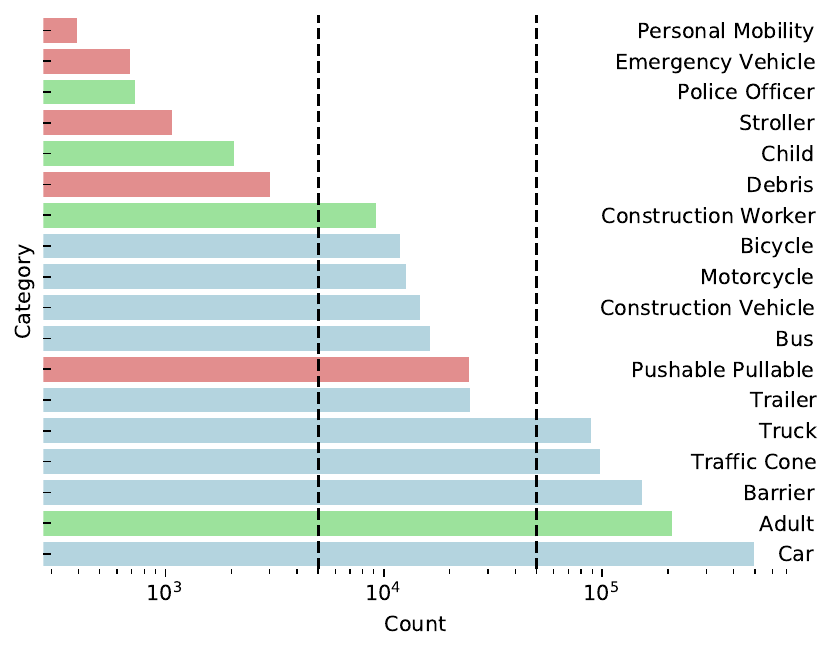} 
\hspace{7mm}
\includegraphics[width=.48\linewidth]{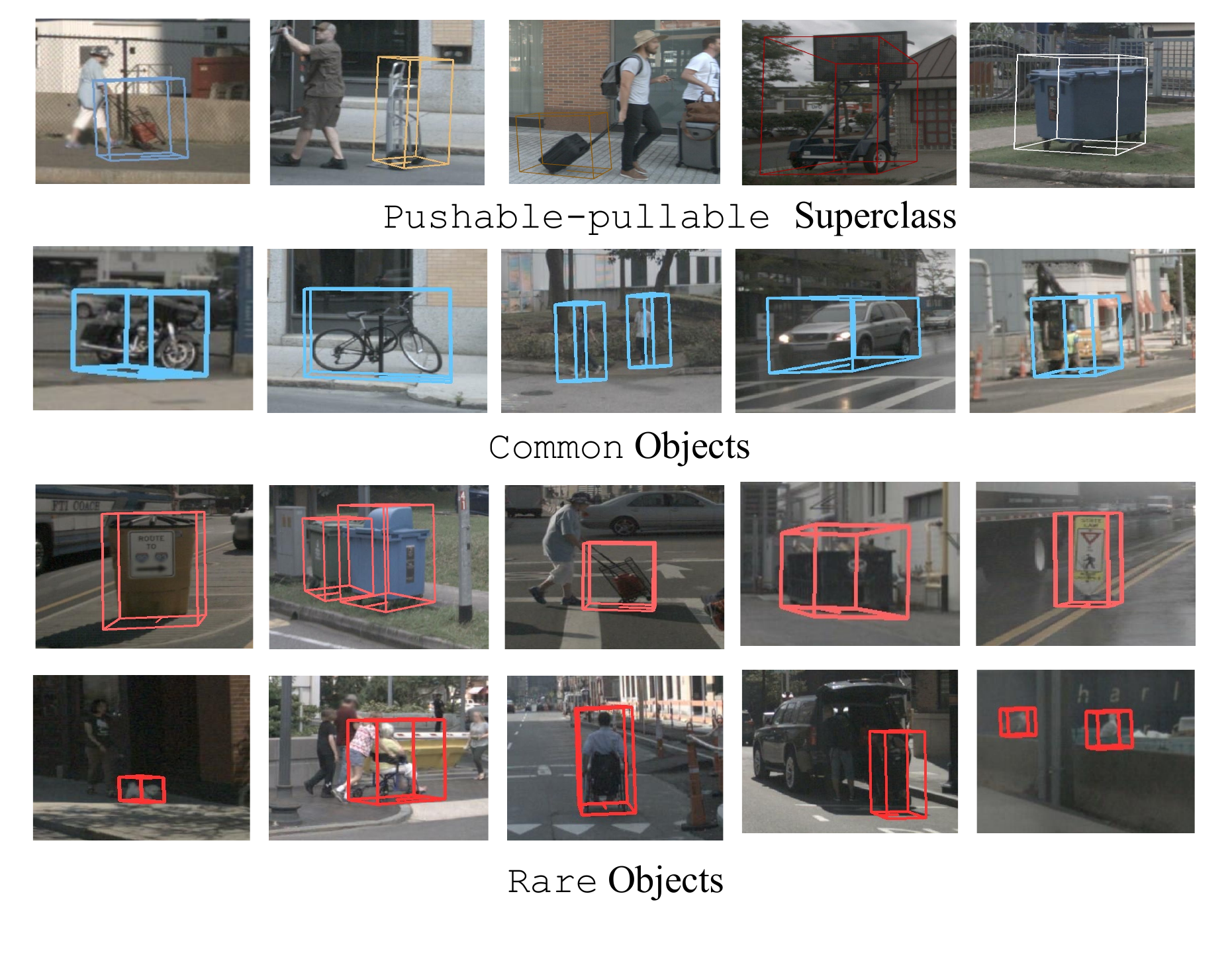} 
\vspace{-1mm}
\caption{\small
\textbf{Left:}
According to the histogram of per-class object counts, 
the nuScenes benchmark \citep{caesar2020nuscenes} focuses on the common classes in {\color{cyan} cyan} (e.g., {\tt car} and {\tt barrier}) and ignores rare ones in {\color{red} red} (e.g., {\tt stroller} and {\tt debris}). 
Moreover, the benchmark creates a superclass {\tt pedestrian} by grouping multiple classes in {\color{darkgreen} green}, including the common class {\tt adult} and several rare ones (e.g., {\tt child} and {\tt police-officer}).
Such a grouping complicates the analysis of detection performance as {\tt pedestrian} performance is dominated by {\tt adult}.
\textbf{Right:} Furthermore, the ignored superclass {\tt pushable-pullable} also contains diverse objects such as {\tt shopping-cart}, {\tt dolly}, {\tt luggage} and {\tt trash-can} as shown in the top row.
Arguably, AVs should detect {\tt rare} classes as they  affect AV behaviors.
To gain a comprehensive analysis of performance,
we report performance in the main paper primarily for three groups of classes ({\tt Many}, {\tt Medium}, and {\tt Few} shown in the left panel) and provide per-class performance in the appendix.}
\label{fig:histogram}
\vspace{-3mm}
\end{figure*}

3D detection is a key component in autonomous vehicles (AVs).
To facilitate research in this space, the AV industry has released multiple large-scale 3D annotated multi-modal datasets~\citep{geiger2012we, chang2019argoverse, wilson2021argoverse, sun2020waymo}. 
However, these datasets often benchmark on only a few {\tt common} classes (e.g., {\tt car} and  {\tt pedestrian}) and ignore {\tt rare} classes (despite being annotated by some datasets \citep{caesar2020nuscenes}) like {\tt stroller} and {\tt emergency vehicle} (Fig.~\ref{fig:histogram}).
In the real world, safe navigation requires AVs to reliably detect {\tt rare}-class objects~\citep{taeihagh2019governing, wong2020identifying}.
This motivates us to study {\em Long-Tailed 3D Detection} (LT3D), a problem that requires detecting objects of both {\tt common} and {\tt rare}  classes.

\begin{figure}[t]
\centering
\includegraphics[width=\linewidth, clip, trim={0cm 0cm 0cm 0cm}]{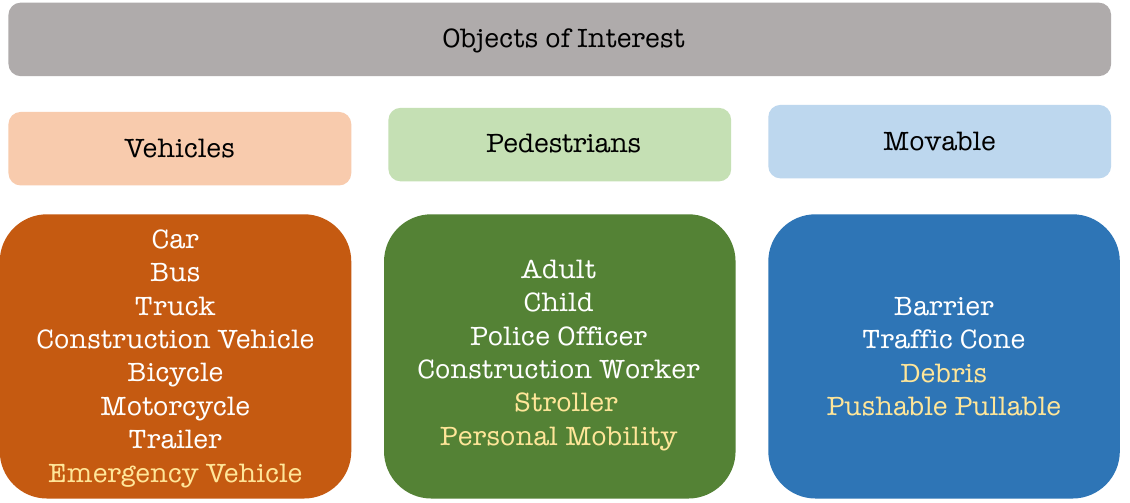} 
\vspace{-3mm}
\caption{\small
The nuScenes dataset defines a hierarchy for the annotated classes (Fig.~\ref{fig:histogram}). 
We mark {\tt common} classes in white and {\tt rare} classes in gold.
The standard nuScenes benchmark makes two choices for dealing with rare classes: (1) ignoring them (e.g., {\tt stroller} and {\tt pushable-pullable}), or (2) grouping them into super classes (e.g., {\tt adult}, {\tt child}, {\tt construction-worker}, {\tt police-officer} are grouped as {\tt pedestrian}).
As the {\tt pedestrian} class is dominated by {\tt adult} (Fig.~\ref{fig:histogram}), the standard benchmarking protocol masks the challenge of detecting rare classes like {\tt child} and {\tt police-officer}. 
}
\label{fig:hierarchy}
\end{figure}

\textbf{Status Quo}.
Among existing AV datasets,
nuScenes~\citep{caesar2020nuscenes} exhaustively annotates objects of various classes crucial for safe navigation (Fig.~\ref{fig:histogram})
and organizes them into a semantic hierarchy (Fig.~\ref{fig:hierarchy}).
We are motivated to study LT3D by re-purposing {\em all} annotated classes in nuScenes because detecting {\tt rare} classes is useful for downstream tasks such as motion planning.
Importantly, LT3D is not simply solved by retraining state-of-the-art methods on both {\tt common} and {\tt rare} classes \citep{peri2022towards}.
For example, 
CCF~\citep{wu2026ccf}, 
a recent multi-modal transformer-based detector, achieves only 9.2 mAP on {\tt rare} categories despite achieving 79.8 mAP on {\tt common} ones (Table~\ref{tab:quick_benchmarking_results}).
We significantly extend our previous work \citep{peri2022towards}, which introduces LT3D for the first time,
by 
(1) delving into RGB detectors and studying whether to fuse 2D or 3D RGB detectors;
(2) delving into fusion and studying whether to fuse in 2D image plane or 3D LiDAR space;
(3) delving into fusion techniques and proposing score calibration and probabilistic fusion;
(4) rigorously comparing recently published 3D detectors.

\textbf{Protocol}. 
LT3D  requires 3D localization and recognition of objects from each of the {\tt common} 
and {\tt rare} classes.
Moreover, for safety-critical robots such as AVs, detecting but mis-classifying {\tt rare} objects (e.g., confusing a {\tt child} with an {\tt adult}) is preferable to failing to detect them at all.
Therefore, we propose a new diagnostic metric to quantify the severity of classification mistakes in LT3D that exploits inter-class relationships w.r.t to the semantic hierarchy 
(Fig.~\ref{fig:hierarchy}). 
We use both the standard and proposed metrics to evaluate 3D detectors on all classes. 
Further, as prior works focus only on a few {\tt common} classes, they miss opportunities to exploit this  hierarchy in training.
We propose hierarchical losses to promote feature sharing across  {\tt common} and {\tt rare} classes, achieving improved performance.

\begin{figure}[t]
\centering
\includegraphics[width=1\linewidth]{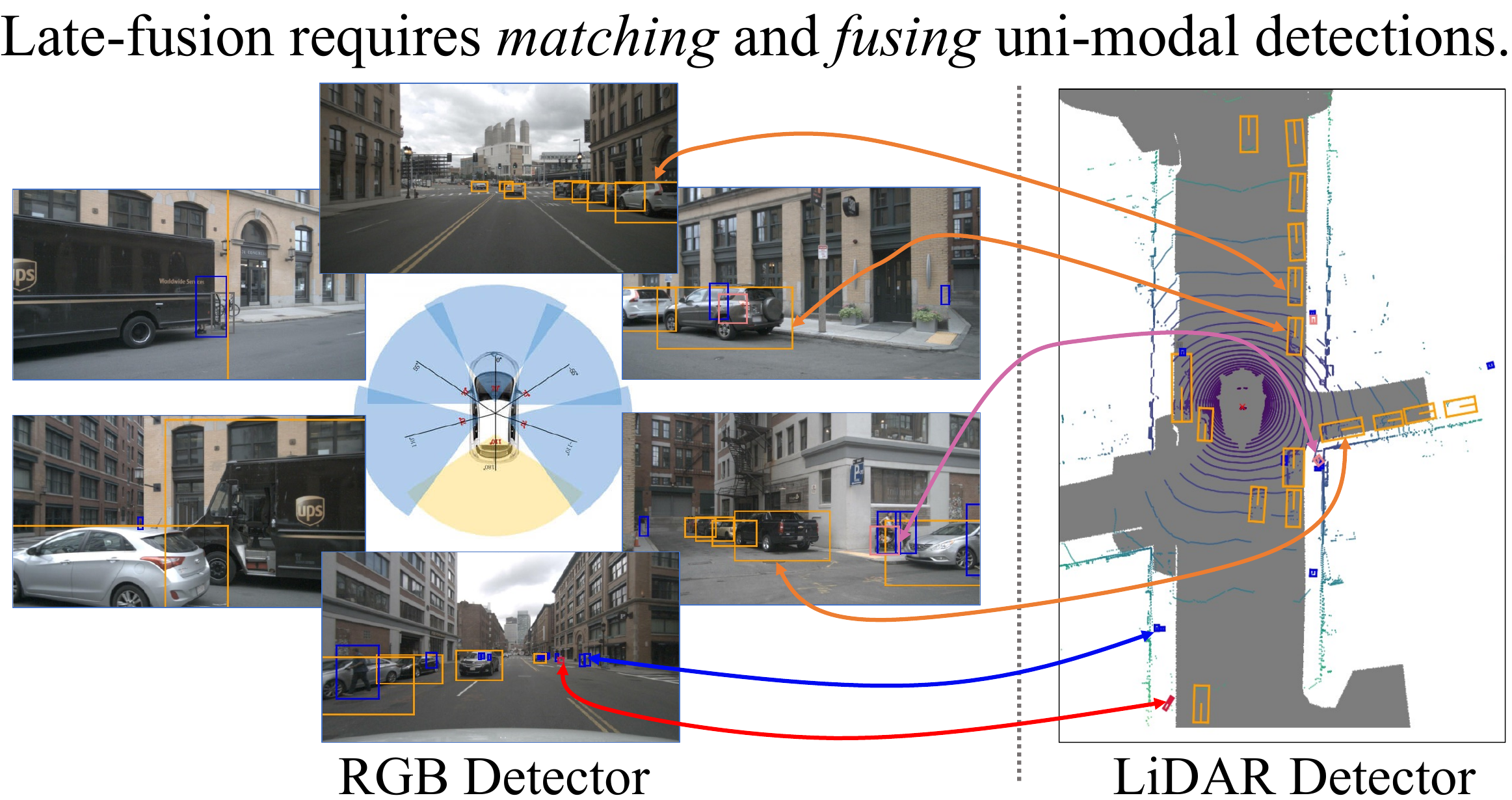}
\vspace{-3mm}
\caption{\small
We adopt a \emph{Multi-Modal Late-Fusion} (MMLF) framework for LT3D by fusing RGB and LiDAR uni-modal detectors.
We rigorously examine three fundamental design choices within MMLF (Fig.~\ref{fig:three-questions}) and propose a simple method that fuses detections produced by a 2D RGB-detector and a 3D LiDAR-detector.
Our method achieves 57.0 mAP on the nuScenes LT3D benchmark, 
significantly improving over the recent 3D detector CCF \citep{wu2026ccf} by 9.9 mAP and the LT3D dedicated approach FOMO-3D \citep{yang2026fomo} by 2.4 mAP
(Table~\ref{tab:quick_benchmarking_results}).
}
\label{fig:flowchart}
\end{figure}

\begin{figure*}[t]
\centering
\includegraphics[width=1\linewidth]{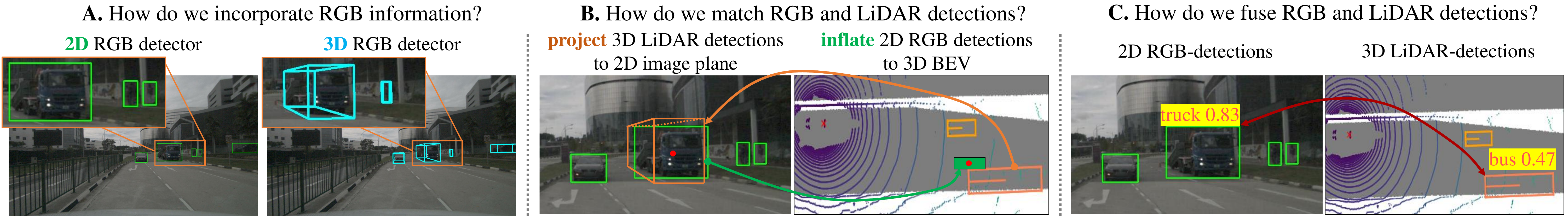}
\vspace{-4mm}
\caption{\small
We examine three key components in multi-modal late-fusion (MMLF) from first principles:
{\bf A.} whether to train a 2D or a 3D RGB detector, 
{\bf B.} whether to match uni-modal detections on the 2D image plane or in the 3D bird's-eye-view (BEV), and
{\bf C.} how to optimally fuse these detections.
We show that using 2D RGB detectors, matching on the 2D image plane, and probabilistically fusing score-calibrated detections achieves state-of-the-art LT3D performance (Table~\ref{tab:analysis_2D-detectors} and Table~\ref{tab:analysis-fusion}). 
}
\label{fig:three-questions}
\end{figure*}

\textbf{Technical Insights}.
To address LT3D, we start by retraining established 3D detectors on
{\em all} classes. 
Surprisingly, these detectors perform rather poorly on {\tt rare} classes, e.g., CenterPoint~\citep{yin2021center} achieves only 0.1 AP on {\tt child} and {\tt stroller} (Table~\ref{tab:ap_h}).
We propose several algorithmic innovations to improve the results. 
First, we allow feature sharing across {\tt common} and {\tt rare} classes by training a single feature trunk, adding in hierarchical losses that ensure features will be useful for all classes (Table~\ref{tab:analysis-fusion}).
Second, noting that LiDAR data is too impoverished even for humans to recognize rare objects that tend to be small (e.g., {\tt strollers}), we propose a \emph{Multi-Modal Late-Fusion} (MMLF) framework (Fig.~\ref{fig:flowchart}),
which fuses detections from a LiDAR-only detector (for precise 3D localization) and an RGB-only detector (for better recognition).
Within MMLF, we first introduce a simple method Multi-Modal Filtering (MMF), which post-processes predictions from 3D LiDAR-only (e.g., CenterPoint \citep{yin2021center}) and 3D RGB-only (e.g., FCOS3D \citep{wang2021fcos3d}) detectors, filtering  detections that are inconsistent across modalities. This significantly improves LT3D performance by 3.2 mAP (Table~\ref{tab:hierarchy}).
Next, we delve into this MMLF framework and study three fundamental design choices (Fig.~\ref{fig:three-questions}):
{\bf (A)} whether to train a 2D or a 3D  RGB detector, 
{\bf (B)} whether to match detections on the 2D image plane or in 3D, and
{\bf (C)} how to fuse these detections.
We reveal that using a 2D RGB detector, matching on the 2D image plane, and probabilistically fusing score-calibrated detections yields state-of-the-art performance, significantly outperforming existing 3D detectors (Table~\ref{tab:quick_benchmarking_results}).

\textbf{Contributions}.
We make three major contributions.
First, we formulate the problem of LT3D, emphasizing the detection of both {\tt common} and {\tt rare} classes in safety-critical applications like AVs.
Second, we design LT3D's benchmarking protocol and develop a diagnostic metric that awards partial credit depending on the severity of misclassifications (e.g., confusing a {\tt child} with an {\tt adult} is less problematic than misclassifying {\tt child} as a {\tt car}). 
Third, we propose several architecture-agnostic approaches to address LT3D, including a simple, non-learned and modular strategy \emph{Multi-Modal Late-Fusion} (MMLF) that generalizes across different RGB and LiDAR architectures (Appendix \ref{sec:transformer-based}). 
We conduct extensive experiments to ablate our design choices and demonstrate that our MMLF framework achieves state-of-the-art  on the nuScenes and Argoverse 2 LT3D benchmarks.

\section{Related Work}
\label{sec:related-work}

\subsection{3D Object Detection for AVs}

Existing approaches are inspired by prior work in 2D detection \citep{xingyi2020centertrack, liu2016ssd, detr2020carion} and can be broadly classified as LiDAR-only, RGB-only, and sensor-fusion methods. 
For example,
LiDAR-based detectors like PointPillars \citep{lang2019pointpillars}, CBGS \citep{zhu2019class}, and PVRCNN++ \citep{shi2022pv} adopt an SSD-like architecture~\citep{liu2016ssd} that regresses amodal bounding boxes from a bird's-eye-view (BEV) feature map.
More recently, CenterPoint \citep{yin2021center} adopts  a center-regression loss that is inspired by CenterNet \citep{xingyi2020centertrack}.
Recent LiDAR-based detectors adopt the transformer architecture or attention modules that achieve further improvements \citep{jin2025unimamba, liu2025fshnet}.
Despite significant progress, LiDAR-based detectors often produce many false positives because it is difficult to distinguish foreground objects from the background given sparse LiDAR returns.

Monocular RGB-based methods have gained popularity in recent years due to increased interest in camera-only perception. FCOS3D \citep{wang2021fcos3d} extends FCOS \citep{tian2019fcos} by additionally regressing the size, depth, and rotation for each object. More recently, methods such as BEVDet, BEVFormer, and DenseBEV \citep{huang2021bevdet, huang2022bevdet4d, li2022bevformer, dahling2026densebev} construct a BEV feature-map by estimating the per-pixel depth of each image feature \citep{philion2020lift}.
PolarFormer \citep{jiang2022polarformer} introduces a polar-coordinate transformation that improves near-field detection. Many of these 3D RGB detectors are commonly pre-trained on large external datasets like DDAD \citep{guizilini20203d}. 
Monocular RGB detectors accurately classify objects but struggle to estimate depth, particularly for far-field detections~\citep{gupta2023far3det}. 
Despite recent advances in LiDAR and RGB 3D detectors, we find that multi-modal fusion is essential for LT3D (detailed next). 
Importantly, using both RGB (for better recognition) and LiDAR (for better 3D localization) helps detect rare classes.
Our work focuses on multi-modal late-fusion (Fig.~\ref{fig:flowchart}) and studies how to effectively fuse RGB and LiDAR uni-modal detectors for LT3D.

\subsection{Multi-modal 3D Detection}
Multi-modal fusion for 3D detection is an active area \citep{bai2022transfusion, liu2022bevfusion, yin2024fusion}. 
Existing methods can be categorized as input-fusion, feature-fusion, and late-fusion. Input-fusion  typically augments LiDAR points using image-level features. 
For example, PointPainting \citep{vora2020pointpainting} projects LiDAR points onto the output mask of a semantic segmentation model and appends corresponding class scores to each point. MVP \citep{yin2021multimodal} densifies regions of LiDAR sweeps that correspond to objects in semantic segmentation masks. Frustum PointNets \citep{qi2018frustum} leverages 2D RGB detections to localize objects within the box frustum using PointNets \citep{qi2017pointnet}. 

Recent works show that feature-fusion is more effective than input-fusion. 
\citep{xu2018pointfusion} fuses global image and point-cloud features prior to detection;
\citep{jiao2022msmdfusion, cai2025dstr} fuse LiDAR and RGB features at multiple scales.
TransFusion \citep{bai2022transfusion} and BEVFusion \citep{liu2022bevfusion} fuse features in the BEV space using multi-headed attention. 
Despite the success of transformers for detecting common objects, \citep{peri2022towards} finds that TransFusion struggles to detect rare classes, and posits that the transformer architecture, as adopted in TransFusion and BEVFusion, suffers from limited training data (particularly for classes in the long tail). For transformers to work well in practice, they should be trained on diverse, large-scale datasets \citep{dosovitskiy2020image,  radford2021learning, wu2026ccf}. It is also worth noting that end-to-end trained multi-modal detectors require paired multi-modal data for training,
increasing the cost of data collection and modality alignment.

We focus on the late-fusion strategy, which ensembles uni-modal detectors and does not require aligned RGB-LiDAR paired training data.
We note that CLOCs \citep{pang2009clocs} and FOMO-3D \citep{yang2026fomo} also adopt late-fusion to combine RGB and LiDAR detections, but they train networks to fuse.
Differently,
ours is non-learned, probabilistic, and modular in design.

\subsection{Long-Tailed Perception}

AV datasets follow a long-tailed class distribution: a few classes like {\tt car} and {\tt pedestrian} are dominant, while others like {\tt stroller} and {\tt debris} are rarely seen. In fact, Long-Tailed Perception (LTP) is not a unique problem in the AV domain~\citep{reed2001pareto} but a long-standing problem in the literature~\citep{openlongtailrecognition, parashar2024neglected} and has been widely studied through the lens of image classification, aiming for high accuracy averaged across imbalanced classes~\citep{openlongtailrecognition, zhang2021deep, alshammari2022long}.
Existing LTP methods propose reweighting losses~\citep{cui2019class, khan2017cost, huang2019deep, zhang2021distribution}, rebalancing data sampling~\citep{drummond2003c4, chawla2002smote, han2005borderline}, balancing gradients computed from imbalanced classes~\citep{tang2020long}, and balancing network weights~\citep{alshammari2022long}. Others study LTP through the lens of 2D object detection~\citep{gupta2019lvis}. 

Compared to long-tailed image classification, long-tailed 3D detection provides unique opportunities and challenges as sensors such as LiDAR directly provide geometric and ego-motion cues that are difficult to extract from 2D images. Further, 2D detectors must detect objects of different scales due to perspective image projection, dramatically increasing the complexity of the output space (e.g., requiring more anchor boxes). In contrast, 3D objects do not exhibit as much scale variation, but far-away objects tend to have sparse LiDAR returns~\citep{gupta2023far3det, peri2023empirical}, imposing different challenges. Finally, 3D detectors often use class-aware heads (i.e. each class has its own binary classifier) while 2D long-tail recognition approaches typically use shared softmax heads.

Recently, CBGS \citep{zhu2019class} explicitly addresses rare-class 3D detection by up-sampling LiDAR-sweeps with instances of rare classes and pasting instances of rare objects copied from different scenes. Although this works well for improving detection of infrequently-seen classes (e.g. classes with {\tt medium} number of examples like {\tt bicycle} and {\tt construction} {\tt vehicle}), it does not provide significant improvement for classes with only a {\tt few} examples like {\tt debris} and {\tt stroller} \citep{peri2022towards}. Additionally, rare classes, such as {\tt child} and {\tt stroller}, are typically small in size and have a limited number of LiDAR returns. 
As a result, LiDAR-only detectors struggle to accurately recognize these rare classes.
In this work, we address LT3D by multi-modal late-fusion, which ensembles RGB and LiDAR uni-modal detectors.

\section{LT3D Benchmarking Protocol}
\label{sec:protocol}

Conceptually, LT3D extends the traditional 3D detection problem, which focuses on identifying objects from  $K$ {\tt common} classes, by further requiring detection of $N$ {\tt rare} classes.
LT3D emphasizes detection performance on {\em all} classes.
In the main paper, we report the metrics for three groups of classes; we provide per-class performance in the appendix.
We group the classes based on their cardinality (Fig.~\ref{fig:histogram}-left): {\em many} ($>$50k instance/class), {\em medium} (5k$\sim$50k instance/class), and {\em few} ($<$5k instance/class). 
We present two metrics below and also evaluate NDS in Appendix \ref{sec:NDS}.

\textbf{Mean Average Precision (mAP)} is a 
well-established metric for object detection~\citep{everingham2015pascal, geiger2012we, Lin2014MicrosoftCC}.
For 3D detection on LiDAR sweeps, a true positive is defined as a detection that has a center distance within a distance threshold to a ground-truth annotation~\citep{caesar2020nuscenes, ma2026roadside}.
The mAP computes the mean of AP over classes, where per-class AP is the averaged area under the precision-recall curves with distance thresholds of [0.5, 1, 2, 4] meters.

\textbf{Hierarchical Mean Average Precision (mAP$_{H}$)}.
For safety-critical applications, 
it is more desirable to localize-but-misclassify an object than fail to detect this object, e.g., detecting but misclassifying a {\tt child} as {\tt adult} is better than not detecting this {\tt child}. 
Therefore, we propose hierarchical mean average precision (mAP$_H$), which considers such semantic relationships across classes to award partial credit. 
We use AP$_{H}$ as a supplementary diagnostic tool to analyze how LT3D detectors make mistakes.
To encode inter-class relationships, we leverage the semantic hierarchy (Fig.~\ref{fig:hierarchy}) such as the one defined by nuScenes~\citep{caesar2020nuscenes}. 
We derive partial credit as a function of semantic similarity using the least common ancestor (LCA) distance metric.
LCA has been proposed for image classification \citep{russakovsky2015imagenet, shi2024lca} but not for object detection. 
Extending this metric to detection is non-trivial as we must jointly evaluate semantic and spatial overlap. Below, we explain how to compute AP$_H$ for a class $c$.

{\bf LCA=0}: Consider the detections and ground-truth boxes for $c$. Label the set of detection that overlap with ground-truth boxes for $c$ as true positives. Other predictions are false positives. {\em This is identical to the standard mAP metric.}

{\bf LCA=1}: 
Consider the detection for $c$, and ground-truth boxes for $c$ and its all sibling classes with LCA distances to $c$ of 1. 
Label the detection that overlap a ground-truth box of $c$ as a true positive. 
Label the detection that overlap sibling classes as {\tt ignored}~\citep{Lin2014MicrosoftCC}. 
All other detections for $c$ are false positives.

{\bf LCA=2}: Consider the detections for $c$ and ground-truth boxes for $c$ and all its sibling classes with LCA distances to $c$ no greater than $2$. For nuScenes, this includes all classes.
Label the set of detections that overlap ground-truth boxes for $c$ as true positives. Label the set of detections that overlap other classes as {\tt ignored}. All other detections for $c$ are false positives.

\textbf{Remarks.}
Although the quantitative metrics above 
provide a meaningful approximation of real-world scenarios from the perspective of object detection,
they do not fully reflect the end-to-end performance of autonomous vehicles.
AV safety depends on much more than detection alone, including multi-object tracking, trajectory forecasting, and motion planning.
Practical deployment therefore requires more comprehensive evaluation protocols that capture the interactions among downstream modules and better assess real-world safety and driving performance.

\section{The Proposed Techniques to Improve LT3D}
\label{sec:methods}

To address LT3D, we first retrain well-established detectors on {\em all} long-tailed classes. We consider the following detectors:
\begin{itemize}[leftmargin=0.5cm, itemsep=0pt, topsep=0.1cm]
\item 
    {\em LiDAR-based 3D Detectors:} CenterPoint~\citep{yin2021center}, PointPillars \citep{lang2019pointpillars},
    TransFusion-L~\citep{bai2022transfusion},
    BEVFusion-L~\citep{liu2022bevfusion},
    CMT-L~\citep{yan2023cross},
    and FSHNet~\citep{liu2025fshnet};
\item
    {\em RGB-based 3D Detectors}: 
    FCOS3D~\citep{wang2021fcos3d},
    PolarFormer~\citep{jiang2022polarformer}, 
    BEVFormer~\citep{li2022bevformer}, and DenseBEV~\citep{dahling2026densebev};    
\item 
    {\em RGB-based 2D Detectors}: 
    YOLOV7~\citep{wang2022yolov7}, 
    DINO~\citep{zhang2022dino}, GroundingDINO~\citep{liu2024grounding}, and Co-DINO~\citep{zong2023detrs};
\item 
    {\em End-to-End Multi-modal 3D Detectors}: TransFusion~\citep{bai2022transfusion},
    BEVFusion~\citep{liu2022bevfusion}, CMT \citep{yan2023cross}, DeepInteraction~\citep{yang2022deepinteraction}, IS-Fusion \citep{yin2024fusion}, EA-LSS \citep{hu2023ea}, and CCF \citep{wu2026ccf}.
\end{itemize}
In Section~\ref{ssec:group-free-head} and \ref{ssec:semantic-hierarchy},
we introduce two modifications that consistently improve LT3D performance.
In Section~\ref{sec:multimodal-filtering}, we present a simple yet effective late-fusion method.

\subsection{Grouping-Free Detector Head}
\label{ssec:group-free-head}
\begin{figure}[t]
\centering
\includegraphics[width=\linewidth, clip, trim={1cm 3cm 7.5cm 3.5cm}]{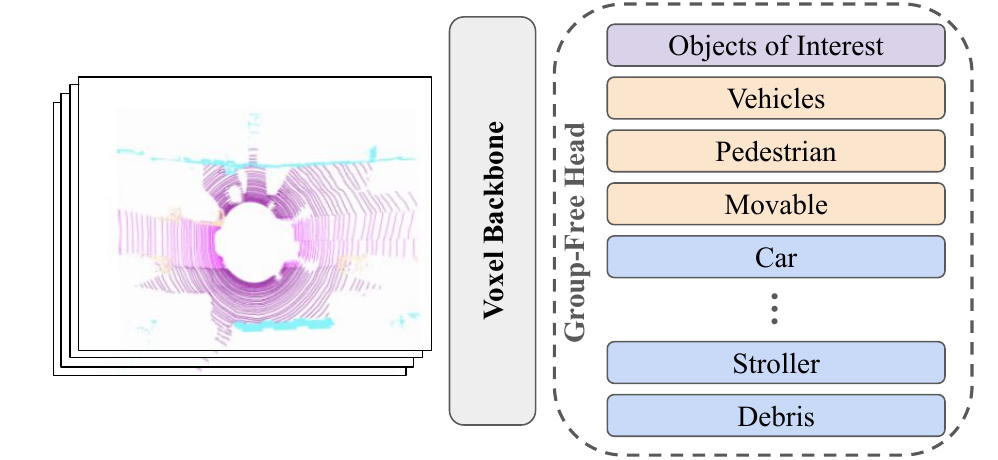} 
\vspace{-4mm}
\caption{\small
We exploit the semantic hierarchy defined in the nuScenes dataset (Fig.~\ref{fig:hierarchy}) to train detectors by predicting class labels at {\em multiple} levels for an object: 
the fine-grained label (e.g., {\tt child}), the coarse class (e.g., {\tt pedestrian}), and the root-level class {\tt object}. That said, the classes in the hierarchy are no longer mutually exclusive.
This allows us to use a sigmoid focal loss to learn a separate heatmap for each class.
}
\label{fig:hierarchy_train}
\end{figure}

Extending existing 3D detectors to train with more classes turns out to be  surprisingly challenging. 
Many networks use a multi-head architecture that groups classes of similar size and shape to facilitate efficient feature sharing~\citep{zhu2019class, yin2021center}. 
For example, CenterPoint~\citep{yin2021center} groups {\tt pedestrian} and {\tt traffic-cone} since these objects are both tall and skinny.
This multi-head grouping design may not work for super-classes like {\tt pushable-pullable} and {\tt debris} that contain diverse objects of different sizes and shapes.
Moreover, in the multi-head design, each head is a group-specific detector that consists of several layers with lots of parameters.
This makes multi-head grouping difficult to scale for a large number of classes.
To address these issues, we treat each class as its own group to avoid hand-crafted grouping heuristics and design a grouping-free strategy, in which each class has only one linear layer as its detector (Fig.~\ref{fig:hierarchy_train}) and all classes share a single detector head.
This design significantly reduces the number of parameters and allows learning the shared feature backbone collaboratively with all classes, effectively mitigating overfitting to rare classes. Adding a new class is as simple as adding a single linear layer to the detector head. 
Importantly, we show that our grouping-free detector outperforms conventional grouping-based methods in Appendix \ref{sec:analysis-class-grouping}.

\subsection{Training with a Semantic Hierarchy}
\label{ssec:semantic-hierarchy}
To handle severity of inter-class misclassifications in autonomous driving, the nuScenes dataset defines a semantic hierarchy (Fig.~\ref{fig:hierarchy}) that  groups semantically similar classes under coarse-grained categories~\citep{caesar2020nuscenes}. We leverage this hierarchy during training.
Specifically, we train detectors to predict three labels for each object: its fine-grained class (e.g., {\tt child}), its coarse class (e.g., {\tt pedestrian}), and the root class {\tt object}.
We adopt a grouping-free detector head (Fig.~\ref{fig:hierarchy_train}) that outputs separate ``multitask'' heatmaps for each class, and use a \emph{per-class sigmoid focal loss}~\citep{lin2017focal} rather than multi-class cross-entropy loss to avoid normalizing class probabilities across fine-grained and super-classes. 
It is worth noting that our approach
does not explicitly enforce a tree-like hierarchy, and can be applied to more complex label relationships~\citep{lin2022continual}.
Crucially, adding a {\tt vehicle} heatmap does not directly interfere with the {\tt car} heatmap. 
Moreover, since this produces repeated detections on the same object (e.g., a {\tt car} is detected as a {\tt car}, a {\tt vehicle}, and an {\tt object}), at test time, we simply ignore coarse detections. 
We explore alternatives in Appendix \ref{sec:analysis-hierarchical-training} and conclude that they achieve similar performance.
Notably, our method improves detection performance not only for {\tt rare} classes, but also for {\tt common} classes (Table~\ref{tab:analysis-fusion}).

\subsection{Multi-Modal Filtering for Detection Fusion} 
\label{sec:multimodal-filtering}

\begin{figure}[t]
\centering
{\footnotesize
\ \hspace{-2.1mm} LiDAR detections \hspace{3mm} RGB detections \hspace{3mm} Filtered results \hspace{1.1mm} \ }
\\
\includegraphics[trim=0cm 8cm 2.5cm 0cm,clip, width=\linewidth]{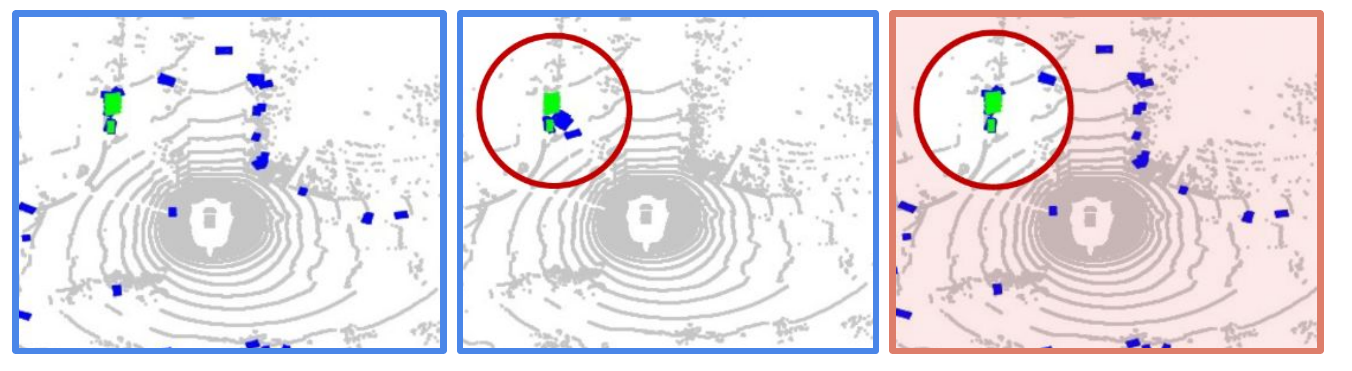}
\vspace{-4mm}
\caption{\small
The proposed method \textit{Multi-Modal Filtering (MMF)} effectively removes high-scoring false-positive LiDAR detections.
The green boxes are ground-truth {\tt strollers}, while the blue boxes are {\tt stroller} detections from the LiDAR-based detector CenterPoint~\citep{yin2021center} ({\bf left}) and RGB-based detector FCOS3D~\citep{wang2021fcos3d} ({\bf mid}). 
MMF removes LiDAR detections not within $m$ meters of any RGB detection, shown in the red region of {\bf right}, and keeps all other LiDAR detections.} 
\label{fig:filtering}
\end{figure}

Rare-class instances are often small
and can be challenging to recognize from sparse (LiDAR) geometry alone: even humans struggle to find {\tt strollers} in LiDAR point clouds.
As a result, 
a LiDAR-based 3D detector tend to misclassify rare-class objects with high confidence, producing many high-scoring false positives (FPs).
Even so, LiDAR-based 3D detectors can achieve very high recall and precise 3D localizations for true positives (TPs) (Appendix \ref{sec:analysis-lidar-recall}).
On the other hand,
RGB-based detections are accurate w.r.t recognition but their 3D localizations are poor. 
Leveraging the strengths of a LiDAR-based 3D detector and an RGB-based 3D detector,
we introduce a Multi-Modal Filtering (MMF) strategy, illustrated in Fig.~\ref{fig:filtering}.
Briefly, for each RGB detection, we keep LiDAR detections within a radius of $m$ meters and remove all the other LiDAR detections.
We denote this method as MMF($D_L$, $D_R$),
where D$_L$ and $D_R$ are any 3D LiDAR detector and 3D RGB detector, respectively.
Table~\ref{tab:hierarchy} demonstrates that MMF greatly improves LT3D.

\begin{figure}[t]
\centering
\hspace{0mm} {\tt Vehicle} \hspace{8mm} {\tt Pedestrian} \hspace{6mm} {\tt Movable} \hspace{0mm} \\
\adjustbox{valign=t}{\includegraphics[width=0.32\linewidth]{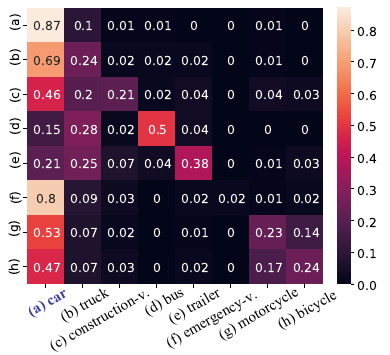}} \hfill
\adjustbox{valign=t}{\includegraphics[width=0.32\linewidth]{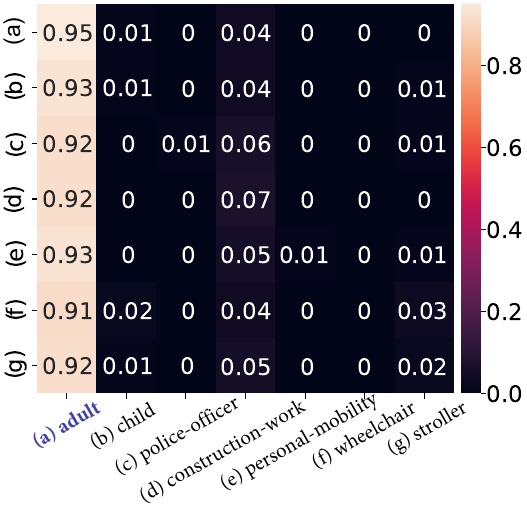}} \hfill
\adjustbox{valign=t}{\includegraphics[width=0.32\linewidth]{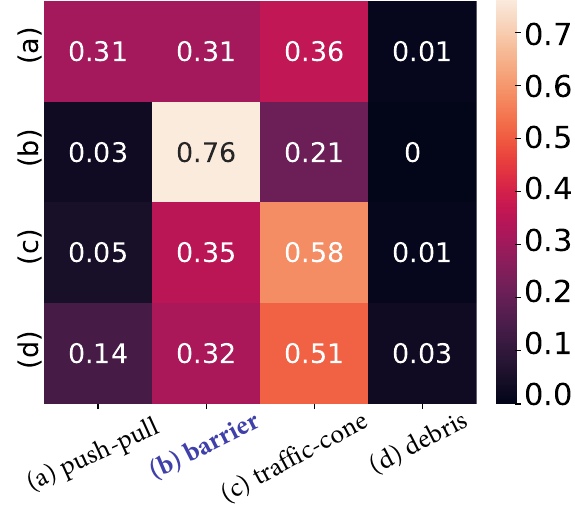}} \hfill
\vspace{-2mm}
\caption{\small 
We analyze the MMF approach. 
Rare classes are most often confused by the {\tt common} class (in {\bf \color{blue}blue}) in each superclass: ({\bf left}) {\tt Vehicle} is dominated by {\tt car}, ({\bf mid}) {\tt Pedestrian} is dominated by {\tt adult}, and ({\bf right}) {\tt Movable} is dominated by {\tt barrier}.
Note that class confusions are quite reasonable. {\tt Car} is often mistaken for {\tt truck}. Similarly, {\tt truck}, {\tt construction-vehicle} and {\tt emergency-vehicle} are most often mistaken for {\tt car}. 
{\tt Bicycle} and {\tt motorcycle} are sometimes misclassified as {\tt car}, presumably because they are sometimes spatially close 
to {\tt cars}.
{\tt Adults} have similar appearance to {\tt police-officer} and {\tt construction-worker}, and they are often co-localized with {\tt child} and {\tt stroller}.
}
\label{fig:conf_matrix}
\end{figure}

\textbf{Analysis}.
Our MMF's performance is largely affected by the classification accuracy of FCOS3D when matching LiDAR and RGB-based detections.
A low classification accuracy can produce many correctly localized but misclassified detections. 
We use a confusion matrix to further analyze the misclassifications within each superclass, as shown in Fig.~\ref{fig:conf_matrix}. 
We explain how to compute a confusion matrix for the detection task:
For each superclass, we construct a confusion matrix, in which the entry $(i,j)$ indicates the misclassification rate of class-$i$ objects as class-$j$.
Specifically, given a class $i$, we find its predictions that match ground-truth boxes within a 2m center-distance of class-$i$ and all its sibling classes (LCA=1, within the corresponding superclass); we ignore all unmatched detections.
This allows us to count the misclassifications of class-$i$ objects into class-$j$. We normalize these counts to produce misclassification rates. 
The analysis shows that rare classes are most often confused by the dominant class in each superclass: {\tt Vehicle} is dominated by {\tt car}, {\tt Pedestrian} is dominated by {\tt adult}, and {\tt Movable} is dominated by {\tt barrier}. We posit that this can be addressed with more accurate RGB-based classifiers,
motivating our study of \emph{multi-modal late-fusion} (MMLF), presented next.

\section{Delving into Multi-Modal Late-Fusion for LT3D}
\label{ssec:multi-modal-late-fusion}
Motivated by the effectiveness of our frustratingly simple MMF,
we delve into this multi-modal late-fusion (MMLF) strategy. 
We explore three questions from first principles (Fig.~\ref{fig:three-questions}): (1) how to effectively incorporate RGB information,
(2) how to match RGB and LiDAR detections, and (3) how to fuse them. 
We elaborate on them next.

\subsection{How Do We Incorporate RGB Information?}
\label{ssec:detectors}
Although LiDAR offers accurate localization, LiDAR-only detectors struggle to recognize objects using sparse LiDAR alone. RGB images provide complementary information that is essential for identifying objects and disambiguating those that are geometrically similar in point clouds but semantically different in images, e.g., {\tt construction}-{\tt worker} vs. {\tt police}-{\tt officer}. 
Building on the late-fusion strategy, 
with a fixed 3D LiDAR detector,
we explore \emph{fusing a 2D RGB detector}.
Surprisingly, this yields more significantly better performance than fusing a 3D RGB detector. 
We present insights on why using a 2D detector to incorporate RGB information is better and ablate the impact of using 2D vs. 3D RGB detectors for late-fusion in Table~\ref{tab:analysis_2D-detectors}.

\begin{figure}[t]
    \centering    \includegraphics[width=\linewidth, clip, trim={1.5cm 5cm 1.5cm 6cm}]{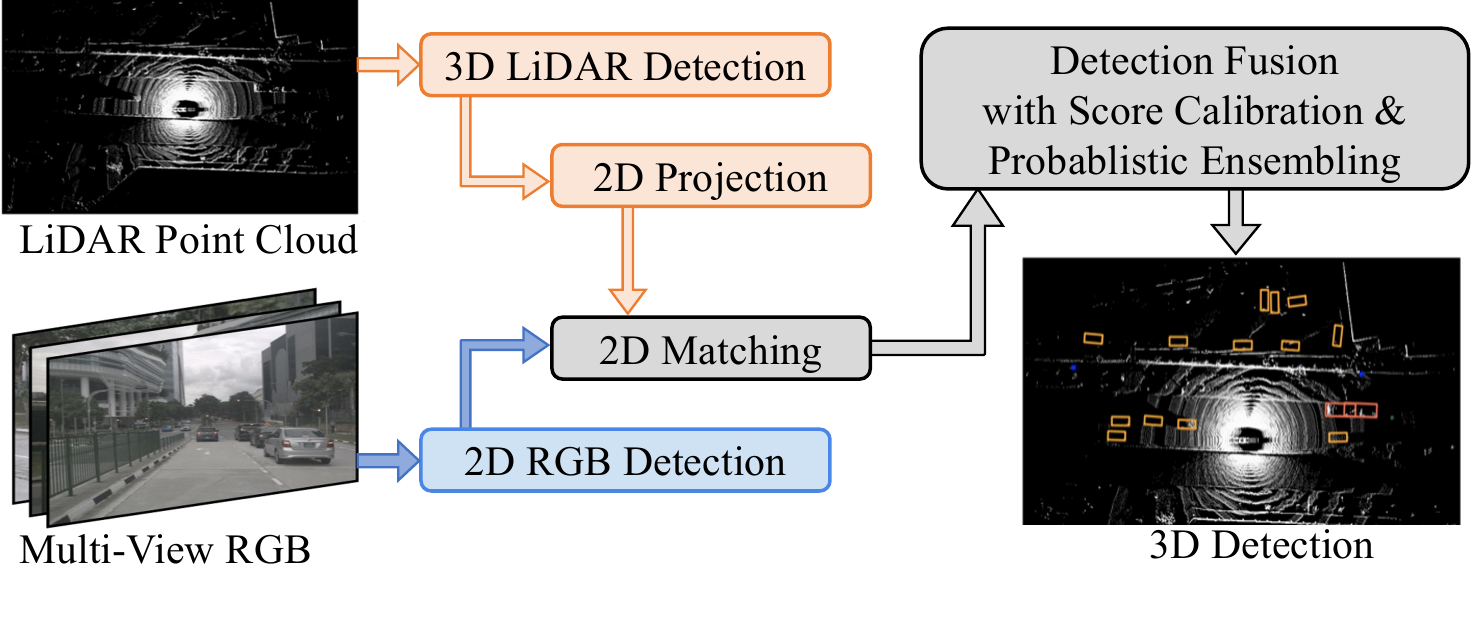}
    \vspace{-6mm}
    \caption{\small
    Our \emph{Multi-Modal Late-Fusion} (MMLF) strategy takes 3D LiDAR and 2D RGB detections as input (Fig.~\ref{fig:three-questions}A), matches the detections on the image plane (Fig.~\ref{fig:three-questions}B), 
    and probabilistically fuses score-calibrated detections towards the final 3D detections.}
    \label{fig:pipeline_diag}
\end{figure}

{\bf 2D RGB detectors are more mature}. 
2D object detection is a fundamental problem in computer vision~\citep{felzenszwalb2009object, Lin2014MicrosoftCC, NIPS2015_fasterRCNN} that has matured in recent years and model trade-offs are well understood~\citep{NIPS2015_fasterRCNN, liu2016ssd, redmon2016you, lin2017focal}. 
Our work considers several well-known 2D RGB detectors: YOLOV7~\citep{wang2022yolov7}, DINO~\citep{zhang2022dino}, 
GroundingDINO \citep{liu2024grounding}, and 
Co-DINO \citep{zong2023detrs}.

\textbf{2D RGB detectors can be trained with more diverse data.} 
Training 2D RGB detectors only requires \emph{2D bounding box} annotations, which are significantly cheaper to collect than 3D cuboids used for training 3D RGB detectors~\citep{wang2021fcos3d, jiang2022polarformer, ma2025towards}. Since annotating 3D amodal cuboids is both expensive and non-trivial (compared to bounding-box annotations for 2D detection), datasets for monocular 3D RGB detection are considerably smaller and less diverse than their 2D detection counterparts.  For example, nuScenes~\citep{caesar2020nuscenes} (published in 2020) annotates 144K RGB images of 23 classes using 3D cuboids, while COCO~\citep{Lin2014MicrosoftCC} (an early 2D detection dataset published in 2014) annotates 330K images of 80 classes using 2D bounding boxes.
This allows us to pretrain 2D RGB detectors on
significantly larger, more diverse, publicly available datasets~\citep{li2022grounded, zhou2022simple, wang2019towards, xu2020universal, redmon2017yolo9000}. 
We demonstrate in Fig.~\ref{fig:curves2} that leveraging existing 2D detection datasets helps train stronger 2D detectors, further improving LT3D performance.

\subsection{How Do We Match Uni-Modal Detections?}
\label{ssec:det-match}

Finding correspondence between two sets of RGB and LiDAR uni-modal detections is an essential step in the late-fusion framework (Fig.~\ref{fig:three-questions}B). 
Our previous  MMF method matches 3D RGB and 3D LiDAR detections using center distance in the 3D space (Fig.~\ref{fig:filtering}). 
However,
precisely matching them in the 3D space is error-prone due to depth estimation errors from the 3D RGB detector.
Hence, we opt to match \emph{2D RGB} (without the need to estimate depth in detection) and 3D LiDAR detections. 
Prior work attempts to inflate 2D detections to 3D for matching by using LiDAR points \citep{wilson20203d}, but we find that matching inflated 2D detections also yields noisy 3D boxes (due to outlier LiDAR points from the background) and reduces overall match quality.
In contrast, we match uni-modal detections by projecting 3D LiDAR detections onto the 2D image plane, avoiding additional noise due to imprecise depth estimates. 
We ablate the impact of matching in 3D versus on the 2D image plane in Table~\ref{tab:analysis_2D-detectors}, and present our 2D matching algorithm below.

{\bf Spatial matching on the 2D image plane}.
Using the available sensor extrinstics, we project 3D LiDAR detections onto the 2D image plane. We then use the IoU metric to determine overlap between projected LiDAR and 2D RGB detections. 
A 2D RGB detection and a projected 3D LiDAR detection are considered a match if their IoU is greater than a threshold. Although conceptually simple, this matching method works significantly better than using center distance to match detections in 3D (cf. bottom four rows in Table~\ref{tab:analysis_2D-detectors}). Spatially matching uni-modal detections using 2D IoU yields three categories of detections: matched detections, unmatched RGB detections (that do not have corresponding LiDAR detections), and unmatched LiDAR detections (that do not have corresponding RGB detections).
We handle unmatched detections below and show how to fuse matched ones in the next subsection.

{\bf Handling unmatched detections}. 
We simply remove unmatched 2D RGB detections,
which are likely to be false positives because LiDAR detectors have already achieved high recall~\citep{peri2022towards} (Appendix \ref{sec:analysis-lidar-recall}).
For unmatched 3D LiDAR detections, we down-weight their detection confidence scores by $w$ (inspired by SoftNMS \citep{bodla2017soft}).
We tune $w$ via grid search on a validation set and set $w=0.4$ in this work.

\subsection{How Do We Fuse Matched Uni-Modal Detections?}
\label{ssec:calib-fuse}

Given a pair of spatially matched RGB and LiDAR detections, we consider two cases: matched detections with semantic disagreement (i.e., detections have different predicted classes), and matched detections with semantic agreement. 
Below, we present our method to address them.

{\bf Addressing semantic disagreement between modalities}.
If the two modalities predict different semantic classes, 
we use the confidence score and class label of the RGB-based detection,
and the 3D box from the LiDAR-based detection. Intuitively, RGB detectors can predict semantics more reliably than LiDAR-only detectors.
This helps correct misclassifications of geometrically similar but semantically different objects produced by the 3D LiDAR detector.
To note, prior late-fusion methods like CLOCS~\citep{pang2009clocs} and our MMF~\citep{peri2022towards} only fuse matched predictions with semantic agreement and do not correct misclassifications.
In contrast, ours handles misclassifications that significantly improves rare class performance (Table~\ref{tab:quick_benchmarking_results}).

{\bf Fusing semantically-agreed detections}.
If both modalities predict the same semantic class, we perform probabilistic fusion~\citep{chen2021multimodal}, presented at the end of this section.
Note that the confidence scores of RGB and LiDAR detections are not directly comparable:
LiDAR-based detectors are often under-confident as it is difficult to distinguish foreground-vs-background using sparse LiDAR alone. Therefore, score calibration is crucial for fusion. Below, we explore score calibration of RGB ($x_{\text{RGB}}$) and LiDAR ($x_{\text{LiDAR}}$) detections.

{\bf Score calibration}.
We calibrate detection confidences per model by tuning a temperature $\tau_c$ for the logit score of class $c$ on a validation set before applying a sigmoid transform~\citep{guo2017calibration, chen2021multimodal}, 
i.e., sigmoid(logit$_c$/$\tau_c$).
To tune  $\tau_c$, we optimize over the $c$-th class AP on the validation set, considering that the final mAP metric evaluates over each class individually.
Note that this score calibration does not increase runtime or complexity. Rather, it is performed only once after training. The tuned $\tau_c$ and $p(c)$ do not need further optimization at test time.
It is also worth noting that score calibration aims to make RGB and LiDAR detections comparable for fusion and it is based on detection mAP over validation set; score calibration is not designed for calibrating posterior probabilities.

{\bf Probabilistic fusion}.
To facilitate the derivation of probabilistic fusion,
we assume conditional independence given the class label $c$,
i.e., 
$p( x_{\text{RGB}}, x_{\text{LiDAR}} | c) = p( x_{\text{RGB}} | c)  p(x_{\text{LiDAR}} | c)$.
Like many assumptions commonly adopted in probabilistic modeling,
this assumption need not hold exactly in practice. 
It is analogous to the widely used Gaussianity assumption, which is rarely satisfied but has proven highly effective for theoretical modeling \citep{dawid1979conditional}.
Now, we compute the final score:
\begin{align}
\small 
p(c | x_{\text{RGB}}, x_{\text{LiDAR}})
& =  \frac{p(x_{\text{RGB}}, x_{\text{LiDAR}} | c) p(c)}{p(x_{\text{RGB}}, x_{\text{LiDAR}})} 
\nonumber
\\
& \propto  \ \ p(x_{\text{RGB}}, x_{\text{LiDAR}} | c)  {p(c)} 
\nonumber
\\
& \propto  \ \ p(x_{\text{RGB}} | c)  p(x_{\text{LiDAR}} | c) {p(c)} 
\nonumber
\\
& \propto  \ \ \frac{p(c | x_{\text{RGB}})  p(c | x_{\text{LiDAR}})}{{p(c)}}
\nonumber
\end{align}%
where $p(c | x_{\text{RGB}})$ and  $p(c | x_{\text{LiDAR}})$ are the posteriors after calibration. 
Unlike assuming balanced class prior distributions in \citep{chen2021multimodal} or using class frequencies based on the training set,
we tune $p(c)$ for each class independently by optimizing over the AP of this class on the validation set.
This is similar to tuning the per-class temperature described above. 
Applying class prior $p(c)$ does not increase inference time.

{
\setlength{\tabcolsep}{1.50mm}
\begin{table*}[t] 
\small
\centering
\scalebox{1}{
\begin{tabular}{lcrrrr}
\toprule 
  \multirow{1}{*}{Method}     & \multicolumn{1}{c}{Modality} & \multicolumn{1}{c}{{\tt All}} & \multicolumn{1}{c}{{\tt Many}} & \multicolumn{1}{c}{{\tt Medium}} & \multicolumn{1}{c}{{\tt Few}}  \\ 
\midrule

\multirow{1}{*}{FCOS3D \citep{wang2021fcos3d} (ICCVW'21)}   
 & \multirow{1}{*}{C} & \multirow{1}{*}{20.9} & \multirow{1}{*}{39.0} & \multirow{1}{*}{23.3} & \multirow{1}{*}{2.9}   \\

\multirow{1}{*}{BEVFormer \citep{li2022bevformer} (ECCV'22)}   
 & \multirow{1}{*}{C} & \multirow{1}{*}{27.3} & \multirow{1}{*}{52.3} & \multirow{1}{*}{31.6} & \multirow{1}{*}{1.4}   \\

\multirow{1}{*}{PolarFormer \citep{jiang2022polarformer} (AAAI'23)}   
 & \multirow{1}{*}{C} & \multirow{1}{*}{28.0} & \multirow{1}{*}{54.0} & \multirow{1}{*}{31.6} & \multirow{1}{*}{2.2}   \\
 
 \multirow{1}{*}{DenseBEV \citep{dahling2026densebev} (WACV'26)}   
 & \multirow{1}{*}{C} & \multirow{1}{*}{29.7} & \multirow{1}{*}{54.9} & \multirow{1}{*}{33.1} & \multirow{1}{*}{4.7}   \\

\midrule

\multirow{1}{*}{CenterPoint \citep{yin2021center} (CVPR'21)}   
& \multirow{1}{*}{L} & \multirow{1}{*}{39.2} & \multirow{1}{*}{76.4} & \multirow{1}{*}{43.1} & \multirow{1}{*}{3.5}   \\

\multirow{1}{*}{TransFusion-L \citep{bai2022transfusion} (CVPR'22)}   
  & \multirow{1}{*}{L} & \multirow{1}{*}{38.5} & \multirow{1}{*}{68.5} & \multirow{1}{*}{42.8} & \multirow{1}{*}{8.4}   \\

\multirow{1}{*}{BEVFusion-L \citep{liu2022bevfusion} (ICRA'23)}   
  & \multirow{1}{*}{L} & \multirow{1}{*}{42.5} & \multirow{1}{*}{72.5} & \multirow{1}{*}{48.0} & \multirow{1}{*}{10.6}   \\

\multirow{1}{*}{CMT-L \citep{yan2023cross} (ICCV'23)}   
 & \multirow{1}{*}{L} & \multirow{1}{*}{34.7}  & \multirow{1}{*}{73.4} & \multirow{1}{*}{35.9} & \multirow{1}{*}{1.1}  \\

\multirow{1}{*}{FSHNet \citep{liu2025fshnet} (CVPR'25)}   
 & \multirow{1}{*}{L} & \multirow{1}{*}{41.6}  & \multirow{1}{*}{79.2} & \multirow{1}{*}{47.6} & \multirow{1}{*}{3.3}  \\
  
\midrule

\multirow{1}{*}{CenterPoint \citep{yin2021center} + RCNN \citep{Girshick_2014_CVPR}}   & C+L & \multirow{1}{*}{34.0} & \multirow{1}{*}{64.8} & \multirow{1}{*}{37.5} & \multirow{1}{*}{4.3}   \\

\multirow{1}{*}{CLOCS (CenterPoint, DINO) \citep{pang2009clocs} (IROS'20)}   
& C+L & \multirow{1}{*}{40.0} & \multirow{1}{*}{68.2} & \multirow{1}{*}{45.7} & \multirow{1}{*}{10.0}   \\

\multirow{1}{*}{CLOCS (CenterPoint, Co-DINO) \citep{pang2009clocs} (IROS'20)}   
& C+L & \multirow{1}{*}{45.2} & \multirow{1}{*}{73.3} & \multirow{1}{*}{50.9} & \multirow{1}{*}{15.1}   \\

\multirow{1}{*}{TransFusion  \citep{bai2022transfusion} (CVPR'22)}   
& C+L & \multirow{1}{*}{39.8} & \multirow{1}{*}{73.9} & \multirow{1}{*}{41.2} & \multirow{1}{*}{9.8}   \\

\multirow{1}{*}{DeepInteraction \citep{yang2022deepinteraction} (NeurIPS '22)}   
& C+L & \multirow{1}{*}{43.7} & \multirow{1}{*}{76.2} & \multirow{1}{*}{51.1} & \multirow{1}{*}{7.9}   \\

\multirow{1}{*}{BEVFusion \citep{liu2022bevfusion} (ICRA'23)}   
& C+L & \multirow{1}{*}{45.5} & \multirow{1}{*}{75.5} & \multirow{1}{*}{52.0} & \multirow{1}{*}{12.8}   \\

\multirow{1}{*}{CMT \citep{yan2023cross} (ICCV'23)}  
& C+L & \multirow{1}{*}{44.4} & \multirow{1}{*}{79.9} & \multirow{1}{*}{53.0} & \multirow{1}{*}{4.8}   \\

\multirow{1}{*}{EA-LSS \citep{hu2023ea} (arXiv'23)}  
& C+L & \multirow{1}{*}{44.6} & \multirow{1}{*}{79.5} & \multirow{1}{*}{52.6} & \multirow{1}{*}{6.2}   \\

\multirow{1}{*}{IS-Fusion \citep{yin2024fusion} (CVPR'24)}  
& C+L & \multirow{1}{*}{44.1} & \multirow{1}{*}{77.9} & \multirow{1}{*}{59.4} & \multirow{1}{*}{3.8}   \\

\multirow{1}{*}{CCF \citep{wu2026ccf} (CVPR'26)}  
& C+L & \multirow{1}{*}{47.1} & \multirow{1}{*}{79.8} & \multirow{1}{*}{56.2} & \multirow{1}{*}{9.2}   \\

\multirow{1}{*}{FOMO-3D \citep{yang2026fomo} (CoRL'25)}  
& C+L & \multirow{1}{*}{54.6} & \multirow{1}{*}{79.9} & \multirow{1}{*}{59.6} & \multirow{1}{*}{27.6}   \\

\midrule

\multirow{1}{*}{\textbf{MMF}(CenterPoint, FCOS3D)
}   
& \multirow{1}{*}{C+L} & \multirow{1}{*}{43.6} & \multirow{1}{*}{77.1} & \multirow{1}{*}{49.0} & \multirow{1}{*}{9.4}   \\

\multirow{1}{*}{\textbf{MMLF}(CenterPoint, DINO \citep{zhang2022dino})
}
& \multirow{1}{*}{C+L} & \multirow{1}{*}{53.1}  & \multirow{1}{*}{78.6} & \multirow{1}{*}{60.3} & \multirow{1}{*}{23.3}  \\

\multirow{1}{*}{\textbf{MMLF}(CenterPoint, GroundingDINO \citep{liu2024grounding})} & 
\multirow{1}{*}{C+L} &
55.2 &
79.9 &
62.5 &
26.0
\\

\multirow{1}{*}{\textbf{MMLF}(CenterPoint, Co-DINO \citep{zong2023detrs})}
& \multirow{1}{*}{C+L} & \multirow{1}{*}{\textbf{57.0}}  & \multirow{1}{*}{\textbf{80.7}} & \multirow{1}{*}{\textbf{64.0}} & \multirow{1}{*}{\textbf{29.2}}  \\

\rowcolor{lightgray}\multirow{1}{*}{\textbf{MMLF}(CenterPoint, ground-truth bbox)}
& \multirow{1}{*}{C+L} & \multirow{1}{*}{77.1}  & \multirow{1}{*}{85.9} & \multirow{1}{*}{73.9} & \multirow{1}{*}{73.6}  \\
\bottomrule
\end{tabular}
}
\caption{\small
{\bf Benchmarking results on nuScenes}.
We denote the RGB and LiDAR modalities by C and L, respectively.
Our simple late-fusion method MMF, 
which uses detections of the 3D RGB-based detector FCOS3D to filter detections of the 3D LiDAR-based detector CenterPoint (improved by using our grouping-free head and training by our hierarchy loss),
rivals most RGB-only, LiDAR-only and multimodal 3D detectors.
Furthermore, our final method MMLF,
which probabilistically fuses 3D LiDAR and 2D RGB detections,
performs the best among non-oracle methods.
}
\label{tab:quick_benchmarking_results} 
\vspace{-4mm}
\end{table*}
}

\section{Experiments}
\label{sec:experiments}
We conduct extensive experiments to better understand the LT3D  problem and gain insights by validating our techniques.
The experiments will answer the following questions:\footnote{Answers: yes, yes, yes, yes, yes.}
\begin{enumerate}[leftmargin=0.5cm, itemsep=0pt, topsep=0.1cm]
\item 
Are {\tt rare} classes more difficult to detect than {\tt common} classes?
\item 
Are objects from {\tt rare} classes sufficiently localized but mis-classified?
\item 
Does training with the semantic hierarchy improve detection performance for LT3D?
\item 
Does our multimodal late-fusion strategy resoundingly outperform existing detectors?
\item 
Does our multi-modal late-fusion help detect {\tt rare} classes?
\end{enumerate}

We compare our MMF and  MMLF approaches (Figs.~\ref{fig:filtering} and \ref{fig:pipeline_diag}) with prior works and present a detailed ablation study to further addresses the three motivating questions in Fig.~\ref{fig:three-questions}. 
Regarding notation,
MMF(X, Y) and MMLF(X, Y) mean that they fuse detections from the LiDAR-based detector X and the RGB-based detector Y.
We provide implementation details in Appendix \ref{sec:impl_details}.
We release our code and models on  \href{https://github.com/cc50121/lt3d-lf}{Github}.

{\bf Datasets}. 
We use the well-established nuScenes~\citep{caesar2020nuscenes} and Argoverse 2.0 (AV2)~\citep{wilson2021argoverse} datasets.
Other AV datasets such as KITTI~\citep{geiger2012we} and Waymo~\citep{sun2020waymo} do not support the study of LT3D as they annotate only three common classes.
In contrast,
nuScenes and AV2 have 18 and 26 fine-grained classes, respectively, which follow long-tailed distributions.
To quantify the long-tail, we compute the imbalance factor (IF), defined as the ratio between the numbers of annotations of the most and the least common classes~\citep{cao2019learning}: nuScenes and AV2 have an IF=1,670 and 2,500 respectively, which are significantly more imbalanced than existing long-tail image recognition benchmarks, e.g., iNaturalist (IF=500)~\citep{van2018inaturalist} and ImageNet-LT (IF=1,000)~\citep{liu2019large}. 
nuScenes arranges classes in a semantic hierarchy (Fig.~\ref{fig:hierarchy}); AV2 does not provide a semantic hierarchy but we construct one based on the nuScenes' hierarchy (refer to Appendix \ref{sec:av2} for details).
Following prior work, we use the official training set for training and the validation set for benchmarking.
As our primary conclusions hold for both datasets,
we report results on nuScenes in the main paper and provide results on AV2 in Appendix \ref{sec:av2}.

\begin{figure*}[t]
\centering
\includegraphics[width=\linewidth]{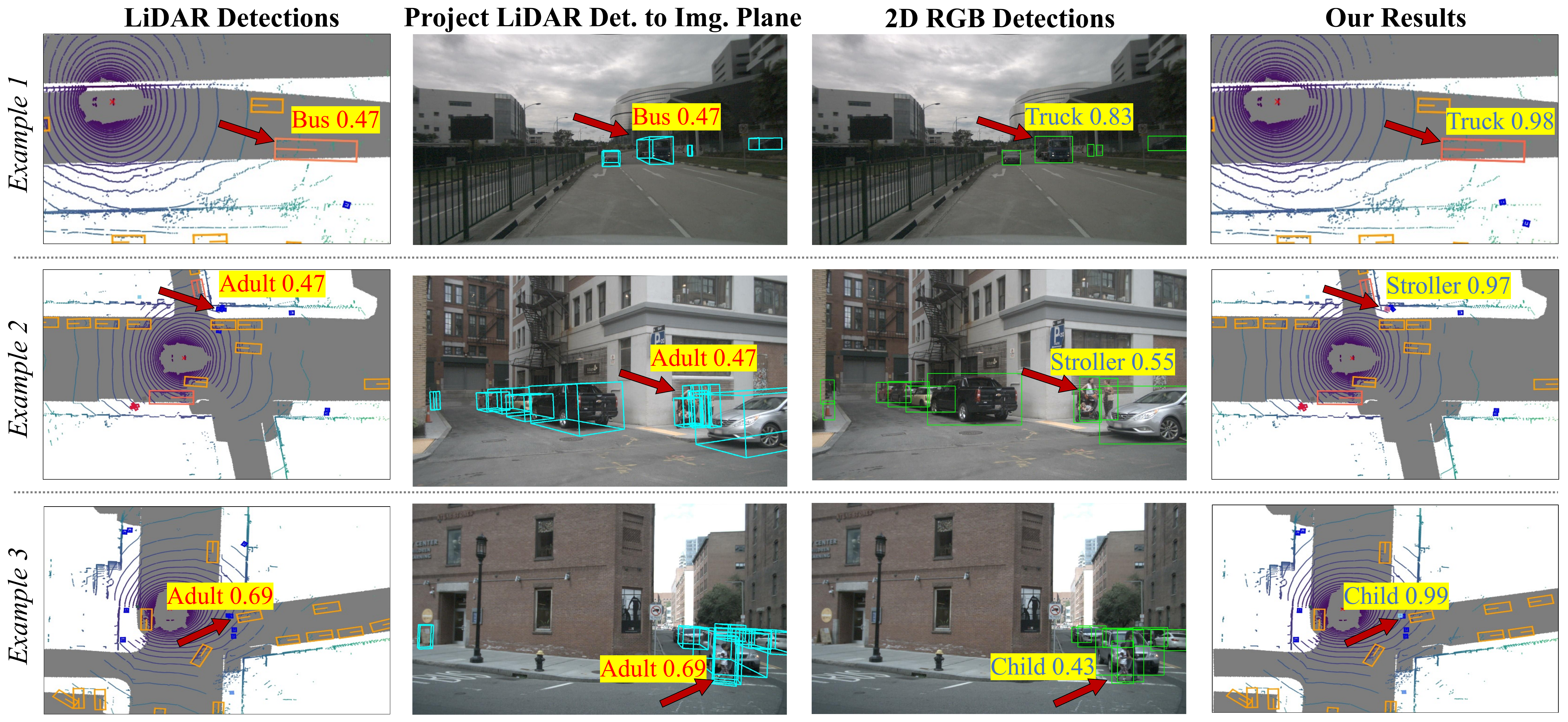}
\vspace{-5mm}
\caption{\small
We demonstrate how our multi-modal late-fusion (MMLF) approach improves LT3D by fusing 2D RGB detections (from Co-DINO) and 3D LiDAR detections (from CenterPoint). 
In the three examples, MMLF correctly relabels detections which are geometrically similar (w.r.t size and shape) in LiDAR but are visually distinct in RGB, such as {\tt bus}-vs-{\tt truck}, {\tt adult}-vs-{\tt stroller}, and {\tt adult}-vs-{\tt child}. 
Refer to Appendix \ref{sec:more-visualizations} for more visualizations. 
}
\label{fig:visual-results}
\end{figure*}

\subsection{Benchmarking Results}
Table~\ref{tab:quick_benchmarking_results} compares our MMF and MMLF  with prior work on nuScenes.
We adapt existing methods (which were previously trained on 10 classes in nuScenes) for LT3D by retraining them on all 18 classes. We provide breakdown analysis per class in Appendix \ref{sec:per-class-results} and show results on the standard 10-class benchmark in Appendix \ref{sec:standard}.

All the compared RGB-based 3D detectors not only struggle in detecting rare classes but also yield significantly lower mAP than LiDAR-based and multimodal 3D detectors.
Moreover, 
compared with the recent LT3D detector FOMO-3D \citep{yang2026fomo},
existing multimodal 3D detectors, such as the recent CCF \citep{wu2026ccf}, are not competitive in LT3D.

Next, we compare another simple baseline,
termed ``CenterPoint + RCNN'', which trains a region-based CNN (RCNN) \citep{Girshick_2014_CVPR} classifier on cropped regions corresponding to projected 3D detections.
Notably, it underperforms CenterPoint by 5.2 mAP,
likely because the cropped-region classifiers do not exploit contextual information useful for recognition.
This suggests that late-fusion cannot be simply solved with a 3D LiDAR detector and a strong 2D classifier. 
MMF keeps CenterPoint detections that are close to monocular 3D RGB detections produced by FCOS3D in 3D and discards all other LiDAR predictions. 
Our MMF achieves 9.4 mAP on classes with {\tt few} training examples.
We also re-implement the learning-based late-fusion method CLOCS \citep{pang2009clocs} that fuses detections from either DINO + CenterPoint, or Co-DINO + CenterPoint.
CLOCS achieves better results by fusing stronger 2D detectors (Co-DINO vs. DINO),
reaffirming the benefit of late fusion's modular design: incorporating stronger uni-modal detectors yields better 3D detection performance.
CLOCS only fuses predictions from the same predicted class, which prevents re-labeling of misclassified LiDAR detections.

\begin{table}[t]
    \centering
    \setlength{\tabcolsep}{0.5mm} 
\resizebox{1.0\linewidth}{!}{%
    \begin{tabular}{l c c c c c c c c c c c}
    \toprule
    \multirow{1}{*}{Method} & \multirow{1}{*}{Modality} & \multicolumn{1}{c}{\tt All} & \multicolumn{1}{c}{\texttt{Many}} & \multicolumn{1}{c}{\texttt{Medium}} & \multicolumn{1}{c}{\tt Few} \\ 
    
    \midrule
    \multirow{1}{*}{FCOS3D}        &     C      &   20.9 &  39.0            & 23.3         &   2.9               \\ 
    \midrule
    \multirow{1}{*}{PointPillars (PP)}                      &      L      &   30.0  &  64.2            & 28.4           &   3.4             \\ 
    \multirow{1}{*}{\quad + Hierarchy Loss}                                         &        L     &   31.2  &  \textbf{66.4}   & 30.4           &   2.9               \\
    \multirow{1}{*}{\quad + MMF(PP, FCOS3D)}                                        & C+L  &   \textbf{35.8} &  66.2            & \textbf{41.0}  &   \textbf{4.4}              \\
    
    \midrule
    \multirow{1}{*}{CenterPoint (CP)}                       &     L       &   39.2 &  76.4             & 43.1         &   3.5                \\ 
    \multirow{1}{*}{\quad + Hierarchy Loss}          &        L                             &   40.4 &  \textbf{77.1}             & 45.1         &   4.3              \\
    \multirow{1}{*}{\quad + MMF(CP, FCOS3D)}         &   C+L                       &   \textbf{43.6}  &  \textbf{77.1}             & \textbf{49.0}          &   \textbf{9.4}            \\ 
    
    \bottomrule
    \end{tabular}
    }
    \caption{\small  
    Training with the semantic hierarchy improves both PointPillars \citep{lang2019pointpillars}  and CenterPoint \citep{yin2021center} for LT3D by $>$1 mAP over {\tt All} classes. 
    Further, MMF improves on  {\tt Medium} and {\tt Few} classes by 1$\sim$13 mAP, showing the benefit of late fusion for LT3D.
    }
    \label{tab:hierarchy}
\vspace{-4mm}
\end{table}

It is worth comparing the recent LT3D method FOMO-3D,
which trains a network to fuse RGB and LiDAR detections in the 3D space and exploits a vision foundation model Metric3D \citep{yin2023metric3d} for depth estimation over RGB frames.
Our method MMLF outperforms FOMO-3D by 2.4 mAP.
As our MMLF adopts a non-learned fusion method without exploiting estimated depth,
our better results corroborate the analysis in Section~\ref{ssec:det-match} that inflating 2D detections to 3D does not necessarily facilitate fusion.
Moreover, using estimated depth does not necessarily improve 3D detection performance.

Lastly,
we note that our MMLF is modular in design that can adopt any 2D RGB detectors and 3D LiDAR detectors.
In the last row of Table~\ref{tab:quick_benchmarking_results},
we provide the results by using ground-truth 2D boxes to fuse 3D LiDAR detections.
These results demonstrate that our MMLF's performance can be significantly improved by using stronger 2D RGB detectors.
Indeed, by using stronger 2D detectors such as GroundingDINO and Co-DINO,
MMLF achieves significantly better LT3D performance.
Fig.~\ref{fig:visual-results} displays qualitative results that demonstrate how MMLF corrects misclassifications on LiDAR detections and improves the final LT3D performance.

\subsection{Ablation Study and Analysis}
We conduct extensive ablations to validate our proposed MMLF approach. 
In appendices, we provide additional results on the impact of calibration noise (Appendix \ref{sec:analysis-calibration-noise}), detector recall (Appendix \ref{sec:analysis-lidar-recall}) and performance breakdowns by object visibility and distance from ego-vehicle (Appendix \ref{sec:analysis-occlusion}).

\subsubsection{Analysis on Training with Semantic Hierarchy} 
We validate the benefits of training with semantic hierarchy by modifying LiDAR-based detectors to jointly predict class labels at different levels of the  hierarchy.
For example, we modify detectors PointPillars~\citep{lang2019pointpillars} and CenterPoint~\citep{yin2021center} to additionally classify {\tt stroller} as {\tt pedestrian} and {\tt object}. 
As the hierarchy naturally groups classes based on shared attributes, training with it helps learn complementary features across levels that are shared 
across {\tt common} and {\tt rare} classes.
Table \ref{tab:hierarchy} highlights that this approach (``+ Hierarchy Loss'') achieves improvements on all classes by $>$1 mAP.

{
\setlength{\tabcolsep}{0.83mm}
\begin{table}[t]
\centering
\resizebox{1\linewidth}{!}{%
\begin{tabular}{l | cccc | cccc}
\hline
\multirow{2}{*}{\quad Method}            & \multicolumn{4}{c|}{Fusion in 3D} &\multicolumn{4}{c}{Fusion in 2D}  \\
      & {\tt All}  & {\tt Many}   & {\tt Medium}   & {\tt Few}   &  {\tt All} & {\tt Many}    & {\tt Medium}    & {\tt Few}     \\
\hline
\cellcolor{lightgrey}\ \ \ CenterPoint  &   \cellcolor{lightgrey}40.4  & \cellcolor{lightgrey}77.1      &   \cellcolor{lightgrey}45.1       &   \cellcolor{lightgrey}4.3       &   \cellcolor{lightgrey}40.4 &  \cellcolor{lightgrey}77.1      &   \cellcolor{lightgrey}45.1       &   \cellcolor{lightgrey}4.3      \\
\hline
\cellcolor{col11}+ FCOS3D    &  \cellcolor{col11}42.9  &   \cellcolor{col11}76.6     &    \cellcolor{col11}48.7      &   \cellcolor{col11}8.1       &   \cellcolor{col11}42.6  &  \cellcolor{col11}75.0      &    \cellcolor{col11}49.4       &   \cellcolor{col11}7.7       \\
\cellcolor{col11}+ BEVFormer   &  \cellcolor{col11}43.2 &   \cellcolor{col11}76.9     &    \cellcolor{col11}50.8      &  \cellcolor{col11}6.3     &   \cellcolor{col11}42.8  &   \cellcolor{col11}75.2      &    \cellcolor{col11}51.4      &   \cellcolor{col11}5.7     \\
\cellcolor{col11}+ PolarFormer  &  \cellcolor{col11}42.8  &  \cellcolor{col11}76.8      &   \cellcolor{col11}50.0       &  \cellcolor{col11}6.1     &   \cellcolor{col11}42.6 &    \cellcolor{col11}75.1     &    \cellcolor{col11}51.1      &   \cellcolor{col11}5.6     \\

\hline

\cellcolor{col3}+ YOLOV7       &  \cellcolor{col3}40.1 &  \cellcolor{col3}76.1      &   \cellcolor{col3}43.8       &   \cellcolor{col3}5.8        &  \cellcolor{col3}48.6 &  \cellcolor{col3}76.7       &   \cellcolor{col3}57.1        &  \cellcolor{col3}15.3      \\
\cellcolor{col3}+ DINO        &  \cellcolor{col3}40.3  &  \cellcolor{col3}76.2      &  \cellcolor{col3}44.1        &  \cellcolor{col3}5.9          &  \cellcolor{col3}53.1 &   \cellcolor{col3}78.6      &    \cellcolor{col3}60.3       &   \cellcolor{col3}22.3       \\
\cellcolor{col3}+ GroundingDINO       &  \cellcolor{col3}42.3 &  \cellcolor{col3}78.1      &   \cellcolor{col3}45.6       &   \cellcolor{col3}8.6        &  \cellcolor{col3}55.2 &  \cellcolor{col3}79.9       &   \cellcolor{col3}62.5        &  \cellcolor{col3}26.0     \\
\cellcolor{col3}+ Co-DINO        &  \cellcolor{col3}42.9  &  \cellcolor{col3}79.0      &  \cellcolor{col3}45.9        &  \cellcolor{col3}9.3          &  \cellcolor{col3}57.0 &   \cellcolor{col3}80.7      &    \cellcolor{col3}64.0       &   \cellcolor{col3}29.2       \\
\hline
\end{tabular}
}
\caption{\small
{\bf Fusing uni-modal detections in 3D vs. on the 2D image plane}. We evaluate the impact of fusing 3D LiDAR detections ({from \setlength{\fboxsep}{1pt}\colorbox{lightgrey}{CenterPoint}} trained with our hierarchy loss) with {\setlength{\fboxsep}{1pt}\colorbox{col11}{2D RGB}} \emph{vs.} and {\setlength{\fboxsep}{1pt}\colorbox{col3}{3D RGB}} detections, in 3D \emph{vs.} on the image plane.
We match and filter detections in 3D using center distance (MMF),
and match and filter detections in the 2D image plane using IoU (MMLF).
When fusing in 3D,
we follow \citep{wilson20203d, khurana2024shelf} that inflate 2D detections to 3D using LiDAR points within their box frustums. 
When fusing on the 2D image plane, we project 3D detections to the image plane using provided sensor extrinsics.
When fusing 3D RGB detections, similar results are achieved regardless of whether matching is performed in 3D or on the 2D image plane.
Unsurprisingly, inflating 2D RGB detections for matching in 3D performs worse than matching 3D RGB detections in 3D.
Yet, late-fusion between LiDAR-based 3D detections and RGB-based 2D detections in the image plane (bottom right panel) significantly improves performance for classes with {\tt medium} and {\tt few} examples by $>$10 mAP.
Refer to Appendix \ref{sec:analysis-2D-detector-perf} for more results.
}
\label{tab:analysis_2D-detectors}
\vspace{-5mm}
\end{table}
}

{
\setlength{\tabcolsep}{1mm}
\begin{table}[t]
\small
\centering
\vspace{-0mm}
\resizebox{1.0\linewidth}{!}{%
\begin{tabular}{lrrrrrrr}
\toprule 
\multirow{1}{*}{Method} & \multicolumn{1}{c}{$\Delta$} & \multicolumn{1}{c}{\tt All} & \multicolumn{1}{c}{{\tt Many}} & \multicolumn{1}{c}{{\tt Medium}} & \multicolumn{1}{c}{{\tt Few}} 
 \\
\midrule
 \multirow{1}{*}{CenterPoint}                        &   & 39.2  & 76.4             & {43.1}          &   3.5             \\ 
 \multirow{1}{*}{\quad + Hierarchy Loss}                        & \darkgreen{+1.2}  & 40.4  & 77.1             & {45.1}          &   4.3             \\ 
\multirow{1}{*}{\quad + MMLF w/ Co-DINO}   
 & \darkgreen{+7.7} & \multirow{1}{*}{48.1} & \multirow{1}{*}{78.8} & \multirow{1}{*}{52.9} & \multirow{1}{*}{16.9}  \\

\multirow{1}{*}{\quad + External Data}   
 & \darkgreen{+4.5} & \multirow{1}{*}{52.6} & \multirow{1}{*}{79.8} & \multirow{1}{*}{58.1} & \multirow{1}{*}{23.5}  \\
 
\multirow{1}{*}{\quad + Score Calibration}   
 & \darkgreen{+2.3} & \multirow{1}{*}{54.9} & \multirow{1}{*}{80.2} & \multirow{1}{*}{60.7} & \multirow{1}{*}{27.1}  \\

 \multirow{1}{*}{\quad + Probabilistic Fusion}   
& \darkgreen{+2.1} & \multirow{1}{*}{\textbf{57.0}} & \multirow{1}{*}{\textbf{80.7}} & \multirow{1}{*}{\textbf{64.0}} & \multirow{1}{*}{\textbf{29.2}}  \\
 \bottomrule
\end{tabular}%
}
\caption{\small
Adding our developed components progressively improves LT3D performance, confirming that 
(1) 2D RGB detectors are better suited for our MMLF,
(2) matching projected 3D LiDAR detections on the 2D image plane outperforms matching 2D RGB detections inflated to 3D,
and (3) score calibration prior and probabilistic fusion improve performance.
A majority of performance improvements is  attributed to fusing detections of the 2D RGB detector Co-DINO on the 2D image plane.
Moreover, training with external data notably improves {\tt rare} class accuracy by 6.6 mAP. 
}
\label{tab:analysis-fusion}
\vspace{-6mm}
\end{table}
}

\subsubsection{Analysis on Multi-Modal Late-Fusion}
We study the trade-off between using 2D and 3D RGB detectors, and matching in the 2D image plane and in 3D in Table~\ref{tab:analysis_2D-detectors}. 
We further examine the impact of using additional data and study different fusion strategies in Table~\ref{tab:analysis-fusion}.

{\bf How to incorporate RGB information?} 
Table~\ref{tab:hierarchy} (+ MMF) demonstrates that incorporating RGB information (by matching and filtering 3D LiDAR detections) using a 3D RGB detector greatly improves LT3D.
In  Table \ref{tab:analysis_2D-detectors},
we evaluate the impact of using 2D RGB-based detectors (e.g., YOLOV7, DINO, GroundingDINO, and Co-DINO) versus 3D RGB-based detectors (e.g., FCOS3D, BEVFormer, PolarFormer, and DenseBEV),
and matching LiDAR detections with 2D RGB detections in 3D versus on the 2D image plane. 
Results show that matching LiDAR detections with 2D RGB detections on the 2D image plane (bottom right) performs best.

{\bf How to match detections?}
MMF keeps LiDAR detections within a radius of $m$ meters for each RGB detection
and removes all the others that are not close to any RGB-based detections. 
This works well when using a 3D RGB detector, 
e.g., MMF with FCOS3D, BEVFormer, and PolarFormer improves over the LiDAR-only detector CenterPoint by 2 mAP, as shown in Table \ref{tab:analysis_2D-detectors}. 
Moreover, matching inflated 2D RGB detections in 3D, performs worse than matching 3D RGB detections, achieving marginally lower accuracy than the LiDAR-only baseline.

Notably, projecting LiDAR detections on the 2D image plane and fusing them with 2D RGB detections significantly improves performance for classes with {\tt medium} and {\tt few} examples by $>$10 mAP. In contrast, projecting 3D RGB detections for matching on the 2D image plane performs worse than matching 2D RGB detections on the 2D image plane, due to erroneous 3D predictions that affect 2D detection quality.

{\bf How to fuse matched detections?}
Prior to fusion, we first calibrate the scores of LiDAR and RGB detections to ensure that they are comparable. This improves accuracy by 2.3 mAP averaged over all classes (Table~\ref{tab:analysis-fusion}). 
Moreover, our probabilistic fusion 
achieves 2.1 mAP improvements.

\subsubsection{Analysis of Using Strong 2D Detectors} 
\begin{figure}[t]
    \centering
    \includegraphics[width=\linewidth]{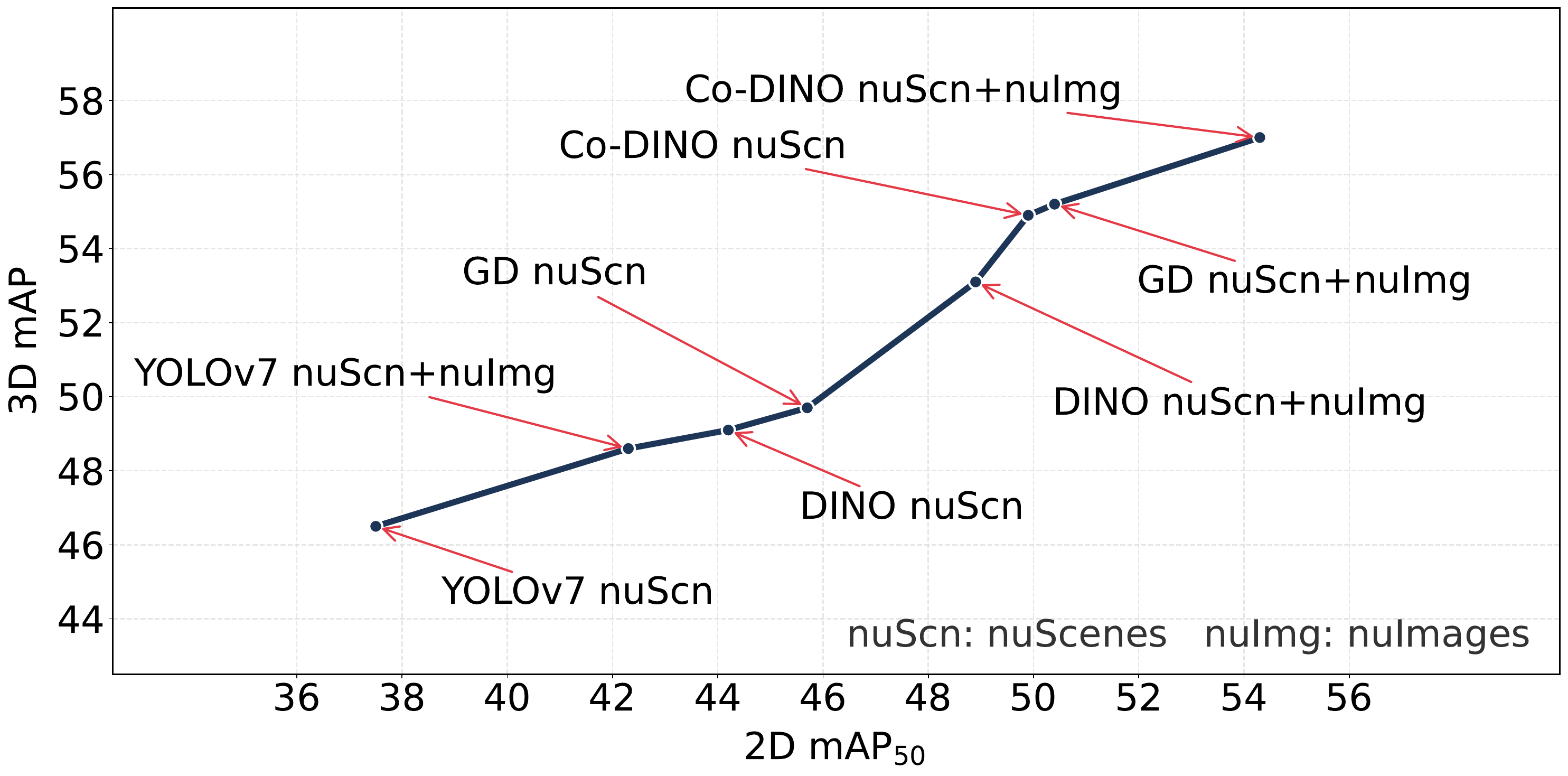}
    \vspace{-2mm}
    \caption{\small Training 2D detectors with more data (e.g., nuScenes + nuImages vs. nuScenes only) and using better 2D detectors (e.g., Co-DINO, GroundingDINO, DINO, compared with YOLOV7) improve performance not only on 2D detection but also on the 3D detection achieved by our MMLF. 
    }
    \vspace{-2mm}
    \label{fig:curves2}
\end{figure}

Fig. \ref{fig:curves2} shows a strong correlation between 2D detection accuracy and the final 3D LT3D performance by MMLF.
Owing to the modular design of MMLF,
simply training better 2D RGB detectors with more data provides a natural pathway for improving LT3D performance. 
This is an exciting signal, because, as more and more powerful foundational 2D detectors become publicly available, our MMLF framework can utilize them  to further improve LT3D performance.

{
\setlength{\tabcolsep}{1mm}
\begin{table}[t] 
\small
\centering
\scalebox{0.75}{
\begin{tabular}{lcrrrr}
\toprule 
\multirow{1}{*}{Method}     & \multicolumn{1}{c}{Modality} & \multicolumn{1}{c}{Time} & \multicolumn{1}{c}{Memory} & \multicolumn{1}{c}{mAP}  \\ 
\midrule

\multirow{1}{*}{FCOS3D \citep{wang2021fcos3d}}   
 & \multirow{1}{*}{C} & \multirow{1}{*}{89} & \multirow{1}{*}{4.2 GB} & \multirow{1}{*}{20.9}   \\

\multirow{1}{*}{BEVFormer \citep{li2022bevformer}}   
 & \multirow{1}{*}{C} & \multirow{1}{*}{327} & \multirow{1}{*}{6.4 GB} & \multirow{1}{*}{27.3}   \\

\midrule

\multirow{1}{*}{CenterPoint \citep{yin2021center}}   
& \multirow{1}{*}{L} & \multirow{1}{*}{323} & \multirow{1}{*}{4.0 GB} & \multirow{1}{*}{40.4}   \\
\multirow{1}{*}{FSHNet \citep{liu2025fshnet}}   
& \multirow{1}{*}{L} & \multirow{1}{*}{296} & \multirow{1}{*}{4.8 GB} & \multirow{1}{*}{41.6}   \\
\midrule

\multirow{1}{*}{CMT \citep{yan2023cross}}  
& L+C & \multirow{1}{*}{520} & \multirow{1}{*}{9.0 GB} & \multirow{1}{*}{44.4}   \\

\multirow{1}{*}{TransFusion \citep{bai2022transfusion}}   
& L+C & \multirow{1}{*}{367} & \multirow{1}{*}{5.6 GB} & \multirow{1}{*}{39.8}   \\

\multirow{1}{*}{DeepInteraction \citep{yang2022deepinteraction}}   
& L+C & \multirow{1}{*}{590} & \multirow{1}{*}{8.8 GB} & \multirow{1}{*}{43.7}   \\

\multirow{1}{*}{CCF \citep{wu2026ccf}}   
& L+C & \multirow{1}{*}{632} & \multirow{1}{*}{10.2 GB} & \multirow{1}{*}{47.1}   \\

\midrule

\multirow{1}{*}{\textbf{MMLF} (CP+DINO) Sequential}
& \multirow{1}{*}{L+C} & \multirow{1}{*}{603}  & \multirow{1}{*}{7.0 GB} & \multirow{1}{*}{53.1}  \\

\multirow{1}{*}{\textbf{MMLF} (CP+DINO) Parallel}
& \multirow{1}{*}{L+C} & \multirow{1}{*}{323}  & \multirow{1}{*}{11.0 GB} & \multirow{1}{*}{53.1}  \\

\multirow{1}{*}{\textbf{MMLF} (Co-DINO) Sequential}
& \multirow{1}{*}{L+C} & \multirow{1}{*}{683}  & \multirow{1}{*}{8.5 GB} & \multirow{1}{*}{57.0}  \\

\multirow{1}{*}{\textbf{MMLF} (Co-DINO) Parallel}
& \multirow{1}{*}{L+C} & \multirow{1}{*}{360}  & \multirow{1}{*}{12.5 GB} & \multirow{1}{*}{57.0}  \\

\bottomrule
\end{tabular}
}
\caption{\small
{\bf Comparison of inference efficiency} with respect to  wall-clock time (ms) and GPU consumption. 
We use NVIDIA A6000 GPUs to measure the computation cost.
Compared with other multimodal 3D detectors, 
MMLF is not only competitive in inference efficiency,
but also significantly better in performance (mAP).
}
\vspace{-3mm}
\label{tab:inference-efficiency} 
\end{table}
}

\begin{table*}[t]
\centering
\resizebox{\textwidth}{!}{
\begin{tabular}{lrrrrrrrrrrrrrrrr}
\toprule
\multirow{2}{*}{Methods}
& \multicolumn{4}{c}{Truck}
& \multicolumn{4}{c}{Adult}
& \multicolumn{4}{c}{Bicycle}
& \multicolumn{4}{c}{Traffic cone} \\
\cmidrule(lr){2-5}\cmidrule(lr){6-9}\cmidrule(lr){10-13}\cmidrule(lr){14-17}
& near & far & sparse & dense
& near & far & sparse & dense
& near & far & sparse & dense
& near & far & sparse & dense \\
\midrule
CenterPoint
& 65.6 & 35.6 & 30.1 & 71.8
& 87.4 & 57.8 & 74.5 & 89.6
& 54.7 &  3.4 & 16.4 & 67.6
& 75.7 &  N/A & 56.3 & 83.7 \\

FSHNet
& 68.5 & 38.5 & 34.0 & 74.4
& 88.3 & 59.3 & 76.1 & 90.8
& 62.1 &  7.8 & 24.3 & 72.1
& 78.7 &  N/A & 61.1 & 85.1 \\

CMT
& 65.8 & 42.5 & 39.1 & 69.9
& 88.6 & 59.7 & 76.1 & 91.3
& 70.0 & 19.6 & 43.2 & 76.7
& 80.1 &  N/A & 63.8 & 84.8 \\

CCF
& 67.2 & 42.9 & 39.5 & 71.1
& 88.8 & 60.8 & 77.1 & 91.7
& 72.2 & 19.7 & 44.7 & 78.2
& 80.8 &  N/A & 64.8 & 85.8 \\

\textbf{MMLF}
& \textbf{71.2} & \textbf{45.2} & \textbf{40.1} & \textbf{76.1}
& \textbf{90.2} & \textbf{63.9} & \textbf{80.1} & \textbf{93.3}
& \textbf{81.3} & \textbf{26.5} & \textbf{58.2} & \textbf{88.4}
& \textbf{82.3} & N/A & \textbf{66.8} & \textbf{86.7} \\

\midrule
\midrule

\multirow{2}{*}{Methods}
& \multicolumn{4}{c}{\color{blue}Personal mobility}
& \multicolumn{4}{c}{\color{blue}Child}
& \multicolumn{4}{c}{\color{blue}Police officer}
& \multicolumn{4}{c}{\color{blue}Stroller} \\
\cmidrule(lr){2-5}\cmidrule(lr){6-9}\cmidrule(lr){10-13}\cmidrule(lr){14-17}
& near & far & sparse & dense
& near & far & sparse & dense
& near & far & sparse & dense
& near & far & sparse & dense \\
\midrule
CenterPoint
& 0.5 & 0.0 & 0.3 & 0.4
&  1.2 & 0.0 &  0.3 &  1.5
&  0.3 & 0.0 & 0.0 &  0.4
&  4.3 & 1.3 &  2.3 &  4.2 \\

FSHNet
& 0.6 & 0.0 & 0.3 & 0.5
&  1.4 & 0.0 & 0.4 & 1.8
&  0.5 & 0.0 & 0.0 & 0.8
&  3.4 & 1.1 & 1.9 & 3.1 \\

CMT
&  4.8 & 0.0 & 3.7 &  1.4
&  5.3 & 0.2 &  0.8 &  8.2
&  8.8 & 0.0 & 0.0 & 15.4
&  0.8 & 0.3 &  0.4 &  0.6 \\

CCF
& 6.1 & 0.0 & 3.9 & 2.1
&  8.1 & 0.4 & 2.1 & 10.2
& 10.1 & 0.0 & 0.0 & 18.3
&  8.1 & 1.9 & 3.7 & 9.9 \\

\textbf{MMLF}
& \textbf{48.9} & 0.0 & \textbf{41.3} & \textbf{74.0}
& \textbf{37.0} & \textbf{1.5} & \textbf{14.0} & \textbf{43.9}
& \textbf{18.7} & 0.0 & 0.0 & \textbf{31.9}
& \textbf{42.0} & \textbf{40.5} & \textbf{32.7} & \textbf{53.4} \\
\bottomrule
\end{tabular}
}
\caption{\small
Breakdown results on objects from common (in top panel) and rare (in bottom panel) classes,
with respect to distance and LiDAR density.
We define far-field and near-field objects when they are $>$30m and $<$30m away from the ego vehicle, respectively.
To define dense or sparse LiDAR returns, 
we consider a per-class threshold to evenly divide the test instances of a class.
We provide the results for all the classes in Table~\ref{tab:multi_class_results} of the Appendix.
The results demonstrate that our method boosts performance across most common and rare classes, both far-field and near-field objects, and both objects with dense and sparse LiDAR returns.
}
\label{tab:selected_stratified_results}
\end{table*}

\subsubsection{Analysis of Inference Efficiency} 

We analyze the inference efficiency of our MMLF.
As MMLF can run the two uni-modal detectors in either  parallel mode or sequential mode, we analyze both modes.
Moreover, because different uni-modal detectors can affect the overall computation of MMLF, we compare DINO and Co-DINO as the RGB 2D detector in MMLF.
Table~\ref{tab:inference-efficiency} lists the computation costs in terms of wall-clock time (in ms) and GPU consumption.
We see a trade-off between the two modes.
Moreover,
comparing with other multimodal 3D detectors,
we conclude that MMLF is not only competitive in inference efficiency but also substantially stronger in LT3D performance (mAP).
Because existing methods are not designed for production deployment or real-time operation, optimizing them to meet real-time requirements is an indispensable engineering step toward practical deployment in real-world applications.

\subsubsection{Performance on Dense/Sparse and Near/Far Objects} 

While objects of rare classes make training difficult,
difficulties can arise from objects that are in the far field \citep{gupta2023far3det} or have sparse LiDAR returns.
We show that our MMLF also improves detection of these objects compared with 
recent 3D detectors, 
including CenterPoint,
CMT \citep{yan2023cross}, FSHNet \citep{liu2025fshnet}, and CCF \citep{wu2026ccf}.
Table~\ref{tab:selected_stratified_results} lists the breakdown results on four common and four rare classes.
We provide per-class results in Table~\ref{tab:multi_class_results} of the Appendix.
The results show that MMLF generally improves performance across  common and rare classes, on objects with varied  distances and LiDAR densities.

\subsubsection{Analysis of Misclassification}
We evaluate CenterPoint~\citep{yin2021center} and CMT~\citep{yan2023cross} using the hierarchical mAP metric (mAP$_H$), which considers semantic relationships across classes to award partial credit.
In safety-critical applications, detecting but misclassifying objects (as a semantically related category) is more desirable than a missed detection (e.g., detecting but misclassifying a {\tt child} as an {\tt adult} is preferable to not detecting this {\tt child}). 
Table~\ref{tab:ap_h} highlights the results of applying our proposed techniques including the grouping-free detector head, the hierarchy loss, and MMLF using the 2D detector Co-DINO.
Results reveal that (1) classes are most often misclassified as their LCA=1 siblings within coarse-grained superclasses,
(2) our developed techniques significantly improve mAP$_H$ (especially on {\tt rare} classes), reducing misclassification between sibling classes.

{
\setlength{\tabcolsep}{1.7mm} 
\begin{table*}[t]
\small
\centering
\resizebox{1.0\linewidth}{!}{%
\begin{tabular}{l l c c c c c c c c c c c c c}
\toprule
  \multirow{1}{*}{Method}  &\multirow{1}{*}{$mAP_H$}  & \multicolumn{1}{c}{Car} & \multicolumn{1}{c}{Adult} & \multicolumn{1}{c}{Truck} & \multicolumn{1}{c}{\color{blue}CV} &  \multicolumn{1}{c}{\color{blue}Bicycle} & \multicolumn{1}{c}{\color{blue}MC} &  \multicolumn{1}{c}{\color{blue}Child} & \multicolumn{1}{c}{\color{blue}CW} & \multicolumn{1}{c}{\color{blue}Stroller} & \multicolumn{1}{c}{\color{blue}PP} \\ 

\midrule

\multirow{2}{*}{CenterPoint~\citep{yin2021center}}    & \multirow{1}{*}{LCA=0}    &     82.4    &   81.2         &    49.4      &      19.7             &   33.6         &    48.9          &      0.1             &   14.2        &    0.1        &      21.7     \\
\multirow{2}{*}{(original)}  & \multirow{1}{*}{LCA=1}                                  &     83.9    &   82.0         &    58.7      &      20.5             &   35.2         &    50.5          &      0.1             &   18.3        &    0.1        &      22.0     \\
    & \multirow{1}{*}{LCA=2}                                &     84.0    &   82.4         &    58.8      &      20.7             &   36.4         &    51.0          &      0.1             &   19.5        &    0.1        &      22.6     \\
\midrule
\multirow{3}{*}{CenterPoint (w/ our grouping-free head)}    & \multirow{1}{*}{LCA=0}        &      88.1      &   86.3         &    62.7           &      24.5      &   48.5        &    62.8          &      0.1        &   22.2         &    4.3         &      32.7    \\
& \multirow{1}{*}{LCA=1}                                                   &      89.0      &   87.1         &    71.6           &      26.7      &   50.2        &    64.7          &      0.1        &   29.4         &    4.5         &      32.9    \\
    & \multirow{1}{*}{LCA=2}                                                 &      89.1      &   87.5         &    71.7           &      26.8      &   51.1        &    65.2          &      0.1        &   30.5         &    4.8         &      33.4    \\
\midrule
    \multirow{2}{*}{CenterPoint (w/ our grouping-free head)}  & \multirow{1}{*}{LCA=0}       &      88.6           &  86.9             &   63.4                  &      25.7          &     50.2       &    63.2       &      0.1           &   25.3         &    8.7     &    36.8    \\ 
    \multirow{2}{*}{\em + Hierarchy Loss}
    & \multirow{1}{*}{LCA=1}              &      89.5           &   87.6            &   72.4                  &      27.5          &     52.2       &    65.2       &      0.1           &   32.4         &    9.4     &    37.0    \\
     & \multirow{1}{*}{LCA=2}                                                 &      89.6           &   88.0            &   72.5                  &      27.7          &     53.2       &    65.7       &      0.1           &   34.0         &    9.8     &    37.6    \\
\midrule
   \multirow{2}{*}{CenterPoint (w/ our grouping-free head)}  & \multirow{1}{*}{LCA=0}                     &      88.5           &  89.2             &   68.1                  &      40.0          &     77.3       &     80.7      &      32.1           &   63.9          &   45.4        &  57.2   \\ 
\multirow{2}{*}{\em + Hier. + MMLF} 
& \multirow{1}{*}{LCA=1}          &      89.3           &  89.9             &   75.4                  &      41.9          &     77.8       &     81.5      &      44.7          &   73.5          &   48.3        &  57.3   \\ 
     & \multirow{1}{*}{LCA=2}                                                 &      89.3           &  90.1             &   75.5                  &      42.2          &     78.2       &     81.7      &      44.9          &   76.5          &   48.6        &  58.3   \\ 
\midrule
\midrule
\multirow{2}{*}{CMT \citep{yan2023cross}}    & \multirow{1}{*}{LCA=0}                              &       88.6          &    87.7           &     65.2                &     36.9           &      66.7      &     76.3      &      4.7           &    34.4         &    0.9       &   34.1    \\ 
\multirow{2}{*}{(RGB + LiDAR)}& \multirow{1}{*}{LCA=1}                        &       89.1          &    88.3           &     73.3                &     38.8           &      67.6      &      77.2     &      7.3           &    50.6         &    1.2       &   34.6    \\ 
    & \multirow{1}{*}{LCA=2}                                                  &       89.1          &    88.6           &     73.4                &     39.0           &      68.3      &     77.7      &      7.7           &    52.6         &    1.3       &   35.6     \\ 
\midrule
\multirow{2}{*}{CMT-L (LiDAR-only)}    & \multirow{1}{*}{LCA=0}                            &      88.4          &    88.6           &   67.6                  &    37.9            &     75.2       &    78.9       &     29.5           &    60.4         &    43.0      &  51.3     \\ 
\multirow{2}{*}{\em + MMLF}
& \multirow{1}{*}{LCA=1}                    &       88.9          &    89.0          &   73.2                  &    39.8            &     76.3       &    79.5       &     41.8          &   69.0         &    46.5       &  56.3    \\ 
    & \multirow{1}{*}{LCA=2}                                                  &       88.9          &    89.1           &   74.9                  &    40.3            &     78.1       &   79.7       &     42.4           &    74.7         &    47.6       &  57.2      \\ 
\bottomrule
\end{tabular}
}
\caption{\small 
{\bf Diagnosis using the mAP$_H$ metric on selected classes}.
We analyze the LiDAR-based 3D detector CenterPoint~\citep{yin2021center} and multi-modal detector CMT~\citep{yan2023cross}, with our proposed grouping-free detector head, hierarchy loss, and MMLF (using the 2D RGB detector Co-DINO \citep{zong2023detrs}).
Comparing the rows of LCA=0 for CenterPoint and CMT, we see that MMLF significantly improves on
{\tt rare} classes (in {\color{blue}blue}), e.g., construction-vehicle (CV), bicycle, motorcycle (MC),
construction-worker (CW), stroller, and pushable-pullable (PP).
Moreover, performance increases significantly from LCA=0 to LCA=1 compared against LCA=1 to LCA=2,
suggesting that objects from {\tt rare} classes are often detected but misclassified as sibling classes.
}
\label{tab:ap_h}
\end{table*}
}

\begin{figure*}[t]
\centering
\includegraphics[width=1\linewidth]{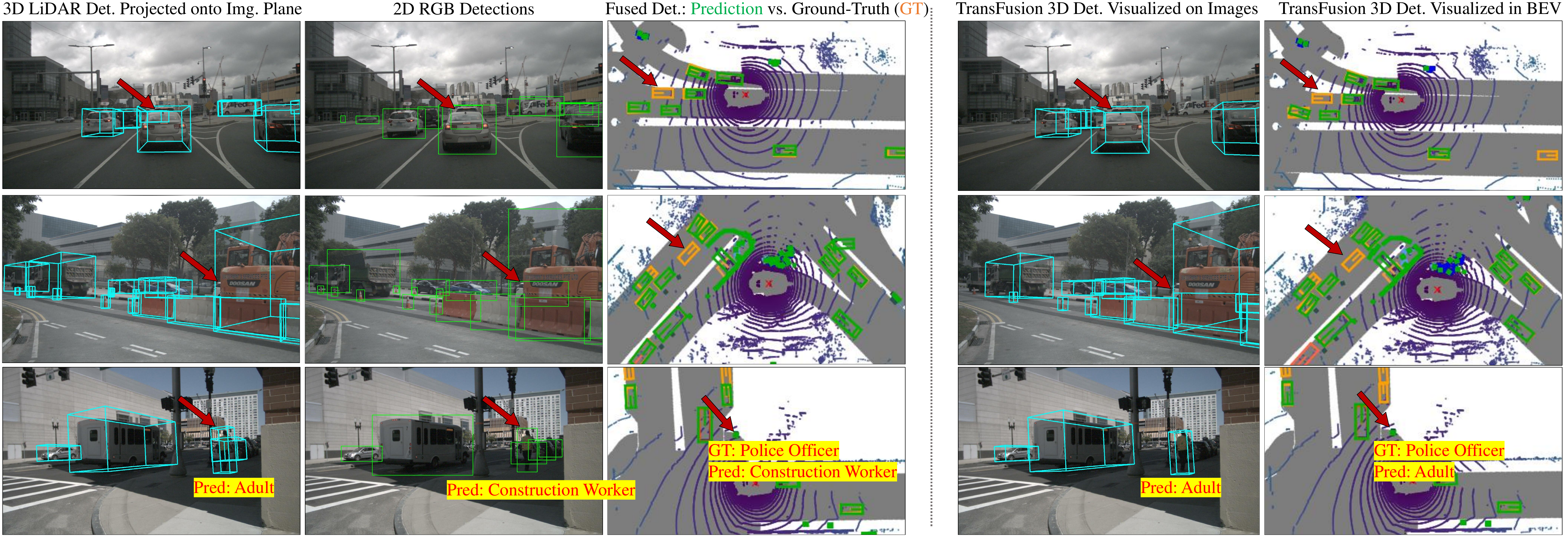}
\vspace{-4mm}
\caption{\small 
Both our MMLF (columns 1-3) and TransFusion~\citep{bai2022transfusion} (columns 4 -5) fail on these cases. In the first and second row, the 2D RGB detector Co-DINO detects the heavily occluded cars but 3D LiDAR detector fails to detect them. As a result, our MMLF misses these cars because it throws away unmatched RGB detections which do not have accurate 3D information. 
In the third row, we see that, although both the LiDAR and RGB detectors fire on the object (whose ground-truth label is {\tt police-officer}), the LiDAR detector classifies it as {\tt adult} and the RGB detector classifies it as {\tt construction-worker}. 
As a result,
the final label is incorrectly predicted.
TransFusion also misclassifies this object, confusing it with {\tt adult}. 
Refer to Appendix \ref{sec:more-visualizations} for more visualizations.
}
\label{fig:failure-cases}
\end{figure*}

\subsubsection{Analysis of Failure Cases} 
In Fig.~\ref{fig:failure-cases}, we visualize common failure cases of MMLF and compare them with the failure cases of TransFusion~\citep{bai2022transfusion}, an end-to-end trained multi-modal detector. 
Both TransFusion and our MMLF fail,
when there are heavy occlusions (i.e.,  3D information is missing). 
Another failure mode of MMLF is due to the misclassifications of the  2D RGB detector adopted by MMLF.
Yet, in such cases (ref. the third example in Fig.~\ref{fig:failure-cases}), TransFusion also fails.
This suggests that the objects misclassified by a strong 2D detector also challenge other 3D detectors.

\section{Conclusion and Future Work}
We explore the problem of long-tailed 3D detection (LT3D), detecting  objects not only from {\tt common} classes but also from many {\tt rare} classes.
This problem is motivated by the operational safety of autonomous vehicles (AVs) but has broad robotic applications, 
e.g., elder-assistive robots \citep{savage2022elderly} that fetch diverse items \citep{grauman2022ego4d} should address LT3D. To study LT3D, we establish a rigorous evaluation protocol that allows for partial credit to better diagnose 3D detectors.
We propose several algorithmic innovations to improve LT3D,
including a grouping-free detector head, a hierarchical loss that promotes feature sharing across long-tailed classes, and present a detailed exploration of multi-modal late-fusion (MMLF) for LT3D.
We find that 2D RGB detectors are better suited for late-fusion, matching projected 3D LiDAR detections on the 2D image plane outperforms matching 2D RGB detections inflated to 3D, and score calibration and probabilistic fusion notably improve performance.
Our MMLF framework not only achieves state-of-the-art performance on long-tailed 3D detection (LT3D) benchmarks, 
but also features a simple, modular design that can be seamlessly integrated with any 2D RGB detector and 3D LiDAR detector, 
making it readily extensible as these component models continue to improve.

Based on our contributions and insights,
below we discuss three directions for future work.

\textbf{Embracing the open world.}
Autonomous vehicles and more general robotic systems operating in the real-world environments inevitably encounter never-before-seen obstacles belonging to novel classes, i.e., the open-set \citep{scheirer2012toward}.
While open-set recognition has been traditionally studied under the assumption that training data contain only the $K$ target classes,
a more realistic learning paradigm embraces open-world data by incorporating them alongside task-specific data in training 
\citep{hendrycks2018deep, kong2021opengan, chakravarthy2025lidar}.
Furthermore,
domain knowledge for autonomous driving is inherently dynamic rather than static \citep{lin2022continual},
evolving through the discovery of new object categories \citep{bendale2015towards} and the continual refinement of semantic ontologies \citep{lin2022continual}.
Consequently,
perception systems should continually adapt as both the environment and the underlying ontology evolve. An important direction for future research is to establish realistic benchmarks and research protocols that jointly address continual learning and open-set recognition in evolving open-world environments.

\textbf{Rethinking data annotation.}
While our work follows the prevailing supervised learning paradigm based on the standard train-validation protocol and large-scale labeled datasets,
it is equally important to rethink the data annotation process itself, the prerequisite for developing machine learning systems.
Today, deploying machine learning in high-stakes applications typically requires massive amounts of annotated data,
which are commonly obtained through crowdsourcing.
In this pipeline, large numbers of non-expert annotators are recruited to label data according to expert-crafted annotation guidelines. Because these annotators often lack domain-specific knowledge,
domain experts carefully define each target concept through detailed textual descriptions and representative visual examples.
For example, the publicly available \textit{nuScenes Annotation Instructions}
illustrate how annotation guidelines specify the target concepts for the
nuScenes dataset \citep{nuscenes-user-guide}.
The annotators then follow these guidelines to produce large-scale annotations.
Although this annotation pipeline has become the \textit{de facto} standard in both academia and industry, it is costly, labor-intensive, and difficult to scale.
An intriguing alternative is to bypass large-scale manual annotation altogether by learning directly from expert-crafted annotation guidelines \citep{ma2025towards, madan2023revisiting, liu2025solving, liu2025surely}.
As 
increasingly capable foundation models become available,
a promising future direction is to adapt these models to the guidelines and learn domain expertise \citep{ma2025towards}, eliminating the need for ordinary annotators to exhaustively label massive datasets.
Since annotation guidelines naturally consist of a few visual demonstrations and textual definitions, learning from them can be formulated as a multi-modal few-shot learning problem.

More broadly, existing foundation models are typically pretrained on large-scale open-world data collected from the Internet or other real-world sources.
Such data are inherently long-tailed, weakly curated, and often lack domain-specific expertise. 
As a result, foundation models can inherit biases from their training data \citep{parashar2024neglected} and often struggle with tasks requiring specialized domain knowledge or expert reasoning  \citep{parashar2023prompting, sun2024alpha}.
These limitations are particularly concerning in high-stakes applications such as autonomous driving, 
where robustness, fairness, and reliability are essential. 
We therefore encourage the community to pursue foundation models tailored to domain-specific applications, investigate effective strategies for democratizing foundation models in specialized domains, and rigorously evaluate the biases, limitations, and safety risks associated with deploying foundation models in high-stakes settings.

\textbf{A warning of foundation models (FMs) for high-stakes applications.}
Despite the rapid progress of FM, including large language models and vision foundation models,
to the best of our knowledge,
there is no standalone FM capable of directly processing LiDAR point clouds for 3D proposal generation or detection.
Developing such FMs to effectively handle point clouds acquired from diverse LiDAR sensors therefore represents an important research direction.
More broadly, 
existing FMs are typically pretrained on large-scale  open-world data collected from the Internet or other real-world sources.
Such data are inherently long-tailed  and lack domain-specific expertise.
Consequently, 
FMs can inherit biases from their training data  \citep{parashar2024neglected} and often struggle with tasks requiring specialized domain knowledge or expert reasoning  \citep{parashar2023prompting, wei2023ov, sun2024alpha}.
These limitations are particularly concerning in high-stakes applications such as autonomous driving,
where robustness, fairness, and reliability are essential.
Hence, we encourage the community to pursue FMs by considering domain-specific applications, 
investigate effective strategies for democratizing FMs in specialized domains,
and rigorously evaluate the biases, limitations, and safety risks of FMs in high-stakes settings.

\section*{Acknowlegements}
Authors thank Shuoquan Wei for running preliminary experiments on 2D detection, and Yanan Li for the helpful discussion on long-tailed learning.

\section*{Funding Declaration}
This work was supported by 
the Science and Technology Development Fund of Macau (0058/2025/RIA2, 0067/2024/ITP2), 
the University of Macau (MYRG-CRG2026-00038-FIC, MYRG-GRG2026-00236-FIC, SRG2023-00044-FST),
the Institute of Collaborative Innovation, 
the CK Foundation, 
the Plato Intiative,
the National Natural Science Foundation of China (52232015),
the CMU Center for Autonomous Vehicle Research, and NSF GRFP (DGE2140739).

\section*{Data Availability Statement}
The datasets used in our experiments are  nuScenes and Argoverse 2, which publicly available and can be downloaded from their official websites: \url{https://www.nuscenes.org/} and \url{https://www.argoverse.org/av2.html}, respectively.

\section*{Code Availability Statement}
We have released the code and models for reproducing the results reported in the paper in the 
The code has been made publicly available in the \href{https://github.com/cc50121/lt3d-lf}{Github repository},
under the MIT License.

\section*{Other Declarations}
\begin{itemize}
\item Ethics approval: N/A
\item Consent to participate: N/A
\item Consent for publication: N/A
\end{itemize}

\section*{Authors' Contributions}
SK and DR designed an initial project back in 2022.
AD helped shape the topic of this project. 
SK extended and directed this project.
YM and NP prepared datasets, trained models, and conducted experiments, supervised by SK, DR and WH.
YM, NP, DR, and SK wrote the paper.
YM, NP, and SK revised the paper.
All the authors contributed to this project.
The authors have no conflict of interest to declare that are relevant to the content of this article.

\bibliographystyle{sn-mathphys-ay}
\bibliography{sn-bibliography}

@String(IJCV = {Int. J. Comput. Vis.})

@String(CVPR= {IEEE Conf. Comput. Vis. Pattern Recog.})

@String(ICCV= {Int. Conf. Comput. Vis.})

@String(ECCV= {Eur. Conf. Comput. Vis.})

@String(ICLR = {Int. Conf. Learn. Represent.})

@String(AAAI = {AAAI})

@String(IJCV  = {IJCV})

@String(CVPR  = {CVPR})

@String(ICCV  = {ICCV})

@String(ECCV  = {ECCV})

@String(ICLR  = {ICLR})

@inproceedings{peri2022towards,
  title={Towards Long-Tailed 3D Detection},
  author={Peri, Neehar and Dave, Achal and Ramanan, Deva and Kong, Shu},
  booktitle={Conference on Robot Learning (CoRL)},
  year={2022}
}

@inproceedings{alshammari2022long,
  title={Long-Tailed Recognition via Weight Balancing},
  author={Alshammari, Shaden and Wang, Yu-Xiong and Ramanan, Deva and Kong, Shu},
  booktitle={IEEE/CVF Conference on Computer Vision and Pattern Recognition},
  year={2022}
}

@inproceedings{ma2025towards,
  title={AutoExpert: Automating 3D LiDAR Annotation from Expert-Crafted Guidelines},
  author={Ma, Yechi and Hua, Wei and Kong, Shu},
  booktitle={Advances in Neural Information Processing Systems (NeurIPS)},
  year={2026}
}

@inproceedings{liu2024grounding,
  title={Grounding dino: Marrying dino with grounded pre-training for open-set object detection},
  author={Liu, Shilong and Zeng, Zhaoyang and Ren, Tianhe and Li, Feng and Zhang, Hao and Yang, Jie and Jiang, Qing and Li, Chunyuan and Yang, Jianwei and Su, Hang and others},
  booktitle={European conference on computer vision},
  year={2024}
}

@article{wei2023ov,
  title={Ov-parts: Towards open-vocabulary part segmentation},
  author={Wei, Meng and Yue, Xiaoyu and Zhang, Wenwei and Kong, Shu and Liu, Xihui and Pang, Jiangmiao},
  journal={Advances in Neural Information Processing Systems},
  year={2023}
}

@inproceedings{sun2024alpha,
  title={Alpha-clip: A clip model focusing on wherever you want},
  author={Sun, Zeyi and Fang, Ye and Wu, Tong and Zhang, Pan and Zang, Yuhang and Kong, Shu and Xiong, Yuanjun and Lin, Dahua and Wang, Jiaqi},
  booktitle={Proceedings of the IEEE/CVF conference on computer vision and pattern recognition},
  year={2024}
}

@inproceedings{liu2025solving,
  title={Solving Semi-Supervised Few-Shot Learning from an Auto-Annotation Perspective},
  author={Liu, Tian and Basu, Anwesha and Caverlee, James and Kong, Shu},
  booktitle={ECCV},
  year={2026}
}

@inproceedings{chang2019argoverse,
  title={Argoverse: 3d tracking and forecasting with rich maps},
  author={Chang, Ming-Fang and Lambert, John and Sangkloy, Patsorn and Singh, Jagjeet and Bak, Slawomir and Hartnett, Andrew and Wang, De and Carr, Peter and Lucey, Simon and Ramanan, Deva and others},
  booktitle={IEEE/CVF Conference on Computer Vision and Pattern Recognition},
  year={2019}
}

@inproceedings{
wilson2021argoverse,
title={Argoverse 2: Next Generation Datasets for Self-Driving Perception and Forecasting},
author={Benjamin Wilson and William Qi and Tanmay Agarwal and John Lambert and Jagjeet Singh and Siddhesh Khandelwal and Bowen Pan and Ratnesh Kumar and Andrew Hartnett and Jhony Kaesemodel Pontes and Deva Ramanan and Peter Carr and James Hays},
booktitle={Neural Information Processing Systems Datasets and Benchmarks Track},
year={2021}
}

@inproceedings{lang2019pointpillars,
  author    = {Alex H. Lang and
               Sourabh Vora and
               Holger Caesar and
               Lubing Zhou and
               Jiong Yang and
               Oscar Beijbom},
  title     = {PointPillars: Fast Encoders for Object Detection From Point Clouds},
  booktitle = {{IEEE} Conference on Computer Vision and Pattern Recognition (CVPR)},
  year      = {2019}
}

@article{zhu2019class,
  title={Class-balanced grouping and sampling for point cloud 3d object detection},
  author={Zhu, Benjin and Jiang, Zhengkai and Zhou, Xiangxin and Li, Zeming and Yu, Gang},
  journal={arXiv preprint arXiv:1908.09492},
  year={2019}
}

@inproceedings{yin2021center,
  title={Center-based 3d object detection and tracking},
  author={Yin, Tianwei and Zhou, Xingyi and Krahenbuhl, Philipp},
  booktitle={CVPR},
  year={2021}
}

@inproceedings{geiger2012we,
  title={Are we ready for autonomous driving? the kitti vision benchmark suite},
  author={Geiger, Andreas and Lenz, Philip and Urtasun, Raquel},
  booktitle={2012 IEEE conference on computer vision and pattern recognition},  
  year={2012},
  organization={IEEE}
}

@inproceedings{xingyi2020centertrack,
  title={Tracking objects as points},
  author={Zhou, Xingyi and Koltun, Vladlen and Kr{\"a}henb{\"u}hl, Philipp},
  booktitle={European conference on computer vision},
  pages={474--490},
  year={2020},
  organization={Springer}
}

@article{liu2025surely,
  title={Visual Species Recognition with Large Multimodal Models as Post-Hoc Correctors},
  author={Liu, Tian and Basu, Anwesha and Caverlee, James and Kong, Shu},
  journal={arXiv preprint arXiv:2512.15748},
  year={2025}
}

@inproceedings{ren2015faster,
  title={Faster r-cnn: Towards real-time object detection with region proposal networks},
  author={Ren, Shaoqing and He, Kaiming and Girshick, Ross and Sun, Jian},
  booktitle={Advances in neural information processing systems},
  year={2015}
}

@inproceedings{caesar2020nuscenes,
  title={nuscenes: A multimodal dataset for autonomous driving},
  author={Caesar, Holger and Bankiti, Varun and Lang, Alex H and Vora, Sourabh and Liong, Venice Erin and Xu, Qiang and Krishnan, Anush and Pan, Yu and Baldan, Giancarlo and Beijbom, Oscar},
  booktitle={CVPR},
  year={2020}
}

@inproceedings{Lin2014MicrosoftCC,
  title={Microsoft COCO: Common Objects in Context},
  author={Tsung-Yi Lin and Michael Maire and Serge J. Belongie and James Hays and Pietro Perona and Deva Ramanan and Piotr Doll{\'a}r and C. Lawrence Zitnick},
  booktitle={European Conference on Computer Vision},
  year={2014}
}

@inproceedings{liu2025few,
  title={Few-shot recognition via stage-wise retrieval-augmented finetuning},
  author={Liu, Tian and Zhang, Huixin and Parashar, Shubham and Kong, Shu},
  booktitle={Proceedings of the IEEE/CVF Conference on Computer Vision and Pattern Recognition},
  year={2025}
}

@article{lin2022continual,
  title={Continual learning with evolving class ontologies},
  author={Lin, Zhiqiu and Pathak, Deepak and Wang, Yu-Xiong and Ramanan, Deva and Kong, Shu},
  journal={Advances in Neural Information Processing Systems},
  year={2022}
}

@inproceedings{bendale2015towards,
  title={Towards open world recognition},
  author={Bendale, Abhijit and Boult, Terrance},
  booktitle={Proceedings of the IEEE conference on computer vision and pattern recognition},
  year={2015}
}

@article{scheirer2012toward,
  title={Toward open set recognition},
  author={Scheirer, Walter J and de Rezende Rocha, Anderson and Sapkota, Archana and Boult, Terrance E},
  journal={IEEE transactions on pattern analysis and machine intelligence},
  volume={35},
  number={7},
  pages={1757--1772},
  year={2012},
  publisher={IEEE}
}

@inproceedings{parashar2023prompting,
  title={Prompting scientific names for zero-shot species recognition},
  author={Parashar, Shubham and Lin, Zhiqiu and Li, Yanan and Kong, Shu},
  booktitle={Proceedings of the 2023 Conference on Empirical Methods in Natural Language Processing},
  pages={9856--9861},
  year={2023}
}

@article{chakravarthy2025lidar,
  title={Lidar panoptic segmentation in an open world},
  author={Chakravarthy, Anirudh S and Ganesina, Meghana Reddy and Hu, Peiyun and Leal-Taix{\'e}, Laura and Kong, Shu and Ramanan, Deva and Osep, Aljosa},
  journal={International Journal of Computer Vision},
  volume={133},
  number={3},
  pages={1153--1174},
  year={2025},
  publisher={Springer}
}

@inproceedings{NIPS2015_fasterRCNN,
 author = {Ren, Shaoqing and He, Kaiming and Girshick, Ross and Sun, Jian},
 booktitle = {Advances in Neural Information Processing Systems},
 title = {Faster R-CNN: Towards Real-Time Object Detection with Region Proposal Networks},
 year = {2015}
}

@inproceedings{ma2026roadside,
  title={Roadside Monocular 3D Detection Prompted by 2D Detection},
  author={Ma, Yechi and Hua, Wei and Li, Yanan and Kong, Shu},
  booktitle={Proceedings of the IEEE/CVF Winter Conference on Applications of Computer Vision},
  year={2026}
}

@article{felzenszwalb2009object,
  title={Object detection with discriminatively trained part-based models},
  author={Felzenszwalb, Pedro F and Girshick, Ross B and McAllester, David and Ramanan, Deva},
  journal={IEEE transactions on pattern analysis and machine intelligence},
  year={2009},
  publisher={IEEE}
}

@inproceedings{xu2020universal,
  title={Universal-rcnn: Universal object detector via transferable graph r-cnn},
  author={Xu, Hang and Fang, Linpu and Liang, Xiaodan and Kang, Wenxiong and Li, Zhenguo},
  booktitle={Proceedings of the AAAI Conference on Artificial Intelligence},
  year={2020}
}

@inproceedings{redmon2017yolo9000,
  title={YOLO9000: better, faster, stronger},
  author={Redmon, Joseph and Farhadi, Ali},
  booktitle={Proceedings of the IEEE conference on computer vision and pattern recognition},  
  year={2017}
}

@inproceedings{wang2019towards,
  title={Towards universal object detection by domain attention},
  author={Wang, Xudong and Cai, Zhaowei and Gao, Dashan and Vasconcelos, Nuno},
  booktitle={CVPR},  
  year={2019}
}

@inproceedings{li2022grounded,
  title={Grounded language-image pre-training},
  author={Li, Liunian Harold and Zhang, Pengchuan and Zhang, Haotian and Yang, Jianwei and Li, Chunyuan and Zhong, Yiwu and Wang, Lijuan and Yuan, Lu and Zhang, Lei and Hwang, Jenq-Neng and others},
  booktitle={CVPR},
  year={2022}
}

@inproceedings{zhou2022simple,
  title={Simple multi-dataset detection},
  author={Zhou, Xingyi and Koltun, Vladlen and Kr{\"a}henb{\"u}hl, Philipp},
  booktitle={CVPR},  
  year={2022}
}

@inproceedings{sun2020waymo,
  author    = {Pei Sun and
               Henrik Kretzschmar and
               Xerxes Dotiwalla and
               Aurelien Chouard and
               Vijaysai Patnaik and
               Paul Tsui and
               James Guo and
               Yin Zhou and
               Yuning Chai and
               Benjamin Caine and
               Vijay Vasudevan and
               Wei Han and
               Jiquan Ngiam and
               Hang Zhao and
               Aleksei Timofeev and
               Scott Ettinger and
               Maxim Krivokon and
               Amy Gao and
               Aditya Joshi and
               Yu Zhang and
               Jonathon Shlens and
               Zhifeng Chen and
               Dragomir Anguelov},
  title     = {Scalability in Perception for Autonomous Driving: Waymo Open Dataset},
  booktitle = {IEEE/CVF Conference on Computer Vision and Pattern Recognition (CVPR)},
  year      = {2020}
}

@inproceedings{gupta2023far3det,
  title={Far3Det: Towards Far-Field 3D Detection},
  author={Gupta, Shubham and Kanjani, Jeet and Li, Mengtian and Ferroni, Francesco and Hays, James and Ramanan, Deva and Kong, Shu},
  booktitle={WACV},
  year={2023}
}

@inproceedings{wong2020identifying,
  title={Identifying unknown instances for autonomous driving},
  author={Wong, Kelvin and Wang, Shenlong and Ren, Mengye and Liang, Ming and Urtasun, Raquel},
  booktitle={CoRL},
  year={2020}
}

@article{dawid1979conditional,
  title={Conditional independence in statistical theory},
  author={Dawid, A Philip},
  journal={Journal of the Royal Statistical Society Series B: Statistical Methodology},
  volume={41},
  number={1},
  pages={1--15},
  year={1979},
  publisher={Oxford University Press}
}

@inproceedings{khurana2024shelf,
  title={Shelf-supervised cross-modal pre-training for 3d object detection},
  author={Khurana, Mehar and Peri, Neehar and Hays, James and Ramanan, Deva},
  booktitle={CoRL},
  year={2024}
}

@article{nuscenes-user-guide,
  title={\url{https://github.com/nutonomy/nuscenes-devkit/blob/master/docs/instructions_nuscenes.md#instructions}},
  author={nuScenes}
}

@inproceedings{hendrycks2018deep,
  title={Deep anomaly detection with outlier exposure},
  author={Hendrycks, Dan and Mazeika, Mantas and Dietterich, Thomas},
  booktitle={ICLR},
  year={2019}
}

@inproceedings{liu2016ssd,
  title={Ssd: Single shot multibox detector},
  author={Liu, Wei and Anguelov, Dragomir and Erhan, Dumitru and Szegedy, Christian and Reed, Scott and Fu, Cheng-Yang and Berg, Alexander C},
  booktitle={ECCV},
  year={2016}
}

@inproceedings{wang2021fcos3d,
  author    = {Tai Wang and
               Xinge Zhu and
               Jiangmiao Pang and
               Dahua Lin},
  title     = {{FCOS3D:} Fully Convolutional One-Stage Monocular 3D Object Detection},
  booktitle = {ICCV},
  year      = {2021}
}

@inproceedings{redmon2016you,
  title={You only look once: Unified, real-time object detection},
  author={Redmon, Joseph and Divvala, Santosh and Girshick, Ross and Farhadi, Ali},
  booktitle={Proceedings of the IEEE conference on computer vision and pattern recognition},  
  year={2016}
}

@inproceedings{wang2022yolov7,
  title={YOLOv7: Trainable bag-of-freebies sets new state-of-the-art for real-time object detectors},
  author={Wang, Chien-Yao and Bochkovskiy, Alexey and Liao, Hong-Yuan Mark},
  booktitle={2023 IEEE/CVF conference on computer vision and pattern recognition (CVPR)},
  pages={7464--7475},
  year={2023},
  organization={Ieee}
}

@article{khan2017cost,
  title={Cost-sensitive learning of deep feature representations from imbalanced data},
  author={Khan, Salman H and Hayat, Munawar and Bennamoun, Mohammed and Sohel, Ferdous A and Togneri, Roberto},
  journal={IEEE transactions on neural networks and learning systems},  
  year={2017},
  publisher={IEEE}
}

@inproceedings{cui2019class,
  title={Class-balanced loss based on effective number of samples},
  author={Cui, Yin and Jia, Menglin and Lin, Tsung-Yi and Song, Yang and Belongie, Serge},
  booktitle={CVPR},
  year={2019}
}

@inproceedings{cao2019learning,
  title={Learning Imbalanced Datasets with Label-Distribution-Aware Margin Loss},
  author={Cao, Kaidi and Wei, Colin and Gaidon, Adrien and Arechiga, Nikos and Ma, Tengyu},
  booktitle={NeurIPS},
  year={2019}
}

@inproceedings{lin2017focal,
  title={Focal loss for dense object detection},
  author={Lin, Tsung-Yi and Goyal, Priya and Girshick, Ross and He, Kaiming and Doll{\'a}r, Piotr},
  booktitle={ICCV},
  year={2017}
}

@article{zhang2021deep,
  author    = {Yifan Zhang and
               Bingyi Kang and
               Bryan Hooi and
               Shuicheng Yan and
               Jiashi Feng},
  title     = {Deep Long-Tailed Learning: A Survey},
  journal   = {{IEEE} Transactions on Pattern Analysis and Machine Intelligence},
  volume    = {45},
  number    = {9},
  pages     = {10795--10816},
  year      = {2023}
}

@inproceedings{zhang2021distribution,
  title={Distribution Alignment: A Unified Framework for Long-tail Visual Recognition},
  author={Zhang, Songyang and Li, Zeming and Yan, Shipeng and He, Xuming and Sun, Jian},
  booktitle={CVPR},
  year={2021}
}

@inproceedings{tang2020long,
  title={Long-Tailed Classification by Keeping the Good and Removing the Bad Momentum Causal Effect},
  author={Tang, Kaihua and Huang, Jianqiang and Zhang, Hanwang},
  booktitle={NeurIPS},
  year={2020}
}

@article{huang2019deep,
  title={Deep imbalanced learning for face recognition and attribute prediction},
  author={Huang, Chen and Li, Yining and Loy, Chen Change and Tang, Xiaoou},
  journal={IEEE transactions on pattern analysis and machine intelligence},
  volume={42},
  number={11},
  pages={2781--2794},
  year={2019},
  publisher={IEEE}
}

@inproceedings{drummond2003c4,
  title={C4. 5, class imbalance, and cost sensitivity: why under-sampling beats over-sampling},
  author={Drummond, Chris and Holte, Robert C and others},
  booktitle={Workshop on learning from imbalanced datasets II},
  year={2003}
}

@inproceedings{han2005borderline,
  title={Borderline-SMOTE: a new over-sampling method in imbalanced data sets learning},
  author={Han, Hui and Wang, Wen-Yuan and Mao, Bing-Huan},
  booktitle={International Conference on Intelligent Computing},
  year={2005},
  organization={Springer}
}

@inproceedings{van2018inaturalist,
  title={The inaturalist species classification and detection dataset},
  author={Van Horn, Grant and Mac Aodha, Oisin and Song, Yang and Cui, Yin and Sun, Chen and Shepard, Alex and Adam, Hartwig and Perona, Pietro and Belongie, Serge},
  booktitle={CVPR},
  year={2018}
}

@inproceedings{gupta2019lvis,
  title={LVIS: A dataset for large vocabulary instance segmentation},
  author={Gupta, Agrim and Dollar, Piotr and Girshick, Ross},
  booktitle={CVPR},
  year={2019}
}

@article{reed2001pareto,
  title={The Pareto, Zipf and other power laws},
  author={Reed, William J},
  journal={Economics letters},
  year={2001},
  publisher={Elsevier}
}

@inproceedings{liu2019large,
  title={Large-scale long-tailed recognition in an open world},
  author={Liu, Ziwei and Miao, Zhongqi and Zhan, Xiaohang and Wang, Jiayun and Gong, Boqing and Yu, Stella X},
  booktitle={CVPR},
  year={2019}
}

@article{everingham2015pascal,
  title={The pascal visual object classes challenge: A retrospective},
  author={Everingham, Mark and Eslami, SM Ali and Van Gool, Luc and Williams, Christopher KI and Winn, John and Zisserman, Andrew},
  journal={IJCV},
  year={2015},
  publisher={Springer}
}

@inproceedings{guo2017calibration,
  title={On calibration of modern neural networks},
  author={Guo, Chuan and Pleiss, Geoff and Sun, Yu and Weinberger, Kilian Q},
  booktitle={ICML},
  year={2017}
}

@article{taeihagh2019governing,
  title={Governing autonomous vehicles: emerging responses for safety, liability, privacy, cybersecurity, and industry risks},
  author={Taeihagh, Araz and Lim, Hazel Si Min},
  journal={Transport Reviews},
  year={2019},
  publisher={Taylor \& Francis}
}

@inproceedings{kong2021opengan,
  title={OpenGAN: Open-set recognition via open data generation},
  author={Kong, Shu and Ramanan, Deva},
  booktitle={ICCV},
  year={2021}
}

@inproceedings{peri2023empirical,
  title={An empirical analysis of range for 3d object detection},
  author={Peri, Neehar and Li, Mengtian and Wilson, Benjamin and Wang, Yu-Xiong and Hays, James and Ramanan, Deva},
  booktitle={Proceedings of the IEEE/CVF International Conference on Computer Vision},  
  year={2023}
}

@article{russakovsky2015imagenet,
  title={Imagenet large scale visual recognition challenge},
  author={Russakovsky, Olga and Deng, Jia and Su, Hao and Krause, Jonathan and Satheesh, Sanjeev and Ma, Sean and Huang, Zhiheng and Karpathy, Andrej and Khosla, Aditya and Bernstein, Michael and others},
  journal={International journal of computer vision},
  year={2015},
  publisher={Springer}
}

@article{chawla2002smote,
  title={SMOTE: synthetic minority over-sampling technique},
  author={Chawla, Nitesh V and Bowyer, Kevin W and Hall, Lawrence O and Kegelmeyer, W Philip},
  journal={Journal of artificial intelligence research},  
  year={2002}
}

@article{hu2023ea,
  title={Ea-lss: Edge-aware lift-splat-shot framework for 3d bev object detection},
  author={Hu, Haotian and Wang, Fanyi and Su, Jingwen and Wang, Yaonong and Hu, Laifeng and Fang, Weiye and Xu, Jingwei and Zhang, Zhiwang},
  journal={arXiv preprint arXiv:2303.17895},
  year={2023}
}

@inproceedings{yin2024fusion,
  title={Is-fusion: Instance-scene collaborative fusion for multimodal 3d object detection},
  author={Yin, Junbo and Shen, Jianbing and Chen, Runnan and Li, Wei and Yang, Ruigang and Frossard, Pascal and Wang, Wenguan},
  booktitle={CVPR},  
  year={2024}
}

@inproceedings{shi2024lca,
  title={LCA-on-the-Line: Benchmarking Out-of-Distribution Generalization with Class Taxonomies},
  author={Shi, Jia and Gare, Gautam and Tian, Jinjin and Chai, Siqi and Lin, Zhiqiu and Vasudevan, Arun and Feng, Di and Ferroni, Francesco and Kong, Shu},
  booktitle={International Conference on Machine Learning (ICML)},
  year={2024}
}

@inproceedings{bodla2017soft,
  title={Soft-NMS--improving object detection with one line of code},
  author={Bodla, Navaneeth and Singh, Bharat and Chellappa, Rama and Davis, Larry S},
  booktitle={ICCV},
  year={2017}
}

@inproceedings{chen2021multimodal,
  title={Multimodal Object Detection via Probabilistic Ensembling},
  author={Chen, Yi-Ting and Shi, Jinghao and Ye, Zelin and Mertz, Christoph and Ramanan, Deva and Kong, Shu},
  booktitle={ECCV},
  year={2022}
}

@inproceedings{vora2020pointpainting,
  title={Pointpainting: Sequential fusion for 3d object detection},
  author={Vora, Sourabh and Lang, Alex H and Helou, Bassam and Beijbom, Oscar},
  booktitle={IEEE/CVF conference on computer vision and pattern recognition},
  year={2020}
}

@article{yin2021multimodal,
    author = {Yin, Tianwei and Zhou, Xingyi and Kr{\"a}henb{\"u}hl, Philipp},
    title={Multimodal Virtual Point 3D Detection},
    journal={NeurIPS},
    year={2021},
}

@inproceedings{pang2009clocs,
  title={CLOCs: Camera-LiDAR Object Candidates Fusion for 3D Object Detection},
  author={Pang, Su and Morris, Daniel and Radha, Hayder},
  booktitle={IEEE/RSJ International Conference on Intelligent Robots and Systems (IROS)},
  year={2020}
}

@inproceedings{qi2018frustum,
  title={Frustum pointnets for 3d object detection from rgb-d data},
  author={Qi, Charles R and Liu, Wei and Wu, Chenxia and Su, Hao and Guibas, Leonidas J},
  booktitle={IEEE conference on computer vision and pattern recognition},
  year={2018}
}

@inproceedings{yan2023cross,
  title={Cross modal transformer: Towards fast and robust 3d object detection},
  author={Yan, Junjie and Liu, Yingfei and Sun, Jianjian and Jia, Fan and Li, Shuailin and Wang, Tiancai and Zhang, Xiangyu},
  booktitle={Proceedings of the IEEE/CVF International Conference on Computer Vision},  
  year={2023}
}

@inproceedings{xu2018pointfusion,
  title={Pointfusion: Deep sensor fusion for 3d bounding box estimation},
  author={Xu, Danfei and Anguelov, Dragomir and Jain, Ashesh},
  booktitle={IEEE conference on computer vision and pattern recognition},
  year={2018}
}

@inproceedings{bai2022transfusion,
  title={Transfusion: Robust lidar-camera fusion for 3d object detection with transformers},
  author={Bai, Xuyang and Hu, Zeyu and Zhu, Xinge and Huang, Qingqiu and Chen, Yilun and Fu, Hongbo and Tai, Chiew-Lan},
  booktitle={CVPR},
  year={2022}
}

@inproceedings{openlongtailrecognition,
  title={Large-Scale Long-Tailed Recognition in an Open World},
  author={Liu, Ziwei and Miao, Zhongqi and Zhan, Xiaohang and Wang, Jiayun and Gong, Boqing and Yu, Stella X.},
  booktitle={CVPR},
  year={2019}
}

@inproceedings{detr2020carion,
  author    = {Nicolas Carion and
               Francisco Massa and
               Gabriel Synnaeve and
               Nicolas Usunier and
               Alexander Kirillov and
               Sergey Zagoruyko},
  title     = {End-to-End Object Detection with Transformers},
  booktitle = {ECCV},
  year      = {2020}
}

@ARTICLE{savage2022elderly,
  author={Savage, Neil},
  journal={Nature}, 
  title={Robots rise to meet the challenge of caring for old people}, 
  year={2022}
  }

@inproceedings{grauman2022ego4d,
  author    = {Kristen Grauman and
               Andrew Westbury and
               Eugene Byrne and
               Zachary Chavis and
               Antonino Furnari and
               Rohit Girdhar and
               Jackson Hamburger and et al.},
  title     = {Ego4D: Around the World in 3, 000 Hours of Egocentric Video},
  booktitle   = {Computer Vision and Pattern Recognition},
  year={2022}
}

@INPROCEEDINGS{li2020classifierimbalance,
  author={Li, Yu and Wang, Tao and Kang, Bingyi and Tang, Sheng and Wang, Chunfeng and Li, Jintao and Feng, Jiashi},
  booktitle={2020 IEEE/CVF Conference on Computer Vision and Pattern Recognition (CVPR)}, 
  title={Overcoming Classifier Imbalance for Long-Tail Object Detection With Balanced Group Softmax}, 
  year={2020},}

@article{Wu2019AHL,
  title={A hierarchical loss and its problems when classifying non-hierarchically},
  author={Cinna Julie Wu and Mark Tygert and Yann LeCun},
  journal={PLoS ONE},
  year={2019},
  volume={14}
}

@inproceedings{jiao2022msmdfusion,
  title={Msmdfusion: Fusing lidar and camera at multiple scales with multi-depth seeds for 3d object detection},
  author={Jiao, Yang and Jie, Zequn and Chen, Shaoxiang and Chen, Jingjing and Ma, Lin and Jiang, Yu-Gang},
  booktitle={2023 IEEE/CVF Conference on Computer Vision and Pattern Recognition (CVPR)},
  pages={21643--21652},
  year={2023},
  organization={IEEE}
}

@inproceedings{liu2022bevfusion,
  title={Bevfusion: Multi-task multi-sensor fusion with unified bird's-eye view representation},
  author={Liu, Zhijian and Tang, Haotian and Amini, Alexander and Yang, Xinyu and Mao, Huizi and Rus, Daniela L and Han, Song},
  booktitle={2023 IEEE international conference on robotics and automation (ICRA)},
  pages={2774--2781},
  year={2023},
  organization={ieee}
}

@inproceedings{li2022bevformer,
  title={Bevformer: Learning bird’s-eye-view representation from multi-camera images via spatiotemporal transformers},
  author={Li, Zhiqi and Wang, Wenhai and Li, Hongyang and Xie, Enze and Sima, Chonghao and Lu, Tong and Qiao, Yu and Dai, Jifeng},
  booktitle={ECCV},
  year={2022},
  organization={Springer}
}

@inproceedings{jiang2022polarformer,
  title={Polarformer: Multi-camera 3d object detection with polar transformer},
  author={Jiang, Yanqin and Zhang, Li and Miao, Zhenwei and Zhu, Xiatian and Gao, Jin and Hu, Weiming and Jiang, Yu-Gang},
  booktitle={Proceedings of the AAAI conference on Artificial Intelligence},
  volume={37},
  number={1},
  pages={1042--1050},
  year={2023}
}

@article{huang2022bevdet4d,
  title={Bevdet4d: Exploit temporal cues in multi-camera 3d object detection},
  author={Huang, Junjie and Huang, Guan},
  journal={arXiv preprint arXiv:2203.17054},
  year={2022}
}

@article{huang2021bevdet,
  title={Bevdet: High-performance multi-camera 3d object detection in bird-eye-view},
  author={Huang, Junjie and Huang, Guan and Zhu, Zheng and Du, Dalong},
  journal={arXiv preprint arXiv:2112.11790},
  year={2021}
}

@inproceedings{radford2021learning,
  title={Learning transferable visual models from natural language supervision},
  author={Radford, Alec and Kim, Jong Wook and Hallacy, Chris and Ramesh, Aditya and Goh, Gabriel and Agarwal, Sandhini and Sastry, Girish and Askell, Amanda and Mishkin, Pamela and Clark, Jack and others},
  booktitle={ICML},  
  year={2021},
  organization={PMLR}
}

@article{dosovitskiy2020image,
  title={An image is worth 16x16 words: Transformers for image recognition at scale},
  author={Dosovitskiy, Alexey and Beyer, Lucas and Kolesnikov, Alexander and Weissenborn, Dirk and Zhai, Xiaohua and Unterthiner, Thomas and Dehghani, Mostafa and Minderer, Matthias and Heigold, Georg and Gelly, Sylvain and others},
  journal={arXiv preprint arXiv:2010.11929},
  year={2020}
}

@inproceedings{philion2020lift,
  title={Lift, splat, shoot: Encoding images from arbitrary camera rigs by implicitly unprojecting to 3d},
  author={Philion, Jonah and Fidler, Sanja},
  booktitle={ECCV},
  year={2020}
}

@article{shi2022pv,
  title={PV-RCNN++: Point-voxel feature set abstraction with local vector representation for 3D object detection},
  author={Shi, Shaoshuai and Jiang, Li and Deng, Jiajun and Wang, Zhe and Guo, Chaoxu and Shi, Jianping and Wang, Xiaogang and Li, Hongsheng},
  journal={IJCV},  
  year={2022},
  publisher={Springer}
}

@inproceedings{tian2019fcos,
  title={Fcos: Fully convolutional one-stage object detection},
  author={Tian, Zhi and Shen, Chunhua and Chen, Hao and He, Tong},
  booktitle={ICCV},  
  year={2019}
}

@inproceedings{guizilini20203d,
  title={3d packing for self-supervised monocular depth estimation},
  author={Guizilini, Vitor and Ambrus, Rares and Pillai, Sudeep and Raventos, Allan and Gaidon, Adrien},
  booktitle={CVPR},  
  year={2020}
}

@inproceedings{qi2017pointnet,
  title={Pointnet: Deep learning on point sets for 3d classification and segmentation},
  author={Qi, Charles R and Su, Hao and Mo, Kaichun and Guibas, Leonidas J},
  booktitle={CVPR},  
  year={2017}
}

@inproceedings{zhang2022dino,
  title={Dino: Detr with improved denoising anchor boxes for end-to-end object detection},
  author={Zhang, Hao and Li, Feng and Liu, Shilong and Zhang, Lei and Su, Hang and Zhu, Jun and Ni, Lionel and Shum, Harry},
  booktitle={ICLR},
  year={2022}
}

@article{wilson20203d,
  title={3d for free: Crossmodal transfer learning using hd maps},
  author={Wilson, Benjamin and Kira, Zsolt and Hays, James},
  journal={arXiv preprint arXiv:2008.10592},
  year={2020}
}

@inproceedings{yang2022deepinteraction,
  title={DeepInteraction: 3D Object Detection via Modality Interaction},
  author={Yang, Zeyu and Chen, Jiaqi and Miao, Zhenwei and Li, Wei and Zhu, Xiatian and Zhang, Li},
  booktitle={NeurIPS},
  year={2022}
}

@InProceedings{Girshick_2014_CVPR,
author = {Girshick, Ross and Donahue, Jeff and Darrell, Trevor and Malik, Jitendra},
title = {Rich Feature Hierarchies for Accurate Object Detection and Semantic Segmentation},
booktitle = {CVPR},
month = {June},
year = {2014}
}

@article{madan2023revisiting,
  title={Revisiting few-shot object detection with vision-language models},
  author={Madan, Anish and Peri, Neehar and Kong, Shu and Ramanan, Deva},
  journal={Advances in Neural Information Processing Systems},
  volume={37},
  pages={19547--19560},
  year={2024}
}

@inproceedings{cai2025dstr,
  title={DSTR: Dual Scenes Transformer for Cross-Modal Fusion in 3D Object Detection},
  author={Cai, Haojie and Yin, Dongfu and Yu, Fei Richard and Xiong, SiTing},
  booktitle={IEEE/CVF Winter Conference on Applications of Computer Vision (WACV)},
  pages={3064--3073},
  year={2025},
  organization={IEEE}
}

@inproceedings{jin2025unimamba,
  title={UniMamba: Unified Spatial-Channel Representation Learning with Group-Efficient Mamba for LiDAR-based 3D Object Detection},
  author={Jin, Xin and Su, Haisheng and Liu, Kai and Ma, Cong and Wu, Wei and Hui, Fei and Yan, Junchi},
  booktitle={IEEE/CVF Conference on Computer Vision and Pattern Recognition Conference},
  pages={1407--1417},
  year={2025}
}

@inproceedings{liu2025fshnet,
  title={FSHNet: Fully Sparse Hybrid Network for 3D Object Detection},
  author={Liu, Shuai and Cui, Mingyue and Li, Boyang and Liang, Quanmin and Hong, Tinghe and Huang, Kai and Shan, Yunxiao},
  booktitle={IEEE/CVF Conference on Computer Vision and Pattern Recognition Conference},
  pages={8900--8909},
  year={2025}
}

@inproceedings{yin2023metric3d,
  title={Metric3d: Towards zero-shot metric 3d prediction from a single image},
  author={Yin, Wei and Zhang, Chi and Chen, Hao and Cai, Zhipeng and Yu, Gang and Wang, Kaixuan and Chen, Xiaozhi and Shen, Chunhua},
  booktitle={Proceedings of the IEEE/CVF international conference on computer vision},
  year={2023}
}

@inproceedings{dahling2026densebev,
  title={DenseBEV: Transforming BEV Grid Cells into 3D Objects},
  author={D{\"a}hling, Marius and Krebs, Sebastian and Z{\"o}llner, J Marius},
  booktitle={Proceedings of the IEEE/CVF Winter Conference on Applications of Computer Vision},
  pages={2370--2379},
  year={2026}
}

@inproceedings{wu2026ccf,
  title={CCF: Complementary Collaborative Fusion for Domain Generalized Multi-Modal 3D Object Detection},
  author={Wu, Yuchen and Wang, Kun and Pan, Yining and Zhao, Na},
  booktitle={Proceedings of the IEEE/CVF Conference on Computer Vision and Pattern Recognition},
  pages={18745--18754},
  year={2026}
}

@inproceedings{yang2026fomo,
  title={Fomo-3d: Using vision foundation models for long-tailed 3d object detection},
  author={Yang, Anqi Joyce and Tu, James and Dvornik, Nikita and Li, Enxu and Urtasun, Raquel},
  booktitle={Conference on Robot Learning (CoRL)},
  year={2025}
}

@inproceedings{parashar2024neglected,
  title={The neglected tails in vision-language models},
  author={Parashar, Shubham and Lin, Zhiqiu and Liu, Tian and Dong, Xiangjue and Li, Yanan and Ramanan, Deva and Caverlee, James and Kong, Shu},
  booktitle={Proceedings of the IEEE/CVF Conference on Computer Vision and Pattern Recognition},
  year={2024}
}

@inproceedings{zong2023detrs,
  title={Detrs with collaborative hybrid assignments training},
  author={Zong, Zhuofan and Song, Guanglu and Liu, Yu},
  booktitle={Proceedings of the IEEE/CVF international conference on computer vision},
  pages={6748--6758},
  year={2023}
}

\clearpage
\begin{appendices}

\appendix
\section{Implementation Details}
\label{sec:impl_details}

We use the open-source codebase det3d
mmcv, mmdetection and mmdetection3d to train FCOS3D, CenterPoint, and TransFusion.
We use the first-party implementations for other detectors, following the training schedule proposed by each respective paper. By default, we train the 2D RGB detector with 2D bounding boxes derived from nuScenes' 3D annotations and additionally train with 2D bounding boxes from nuImages where denoted. 
For 2D RGB detectors including YOLOV7, DINO, GroundingDINO, and Co-DINO,
we download their publicly released checkpoints.
To train stronger versions of them,
we adopt the stage-wise finetuning protocol \citep{liu2025few} that finetunes them over larger-scale and mixed datasets first,
and then finetunes on nuImages.
We describe important implementation details below.

\begin{itemize}
    \item {\em Input}. We adopt 10-frame aggregation for LiDAR densification when training LiDAR-based detectors on nuScenes and a 5-frame aggregation on Argoverse 2. We assume that we are provided with ego-vehicle poses for prior frames to align all LiDAR sweeps to the current ego-vehicle pose. Since LiDAR returns are sparse, this densification step is essential for accurate 3D detection.  By default, we train the 2D RGB detectors on the 2D bounding boxes derived by projecting 3D annotations to the 2D image plane and additionally train with 2D bounding boxes from nuImages where denoted. 
    
    \item {\em Model Architecture.} We adopt the architecture in \citep{zhu2019class} but make an important modification. The original architecture (for the standard nuScenes benchmark) has six heads designed for ten classes; each head has 64 filters. We first adapted this architecture for LT3D using seven heads designed for 18 classes. We then replace these seven heads with a single head consisting of 512 filters shared by all classes. 
    \item {\em Training Losses.} We use the sigmoid focal loss (for recognition) \citep{lin2017focal} and L1 regression loss (for localization) below. Existing works also use the same losses but only with fine labels; we apply the loss to both coarse and fine labels. Concretely, our loss function for CenterPoint is as follows: $L = L_{HM} + \lambda L_{REG}$, where $L_{HM} = \sum_{i=0}^{C} SigmoidFocalLoss(X_i, Y_i)$ and $L_{REG} = |X_{BOX} - Y_{BOX}|$, where $X_i$ and $Y_i$ are the $i^{th}$ class' predicted and ground-truth heat maps, while $X_{BOX}$ and $Y_{BOX}$ are the predicted and ground-truth box attributes. 
    Without our hierarchical loss, $C$=18. With our hierarchical loss, $C$=22 (18 fine grained + 3 coarse + 1 object class). $\lambda$ is set to 0.25. Modifications for other detectors similarly follow.
    \item {\em Optimization. } We train all LiDAR-only detectors for 20 epochs using an AdamW optimizer and a cyclic learning rate. We adopt a basic set of data augmentations, including global 3D tranformations, flip in BEV, and point shuffling during training. We train our model with 8 RTX 3090 GPUs and a batch size of 1 per GPU. The training noise (from random seed and system scheduling) is $<$ 1\% of the accuracy (standard deviation normalized by the mean).
    \item {\em Post-processing.} 
    We use non-maximum suppression (NMS) on detections {\em within} each class to suppress lower-scoring detections. In contrast, existing works apply NMS on all detections {\em across} classes, i.e., suppressing detections overlapping other classes' detections (e.g., a {\tt pedestrian} detection can suppress other {\tt pedestrian} {\em and} {\tt traffic-cone} detections).
\end{itemize}

\section{Analysis on Class Grouping}
\label{sec:analysis-class-grouping}

{
\setlength{\tabcolsep}{2mm} 
\begin{table*}[t] 
\small
\centering
\scalebox{0.85}{
\begin{tabular}{l c c c c c c c c c c c c c}
\toprule
CenterPoint & \multicolumn{1}{c}{Multi-Head} & \multicolumn{1}{c}{Car} & \multicolumn{1}{c}{Ped.} & \multicolumn{1}{c}{Barrier} & \multicolumn{1}{c}{TC} &  \multicolumn{1}{c}{Truck} & \multicolumn{1}{c}{Bus} & \multicolumn{1}{c}{Trailer} &  \multicolumn{1}{c}{CV} & \multicolumn{1}{c}{MC} & \multicolumn{1}{c}{Bicyc.}  \\ 
\midrule
Original                  & \checkmark &  87.7   &  87.7   &  70.7   &  74.0   &  63.6   & 72.7  &  \textbf{45.1}   &  \textbf{26.3}   &  64.7  &  47.9   \\ 
                             &            &  \textbf{89.1}   &  \textbf{88.4}   &  \textbf{70.8}   &  \textbf{74.3}   &  \textbf{64.8}   & \textbf{72.9}  &  42.0   &  25.7   &  \textbf{65.9}  &  \textbf{53.6}   \\ 
\midrule
\multirow{1}{*}{for LT3D}    & \checkmark &  82.4   &  ---   &  62.0   &  60.1   &  49.4   & 55.7  &  28.9   &  19.7   &  48.9  &  33.6   \\ 
                             &            &  \textbf{88.1}   &  ---   &  \textbf{72.4}   &  \textbf{72.7}   &  \textbf{62.7}   & \textbf{70.8}  &  \textbf{40.2}   &  \textbf{24.5}   &  \textbf{62.8}  &  \textbf{48.5}   \\ 
\midrule
\midrule 
\multirow{1}{*}{} & \multirow{1}{*}{}  & \multicolumn{1}{c}{Adult} & \multicolumn{1}{c}{\color{blue}PP} & \multicolumn{1}{c}{\color{blue}CW} & \multicolumn{1}{c}{\color{blue}Debris} &  \multicolumn{1}{c}{\color{blue}Child} & \multicolumn{1}{c}{\color{blue}Stroller} & \multicolumn{1}{c}{\color{blue}PO} &  \multicolumn{1}{c}{\color{blue}EV} & \multicolumn{1}{c}{\color{blue}PM} & \multicolumn{1}{c}{{\tt All}} \\ 
\midrule
Original                  & \checkmark &  ---   &  ---   &  ---   &  ---   &  ---   & ---  &  ---  &  ---  &  --- & 64.0 \\ 
                          &            &  ---   &  ---   &  ---   &  ---   &  ---   & ---  &  ---  &  ---   & --- & \textbf{64.8} \\ 
\midrule
\multirow{1}{*}{for LT3D}    & \checkmark &  81.2   &  21.7   &  14.2   & 1.1   &  \textbf{0.1}     & 0.1  &  \textbf{1.3}   &  0.1          &   \textbf{0.1}     & 31.2      \\ 
                             &            &  \textbf{86.3}   &  \textbf{32.7}   &  \textbf{22.2}   & \textbf{4.3}   &  \textbf{0.1}     & \textbf{4.3}  &  \textbf{1.8}    & 10.3          &   \textbf{0.1}  & \textbf{39.2}       \\ 
\bottomrule
\end{tabular}
}
\caption{\small 
Our proposed grouping-free detector head architecture consistently outperforms grouping-based approaches on both the standard and LT3D \ benchmarks. We note that sub-optimal grouping strategies (such as those adopted for LT3D) may yield significantly diminished performance, whereas optimized grouping strategies (such as those adopted for the standard setup) have comparable performance to the grouping-free approach. Note, {\tt TC} is {\tt traffic-cone}, {\tt CV} is {\tt construction vehicle}, {\tt MC} is {\tt motorcycle}, {\tt PP} is {\tt pushable-pullable}, {\tt CW} is {\tt construction-worker}, and {\tt PO} is {\tt police-officer}. We highlight  {\tt Medium} and {\tt Few} classes in {\color{blue}blue}.
}
\label{tab:segm}
\end{table*}
}

Many contemporary networks use a multi-head architecture that groups classes of similar size and shape to facilitate efficient feature sharing. For example, CenterPoint groups {\tt pedestrian} and {\tt traffic-cone} since these objects are both tall and skinny. We study the impact of grouping for both the standard and LT3D problem setups. We define the groups used for this study below. Each group is enclosed in curly braces. Our grouping-free head includes all classes into a single group.

\begin{itemize}
    \item Original: \{{\tt Car}\}, \{{\tt Truck}, {\tt Construction Vehicle}\}, \{{\tt Bus}, {\tt Trailer}\}, \{{\tt Barrier}\}, \{{\tt Motorcycle}, {\tt Bicycle}\}, \{{\tt Pedestrian}, {\tt Traffic Cone}\}
    \item LT3D: \{{\tt Car}\}, \{{\tt Truck}, {\tt Construction Vehicle}\}, \{{\tt Bus}, {\tt Trailer}\}, \{{\tt Barrier}\}, \{{\tt Motorcycle}, {\tt Bicycle}\}, \{{\tt Adult}, {\tt Child}, {\tt Construction Worker}, {\tt Police Officer}, {\tt Traffic Cone}\}, \{{\tt Pushable Pullable}, {\tt Debris}, {\tt Stroller}, {\tt Personal Mobility}, {\tt Emergency Vehicle}\}
\end{itemize}

We use the class groups proposed by prior works \citep{zhu2019class, yin2021center} for the standard benchmark and adapt this grouping for LT3D. Our proposed grouping-free detector head architecture consistently outperforms grouping-based approaches on both the standard and LT3D  benchmarks, as shown in Table~\ref{tab:segm}. We note that sub-optimal grouping strategies (such as those adopted for LT3D) may yield significantly diminished performance, whereas optimized grouping strategies (such as those adopted for the standard setup) have comparable performance to the grouping-free approach. The grouping-free approach simplifies architecture design, while also providing competitive performance.

Two insights allow us to train the grouping-free architecture. First, we make the grouping-free head proportionally larger to train more classes. The standard grouping setup contains 6 heads, each with 64 convolutional filters. Scaling up to the nearest power of two, our grouping-free head has 512 convolutional filters. Second, we do not perform between-class NMS. The standard setup performs NMS between classes in each group (e.g., since pedestrians and traffic cones are tall and skinny, the model should only predict that an object is either a traffic cone or a pedestrian). However, performing NMS between classes requires that confidence scores are calibrated, which is not the case. Moreover, for LT3D, score calibration becomes more important for {\tt rare} classes as these classes have lower confidence scores than {\tt common} classes on average, meaning that {\tt common} objects will likely suppress {\tt rare} objects within the same group. Our solution is to only perform within-class NMS, which is standard for 2D detectors \citep{ren2015faster}.

\section{Analysis of Hierarchical Training}
\label{sec:analysis-hierarchical-training}

{
\setlength{\tabcolsep}{1mm} 
\begin{table}[t]
\small
\centering
\vspace{1mm}
\scalebox{0.91}{
\begin{tabular}{l c c c c c c c c c c c}
\toprule
\multirow{1}{*}{Method} & \multirow{1}{*}{Hier.} & \multicolumn{1}{c}{\tt Many} & \multicolumn{1}{c}{\texttt{Medium}} & \multicolumn{1}{c}{\texttt{Few}} & \multicolumn{1}{c}{\tt All} \\ 

\midrule
\multirow{1}{*}{CenterPoint (w/o Hier.)}                       &      n/a      &  76.4             & 43.1         &   3.5        &   39.2        \\ 
\midrule
\multirow{4}{*}{CenterPoint w/ Hier.}        &  (a)   & \textbf{77.1}             & {45.1}          &   4.3      &   40.4       \\ 
\multirow{1}{*}{}                                &  (b)   &  {76.4}             & {45.0}          &   5.3      &   40.5       \\ 
\multirow{1}{*}{}                                &  (c)   &  {76.5}             & \textbf{45.2}          &   5.2      &   \textbf{40.6}       \\
\multirow{1}{*}{}                                &  (d)   &  {74.5}             & {43.5}          &   \textbf{5.6}      &   39.5       \\ 
\bottomrule
\end{tabular}
}
\caption{\small 
Different variants achieve similar performance. We note that other methods do improve accuracy in the tail by sacrificing performance in the head, suggesting that hybrid approaches that apply different techniques for head-vs-tail classes may further improve accuracy. Unlike \citep{li2020classifierimbalance, Wu2019AHL} which requires a strict label hierarchy, our approach is not limited to a hierarchy.
}
\label{tab:hierarchy_loss}
\vspace{-4mm}
\end{table}
}

Classic methods train a hierarchical softmax (in contrast to our simple approach of sigmoid focal loss with both fine and coarse classes), where one multiplies the class probabilities of the hierarchical predictions during training and inference \citep{Wu2019AHL}. We implemented such an approach, but found the training did not converge. Interestingly, \citep{Wu2019AHL} shows such a hierarchical softmax loss has little impact on long-tailed object detection (in 2D images), which is one reason they have not been historically adopted. Instead, we found better results using the method from \citep{li2020classifierimbalance} (a winning 2D object detection system on the LVIS \citep{gupta2019lvis} benchmark) which multiples class probabilities of predictions (e.g. $P_{CAR} = P_{OBJ} * P_{CAR}$) at test-time, even when such predictions are not trained with a hierarchical softmax. We tested three variants and compared it to our approach (which recall, uses only fine-grained class probabilities at inference). Table~\ref{tab:hierarchy_loss} compares their performance for LT3D.

\begin{enumerate}[label=(\alph*)]
    \item Ours (e.g., Fine-class score only)
    \item Object score * Fine-class  score (\citep{li2020classifierimbalance}, e.g. $P_{CAR} = P_{OBJ} * P_{CAR}$) 
    \item Coarse score * Fine-class score (Variant-1 of \citep{li2020classifierimbalance}, e.g. $P_{CAR} = P_{VEHICLE} * P_{CAR}$)
    \item Object score * Coarse score * Fine-class  score (Variant-2 of \citep{li2020classifierimbalance}, e.g. $P_{CAR} = P_{OBJ} * P_{VEHICLE} * P_{CAR}$)
\end{enumerate}

Unlike \citep{li2020classifierimbalance, Wu2019AHL} which require a strict label hierarchy, our approach is not limited to a hierarchy. We find that other hierarchical methods improve accuracy in the tail by sacrificing performance in the head, suggesting that hybrid approaches that apply different techniques for head-vs-tail classes may further improve accuracy.

\section{Analysis of 2D Detection Performance}
\label{sec:analysis-2D-detector-perf}

{
\setlength{\tabcolsep}{0.15em}
\begin{table}[t] 
\centering
\vspace{1mm}
\scalebox{0.91}{
\begin{tabular}{lrrrr}
\toprule
\multirow{1}{*}{Method}     
  & \multicolumn{1}{c}{{\tt All}} & \multicolumn{1}{c}{{\tt Many}} & \multicolumn{1}{c}{{\tt Medium}}  & \multicolumn{1}{c}{{\tt Few}} \\ 
\midrule
\multirow{1}{*}{FCOS3D (\textbf{3D} RGB Detector)}
 & \multirow{1}{*}{18.3} & \multirow{1}{*}{36.0} & \multirow{1}{*}{21.1} & \multirow{1}{*}{0.2}   \\
\multirow{1}{*}{BEVFormer (\textbf{3D} RGB Detector)} 
 & \multirow{1}{*}{23.0} & \multirow{1}{*}{40.8} & \multirow{1}{*}{28.2} & \multirow{1}{*}{2.1}   \\
 \multirow{1}{*}{PolarFormer (\textbf{3D} RGB Detector)}
 & \multirow{1}{*}{20.7} & \multirow{1}{*}{37.5} & \multirow{1}{*}{25.2} & \multirow{1}{*}{1.6}   \\
  \multirow{1}{*}{DenseBEV (\textbf{3D} RGB Detector)}
 & \multirow{1}{*}{24.4} & \multirow{1}{*}{41.2} & \multirow{1}{*}{30.1} & \multirow{1}{*}{3.7}   \\
\midrule
\multirow{1}{*}{YOLOV7 (\textbf{2D} RGB Detector)}
& \multirow{1}{*}{{42.3}} & \multirow{1}{*}{61.1} & \multirow{1}{*}{{46.3}} & \multirow{1}{*}{{22.0}}   \\
\multirow{1}{*}{DINO (\textbf{2D} RGB Detector)}
& \multirow{1}{*}{48.9} & \multirow{1}{*}{67.6} & \multirow{1}{*}{52.9} & \multirow{1}{*}{28.7}   \\
\multirow{1}{*}{GroundingDINO (\textbf{2D} RGB Detector)}
& \multirow{1}{*}{50.4} & \multirow{1}{*}{68.7} & \multirow{1}{*}{55.0} & \multirow{1}{*}{29.8}   \\
\multirow{1}{*}{Co-DINO (\textbf{2D} RGB Detector)}
& \multirow{1}{*}{\textbf{54.3}} & \multirow{1}{*}{\textbf{72.2}} & \multirow{1}{*}{\textbf{58.6}} & \multirow{1}{*}{\textbf{34.5}}   \\
\hline
\end{tabular}
}
\caption{\small
{\bf Evaluating 2D detection performance}.
Simply projecting 3D detections from RGB-only 3D detectors to 2D image plane yields considerably lower 2D detection mAP across all classes. 
However, 2D RGB detectors achieve higher performance.
Intuitively, 2D RGB detectors achieve better classification performance and predict tighter fitting bounding boxes than those derived by projecting 3D detections to the image plane. This explains why fusing detections from 2D RGB detectors yields better LT3D performance than detections from 3D RGB detectors. All detectors in this table are only trained on the nuScenes train-set.
Refer to Table \ref{tab:fine-grained 2D dectector} for per-class results of ensembling the strong 2D RGB detectors with the 3D LiDAR detector CenterPoint.
}
\label{tab:2d_perf} 
\end{table}
}

{
\setlength{\tabcolsep}{0.4em}
\begin{table}[t]
\small
\centering
\vspace{1mm}
\scalebox{0.72}{
\begin{tabular}{lcccc}
\toprule
  \multirow{1}{*}{Class name}     
  & \multirow{1}{*}{w/ YOLOV7} & \multirow{1}{*}{w/ DINO} & \multirow{1}{*}{w/ GD} & \multirow{1}{*}{w/ Co-DINO}\\
  \midrule
\rowcolor{col11}\multirow{1}{*} {Personal Mobility}
 & \multirow{1}{*}{18.9} & \multirow{1}{*}{35.8} & \multirow{1}{*}{37.1} & \multirow{1}{*}{42.3}\\
  \rowcolor{col11}\multirow{1}{*} {Emergency Vehicle}
 & \multirow{1}{*}{12.5} & \multirow{1}{*}{21.1} & \multirow{1}{*}{24.7} & \multirow{1}{*}{28.0}\\
    \rowcolor{col11}\multirow{1}{*} {Police Officer}
 & \multirow{1}{*}{10.9} & \multirow{1}{*}{14.4} & \multirow{1}{*}{15.9} & \multirow{1}{*}{16.1}\\
   \rowcolor{col11}\multirow{1}{*} {Stroller}
 & \multirow{1}{*}{27.5} & \multirow{1}{*}{36.9} & \multirow{1}{*}{40.8} & \multirow{1}{*}{45.4}\\
  \rowcolor{col11}\multirow{1}{*} {Child}
 & \multirow{1}{*}{16.7} & \multirow{1}{*}{24.9} & \multirow{1}{*}{28.6} & \multirow{1}{*}{32.1}\\
    \rowcolor{col11}\multirow{1}{*} {Debris}
 & \multirow{1}{*}{5.3} & \multirow{1}{*}{6.1} & \multirow{1}{*}{8.9} & \multirow{1}{*}{11.4}\\
  \rowcolor{yellow}\multirow{1}{*} {Construction Worker}
 & \multirow{1}{*}{61.5} & \multirow{1}{*}{61.9} & \multirow{1}{*}{63.7} & \multirow{1}{*}{63.9}\\
  \rowcolor{yellow}\multirow{1}{*} {Bicycle}
 & \multirow{1}{*}{66.6} & \multirow{1}{*}{71.2} & \multirow{1}{*}{76.7} & \multirow{1}{*}{77.3}\\
 \rowcolor{yellow}\multirow{1}{*} {Motorcycle}
 & \multirow{1}{*}{73.2} & \multirow{1}{*}{77.6} & \multirow{1}{*}{79.5} & \multirow{1}{*}{80.7}\\
  \rowcolor{yellow}\multirow{1}{*} {Construction Vehicle}
 & \multirow{1}{*}{31.9} & \multirow{1}{*}{35.5} & \multirow{1}{*}{38.1} & \multirow{1}{*}{40.0}\\
  \rowcolor{yellow}\multirow{1}{*} {Bus}
 & \multirow{1}{*}{75.7} & \multirow{1}{*}{76.0} & \multirow{1}{*}{77.1} & \multirow{1}{*}{79.2}\\
     \rowcolor{yellow}\multirow{1}{*} {Pushable Pullable}
 & \multirow{1}{*}{49.9} & \multirow{1}{*}{54.3} & \multirow{1}{*}{55.1} & \multirow{1}{*}{57.2}\\
  \rowcolor{yellow}\multirow{1}{*} {Trailer}
 & \multirow{1}{*}{40.9} & \multirow{1}{*}{45.6} & \multirow{1}{*}{47.3} & \multirow{1}{*}{49.4}\\
 \rowcolor{col3}\multirow{1}{*} {Truck}
 & \multirow{1}{*}{59.2} & \multirow{1}{*}{64.8} & \multirow{1}{*}{68.0} & \multirow{1}{*}{68.1}\\
\rowcolor{col3}\multirow{1}{*} {Traffic Cone}
 & \multirow{1}{*}{78.0} & \multirow{1}{*}{79.6} & \multirow{1}{*}{80.8} & \multirow{1}{*}{82.3}\\
    \rowcolor{col3}\multirow{1}{*} {Barrier} 
 & \multirow{1}{*}{73.1} & \multirow{1}{*}{74.4} & \multirow{1}{*}{75.1} & \multirow{1}{*}{75.6}\\
 \rowcolor{col3}\multirow{1}{*} {Adult}
 & \multirow{1}{*}{86.4} & \multirow{1}{*}{87.3} & \multirow{1}{*}{87.9} & \multirow{1}{*}{89.2}\\
  \rowcolor{col3}\multirow{1}{*}{Car} &
  \multirow{1}{*}{86.8} &
  \multirow{1}{*}{86.9} &
  \multirow{1}{*}{87.7} &
  \multirow{1}{*}{88.5}\\ 
 \midrule
 \rowcolor{col11}\multirow{1}{*}{\tt Few}
 & \multirow{1}{*}{15.3} & \multirow{1}{*}{23.3} & \multirow{1}{*}{26.0} & \multirow{1}{*}{29.2}\\
      \rowcolor{yellow}\multirow{1}{*}{\tt Medium}
 & \multirow{1}{*}{57.1} & \multirow{1}{*}{60.3} & \multirow{1}{*}{62.5} & \multirow{1}{*}{64.0}\\
     \cellcolor{col3}\multirow{1}{*}{\tt Many}
 & \cellcolor{col3}\multirow{1}{*}{76.7} & \cellcolor{col3}\multirow{1}{*}{78.6} 
 & \cellcolor{col3}\multirow{1}{*}{79.9} & \cellcolor{col3}\multirow{1}{*}{80.7} \\
  \midrule
       \multirow{1}{*} {All}
 & \multirow{1}{*}{48.6} & \multirow{1}{*}{53.1} & \multirow{1}{*}{55.2} & \multirow{1}{*}{57.0}\\
\bottomrule
\end{tabular}
}
\caption{\small
{\bf Using a stronger 2D detector improves 3D detection by MMLF.}
We compare YOLOV7, DINO, GroundingDINO, and Co-DINO as the 2D detector,
applied along with the 3D LiDAR detector CenterPoint~\citep{yin2021center} in our MMLF.
Results demonstrate that using stronger 2D detectors, especially Co-DINO, yields significantly better 3D detection performance on all classes, particularly on {\tt medium} and {\tt few} classes.
Refer to Fig.~\ref{fig:histogram} for the distribution of these classes.
}
\vspace{-4mm}
\label{tab:fine-grained 2D dectector} 
\end{table}
}

Table \ref{tab:2d_perf} shows that 2D RGB detectors, especially Co-DINO, outperform state-of-the-art 3D RGB detectors (e.g., DenseBEV) for \emph{2D detection} on the nuScenes val-set. Importantly, Co-DINO performs significantly better than DenseBEV (34.5 vs. 3.7 mAP) on rare classes.
Furthermore, Table~\ref{tab:fine-grained 2D dectector} compares using Co-DINO vs. other methods as the 2D detector,
applied along with the 3D LiDAR detector CenterPoint in our MMLF.
Results show that using Co-DINO, a stronger detector than others, yields significantly better 3D detection performance on all classes, particularly on {\tt medium} and {\tt few} classes.

\section{Analysis on Transformer 3D Detectors}
\label{sec:transformer-based}

{
\setlength{\tabcolsep}{1.6mm}
\begin{table}[t]
\small
\centering
\scalebox{0.8}{
\begin{tabular}{lccccccccccccccccccccccccccccc}
\toprule 
\multirow{1}{*}{Method} & \multicolumn{1}{c}{\tt All} & \multicolumn{1}{c}{{\tt Many}} & \multicolumn{1}{c}{{\tt Medium}} & \multicolumn{1}{c}{{\tt Few}} 
 \\
\midrule

\multirow{1}{*}{CenterPoint {\em w/ Hier.}}   
  & \multirow{1}{*}{40.4} & \multirow{1}{*}{77.1} & \multirow{1}{*}{45.1} & \multirow{1}{*}{4.3} \\ 

 \multirow{1}{*}{MMLF(CenterPoint, Co-DINO)}   
 & \multirow{1}{*}{57.0} & \multirow{1}{*}{80.7} & \multirow{1}{*}{\bf 64.0} & \multirow{1}{*}{\textbf{29.2}}  \\ \midrule

\multirow{1}{*}{BEVFusion-L}   
 & \multirow{1}{*}{42.5} & \multirow{1}{*}{72.5} & \multirow{1}{*}{48.0} & \multirow{1}{*}{10.6}  \\ 

 \multirow{1}{*}{MMLF(BEVFusion-L, Co-DINO)}   
 & \multirow{1}{*}{56.2} & \multirow{1}{*}{80.1} & \multirow{1}{*}{63.2} & \multirow{1}{*}{28.1}  \\ \midrule

\multirow{1}{*}{CMT-L}   
 & \multirow{1}{*}{34.7} & \multirow{1}{*}{73.4} & \multirow{1}{*}{35.9} & \multirow{1}{*}{1.1}  \\ 

  \multirow{1}{*}{MMLF(CMT-L, Co-DINO)}   
 & \multirow{1}{*}{{\bf 57.1}} & \multirow{1}{*}{\textbf{81.2}} & \multirow{1}{*}{63.8} & \multirow{1}{*}{29.2}  \\ 
 
 \midrule
 
 \multirow{1}{*}{BEVFusion (RGB + LiDAR)}   
 & \multirow{1}{*}{45.5}  & \multirow{1}{*}{75.5} & \multirow{1}{*}{52.0} & \multirow{1}{*}{12.8} \\ 

  \multirow{1}{*}{CMT (RGB + LiDAR)}   
 & \multirow{1}{*}{44.4}  & \multirow{1}{*}{79.9} & \multirow{1}{*}{53.0} & \multirow{1}{*}{4.8} \\ 

 \bottomrule
 
\end{tabular}
}
\caption{\small
{\bf Transformer-based 3D LiDAR detector}. We evaluate the late-fusion performance of BEVFusion-L and CMT-L (LiDAR-only) with Co-DINO, and find that our late fusion strategy improves over BEVFusion-L by 13.7 mAP overall and CMT-L by 22.4 mAP. This suggests that our late-fusion approach generalizes across different 3D detector architectures.
}
\label{tab:bevfusion} 
\end{table}
}

In the main paper, we evaluate all late-fusion methods by filtering LiDAR detections from CenterPoint with different RGB detectors for fair comparison to \citep{peri2022towards}. However, our late-fusion approach generalizes to other 3D LiDAR detectors as well. To this end, we evaluate our best-performing late-fusion configuration from the main paper with LiDAR detections from BEVFusion-L \citep{liu2022bevfusion} and CMT-L \citep{yan2023cross} in Table \ref{tab:bevfusion}.

First, we note that BEVFusion-L performs 2.1 mAP better than CenterPoint on LT3D, notably improving on classes with {\tt few} examples by 6.3 mAP. However, BEVFusion-L performs 4.6 mAP worse than CenterPoint on classes with {\tt many} examples, suggesting that learning robust features for both {\tt common} and {\tt rare} categories is challenging. Surprisingly, CMT-L performs worse than CenterPoint on all classes. 
We find that filtering BEVFusion-L's 3D detections with Co-DINO's 2D detections yields a further 13.7 mAP improvement overall, and a 17.5 mAP improvement for rare classes. This further supports our hypothesis that late-fusion with better 2D detectors can improve LT3D accuracy. In contrast, the end-to-end multi-modal variant of BEVFusion only improves over the LiDAR-only baseline by 3 mAP overall, and performs 10.7 mAP worse than our BEVFusion-L late-fusion variant.
Similarly, filtering CMT-L's 3D detections with Co-DINO's 2D detections yields a 22.4 mAP improvement overall, with a 28.1 mAP improvement for rare classes. Fusing CMT-L with Co-DINO performs marginally better than CenterPoint with Co-DINO.

We posit that the relative improvement from late-fusion can be attributed more generally to LiDAR-based detectors achieving high recall and low precision rather than the specific network architecture. Our late-fusion approach will be less effective for LiDAR-based detectors that achieve high precision and low recall since our method does not add new 3D detections (by lifting 2D detections), but rather only filters existing ones.

\section{Per-Class Breakdown Results}
\label{sec:per-class-results}

{
\setlength{\tabcolsep}{2.5mm}
\begin{table*}[t] 
\small
\centering
\scalebox{0.86}{
\begin{tabular}{lcccc}
\toprule
Class name & BevFormer   & Deepinteraction  & CMT  & {\bf MMLF}(CenterPoint, Co-DINO) \\
&  \citep{li2022bevformer}  & \citep{yang2022deepinteraction} & \citep{yan2023cross} & (Ours) \\
\midrule
\rowcolor{col11}Personal Mobility & 0.7 & 2.5 & 4.1 & 42.3 \\
\rowcolor{col11}Emergency Vehicle & 0.2 & 1.5 & 6.9 & 28.0 \\
\rowcolor{col11}Police Officer & 0.9 & 0.0 & 7.3 & 16.1 \\
\rowcolor{col11}Stroller & 1.1 & 30.9 & 0.9 & 45.4 \\
\rowcolor{col11}Child & 3.9 & 6.0 & 4.7 & 32.1 \\
\rowcolor{col11}Debris & 1.9 & 6.8 & 4.6 & 11.4 \\
\rowcolor{yellow}Construction Worker & 21.5 & 30.7 & 34.4 & 63.9 \\
\rowcolor{yellow}Bicycle & 41.9 & 63.3 & 66.7 & 77.3 \\
\rowcolor{yellow}Motorcycle & 47.0 & 76.2 & 76.3 & 80.7 \\
\rowcolor{yellow}Construction Vehicle & 16.9 & 35.3 & 36.9 & 40.0 \\
\rowcolor{yellow}Bus & 43.0 & 75.9 & 76.4 & 79.2 \\
\rowcolor{yellow}Pushable Pullable & 29.1 & 30.8 & 34.1 & 57.2 \\
\rowcolor{yellow}Trailer & 21.4 & 45.4 & 46.5 & 49.4 \\
 \cellcolor{col3}Truck &  \cellcolor{col3}37.1 &  \cellcolor{col3}63.2 & \cellcolor{col3}65.2 &  \cellcolor{col3}68.1 \\
 \cellcolor{col3}Traffic Cone &  \cellcolor{col3}59.9 &  \cellcolor{col3}71.8 &  \cellcolor{col3}80.1 &  \cellcolor{col3}82.3 \\
 \cellcolor{col3}Barrier &  \cellcolor{col3}53.2 &  \cellcolor{col3}75.3 &  \cellcolor{col3}77.9 &  \cellcolor{col3}75.6 \\
 \cellcolor{col3}Adult &  \cellcolor{col3}50.3 &  \cellcolor{col3}85.9 &  \cellcolor{col3}87.7 &  \cellcolor{col3}89.2 \\
 \cellcolor{col3}Car &  \cellcolor{col3}60.8 &  \cellcolor{col3}84.9 &  \cellcolor{col3}88.6 &  \cellcolor{col3}88.5 \\
\midrule
\rowcolor{col11}Few & 1.4 & 7.9 & 4.8 & 29.2 \\
\rowcolor{yellow}Medium & 31.6 &  51.1 & 53.0 & 64.0\\
 \cellcolor{col3}Many &  \cellcolor{col3}52.3 &  \cellcolor{col3}76.2 &  \cellcolor{col3}79.9 &  \cellcolor{col3}80.7 \\
\midrule
All & 27.3 & 43.7 & 44.4 & 57.0 \\
\bottomrule
\end{tabular}
}
\caption{\small
{\bf Per-class performance on nuScenes.} 
Our approach MMLF fuses detections from the LiDAR-based 3D detector CenterPoint and the RGB-based 2D detector Co-DINO, achieving the highest per-class AP on 16 out of 18 classes. 
It is worth noting that, compared to CMT, our approach improves {\tt construction worker} by 29.5 AP, 
{\tt stroller} by 44.5 AP, and {\tt pushable-pullable} by 23.1 AP. 
}
\vspace{-1mm}
\label{tab:breakdown} 
\end{table*}
}

We compare recent multi-modal methods w.r.t per-class AP in Table \ref{tab:breakdown}. 
All multi-modal methods perform similarly on {\tt common} classes but  considerably worse on {\tt Medium} and {\tt Few} classes, highlighting the need for further investigation of LT3D by the research community. Notably, our late-fusion approach achieves 23.1 higher AP on {\tt pushable-pullable} and 14.5 higher AP on {\tt stroller} than prior work. In general, our late-fusion approach yields considerable improvement on classes with {\tt medium} and {\tt few} examples. Despite significant improvements in rare class detection accuracy, our approach detects {\tt child} with 32.1 AP, which remains relatively low compared with other classes.
We posit that it is difficult to differentiate {\tt child} from {\tt adult} due to perspective geometry, since a small child close to the camera looks similar to a tall adult far away from the camera.

\section{Analysis of AP vs. NDS Results}
\label{sec:NDS}

{
\setlength{\tabcolsep}{0.21em}
\begin{table*}[t] 
\centering
\scalebox{0.70}{
\begin{tabular}{lccccccccc}
\toprule
& FCOS3D  
& BEVFormer
& CP 
& TransFusion 
& DeepInteraction 
& CMT
& {\bf MMLF(CP, Co-DINO)} \\
& \citep{wang2021fcos3d} 
& \citep{li2022bevformer}
& \citep{yin2021center}
& \citep{bai2022transfusion}
& \citep{yang2022deepinteraction}
& \citep{yan2023cross}
& {\bf (Ours)}\\
\midrule
mAP &  20.9 
& 27.3
& 39.2 
& 39.8 
& 43.7
& 44.4
& 57.0 \\
NDS &30.4 
& 38.8
& 54.9 
& 53.9  
& 54.4
& 55.9
& 62.8 \\
\hline
\end{tabular}
}
\caption{\small
{\bf AP vs. NDS results}. We compare methods reported in the main paper w.r.t nuScenes Detection Score (NDS) and mAP. We find that the methods show broadly consistent rankings on both metrics, with MMLF ranking first under both mAP and NDS.
}
\label{tab:nds} 
\end{table*}
}

We report nuScenes Detection Score (NDS) and mAP results in Table~\ref{tab:nds}. We find that the methods show broadly consistent rankings on both metrics, with MMLF ranking first under both mAP and NDS. This is unsurprising because NDS is computed as a weighted sum of mAP and other true positive metrics, where mAP is weighted five times greater than other components.

\section{Impact of Occlusion and Range}
\label{sec:analysis-occlusion}

We report average recall at four visibility levels in Table \ref{tab:visibility}.
Results show that detecting occluded and partially occluded objects is significantly more challenging than detecting un-occluded objects, likely because the visible regions of partially occluded objects can be easily mistaken as part of the background and hence lead to more false positive detections. Interestingly, the LiDAR-based 3D detector CenterPoint achieves the highest recall across all occlusion levels.

We also examine detection accuracy across three distance ranges in Table \ref{tab:distance} and observe that detecting objects closer to the ego-vehicle is easier due to the higher LiDAR point density. While CenterPoint's performance declines significantly beyond 10m, CMT and MMLF demonstrate greater robustness, suggesting that both feature-fusion and late-fusion effectively mitigate the impact of LiDAR sparsity at range. Furthermore, our multi-modal fusion approach significantly improves CenterPoint’s detection accuracy in the 10 -- 20m and 20 -- 30m ranges by reducing false positives, increasing AP from 40.1 to 61.5 and from 26.7 to 39.3, respectively.

{
\setlength{\tabcolsep}{0.2em} 
\begin{table}[t]
\small
\centering
\scalebox{0.75}{
\begin{tabular}{l c c c c c c c c c c c}
\toprule
\multirow{1}{*}{Method} & \multirow{1}{*}{Modality} & \multicolumn{1}{c}{0-40\%}  & \multicolumn{1}{c}{40-60\%} & \multicolumn{1}{c}{60-80\%} & \multicolumn{1}{c}{80-100\%} \\ 

\midrule
\multirow{1}{*}{CenterPoint (CP)}  & {\multirow{1}{*}{L} } & {\multirow{1}{*}{84.1}} & {\multirow{1}{*}{90.9}} & {\multirow{1}{*}{92.8}} & {\multirow{1}{*}{96.2}}  \\
\midrule

\multirow{1}{*}{CMT}  & {\multirow{1}{*}{C + L} } & {\multirow{1}{*}{84.3}} & {\multirow{1}{*}{91.6}} & {\multirow{1}{*}{93.1}} & {\multirow{1}{*}{96.1}}  \\
\midrule

{\textbf{MMLF(CP, Co-DINO) }} 
& {\multirow{1}{*}{C + L} } & {\multirow{1}{*}{83.2}} & {\multirow{1}{*}{90.4}} & {\multirow{1}{*}{92.5}} & {\multirow{1}{*}{95.9}}  \\
\bottomrule
\end{tabular}
}
\caption{\small 
{\bf Average recall for occluded targets at different visibility levels}. We evaluate average recall at four visibility levels (defined by nuScenes). Perhaps unsurprisingly, detection accuracy drops as objects become more occluded. Note that visibility attributes are not predicted by our model and are manually annotated for the ground truth, making it difficult to measure precision.
}
\label{tab:visibility}
\end{table}
}

{
\setlength{\tabcolsep}{0.15em} 
\begin{table}[t]
\small
\centering
\scalebox{0.74}{
\begin{tabular}{l c c c c c c c c c c c}
\toprule
\multirow{1}{*}{Method} & \multirow{1}{*}{Modality} & \multicolumn{1}{c}{0-10m}  & \multicolumn{1}{c}{10-20m} & \multicolumn{1}{c}{20-30m} & \multicolumn{1}{c}{Full Range} \\ 

\midrule
\multirow{1}{*}{CenterPoint (CP)}  & {\multirow{1}{*}{L} } & {\multirow{1}{*}{45.5}} & {\multirow{1}{*}{40.1}} & {\multirow{1}{*}{26.7}} & {\multirow{1}{*}{40.4}}  \\
\midrule

\multirow{1}{*}{CMT}  & {\multirow{1}{*}{C + L} } & {\multirow{1}{*}{42.0}} & {\multirow{1}{*}{43.2}} & {\multirow{1}{*}{31.4}} & {\multirow{1}{*}{44.4}}  \\
\midrule

{\textbf{MMLF(CP, Co-DINO)}} 
& {\multirow{1}{*}{C + L} } & {\multirow{1}{*}{57.7}} & {\multirow{1}{*}{61.5}} & {\multirow{1}{*}{39.3}} & {\multirow{1}{*}{57.0}}  \\
\bottomrule
\end{tabular}
}
\caption{\small 
{\bf Detection accuracy at different distance thresholds}. We evaluate detection accuracy at three distance ranges. Detecting objects closer to the ego-vehicle is easier (due to higher LiDAR point density) than detecting objects further away.
}
\label{tab:distance}
\end{table}
}

\section{Analysis of Recall}
\label{sec:analysis-lidar-recall}

{
\setlength{\tabcolsep}{0.3em} 
\begin{table}[t]
\small
\centering
\scalebox{0.80}{
\begin{tabular}{l c c c c c c c c c c c}
\toprule
\multirow{1}{*}{Method} & \multirow{1}{*}{Modality} & \multicolumn{1}{c}{{\tt Vehicle}}  & \multicolumn{1}{c}{{\tt Pedestrian}} & \multicolumn{1}{c}{{\tt Movable}} & \multicolumn{1}{c}{{\tt Object}} \\

\midrule
\multirow{1}{*}{CenterPoint }  & {\multirow{1}{*}{L} } & {\multirow{1}{*}{85.5}} & {\multirow{1}{*}{95.5}} & {\multirow{1}{*}{90.7}} & {\multirow{1}{*}{89.3}}  \\
\midrule

\multirow{1}{*}{CMT}  & {\multirow{1}{*}{C + L} } & {\multirow{1}{*}{86.4}} & {\multirow{1}{*}{91.2}} & {\multirow{1}{*}{87.2}} & {\multirow{1}{*}{87.9}}  \\
\bottomrule
\end{tabular}
}
\caption{\small 
{\bf Average recall between detectors}. CenterPoint (LiDAR-only) achieves higher recall than CMT for the {\tt Pedestrian} (LCA=1), {\tt Movable} (LCA=1), and {\tt Object} (LCA=2) superclasses.
}
\label{tab:recall}
\end{table}
}

As shown in Table \ref{tab:recall}, the LiDAR-only 3D detector CenterPoint achieves higher recall than CMT (multi-modal) for the {\tt Pedestrian} (LCA=1), {\tt Movable} (LCA=1), and {\tt Object} (LCA=2) superclasses. The lower recall of CMT can be attributed to its limited number of object queries (N=300) per frame compared to the (nearly) unbounded number of predictions from anchor-free detectors like CenterPoint. Importantly, the high recall of CenterPoint is essential for our late-fusion approaches (cf. Fig.~\ref{fig:filtering}).

\section{Impact of Calibration Noise}
\label{sec:analysis-calibration-noise}

Following prior literature, we benchmark on datasets (nuScenes and Argoverse 2) collected in-the-wild, which {\em already} contain realistic calibration errors.  However, we study the impact of introducing more aggressive calibration errors in Table \ref{tab:calibration_noise} by applying synthetic perturbations to the camera-to-LiDAR translation and rotation matrices.  We apply additive Gaussian noise $N_T$ to each of the $X$, $Y$ and $Z$ components of the translation vector and apply additive noise $N_R$ to the yaw, pitch and roll of the rotation matrix. We find that translation noise does not significantly affect detection accuracy. In contrast, rotation noise has a much larger impact on accuracy. Unsurprisingly, combining translation and rotation noise yields significantly degraded accuracy. Such calibration errors affect all (early, mid, and late) fusion methods since they require calibrated extrinsics \citep{philion2020lift}. We find that our late-fusion approach is less robust to noise than end-to-end fusion methods. However, our approach still achieves higher accuracy than CMT.

{
\setlength{\tabcolsep}{0.5em} 
\begin{table}[t]
\small
\centering
\scalebox{0.84}{
\begin{tabular}{l c c c c c c c c c c c}
\toprule
\multirow{1}{*}{Translation Noise $N_T$}  & \multicolumn{1}{c}{0.01m}  & \multicolumn{1}{c}{0.02m} & \multicolumn{1}{c}{0.03m} & \multicolumn{1}{c}{0.04m} \\

\midrule

\multirow{1}{*}{CMT}   & {\multirow{1}{*}{44.4}} & {\multirow{1}{*}{44.3}} & {\multirow{1}{*}{44.3}} & {\multirow{1}{*}{44.2}}  \\
{\textbf{MMLF(CP, Co-DINO)}} 
& {\multirow{1}{*}{57.0}} & {\multirow{1}{*}{56.9}} & {\multirow{1}{*}{56.8}} & {\multirow{1}{*}{56.8}}  \\

\midrule
\midrule

\multirow{1}{*}{Rotation Noise $N_R$}  
& \multicolumn{1}{c}{1$^\circ$}  & \multicolumn{1}{c}{2$^\circ$} & \multicolumn{1}{c}{3$^\circ$} & \multicolumn{1}{c}{4$^\circ$} \\

\midrule

\multirow{1}{*}{CMT}   & {\multirow{1}{*}{44.3}} & {\multirow{1}{*}{43.1}} & {\multirow{1}{*}{42.7}} & {\multirow{1}{*}{41.1}}  \\

{\textbf{MMLF(CP,  Co-DINO)}} 
& {\multirow{1}{*}{54.3}} & {\multirow{1}{*}{53.4}} & {\multirow{1}{*}{51.7}} & {\multirow{1}{*}{51.3}}  \\

\midrule
\midrule

\multirow{1}{*}{Translation Noise $N_T$}  & \multicolumn{1}{c}{0.01m}  & \multicolumn{1}{c}{0.02m} & \multicolumn{1}{c}{0.03m} & \multicolumn{1}{c}{0.04m} \\ 

\multirow{1}{*}{Rotation Noise $N_R$}  & \multicolumn{1}{c}{1$^\circ$}  & \multicolumn{1}{c}{2$^\circ$} & \multicolumn{1}{c}{3$^\circ$} & \multicolumn{1}{c}{4$^\circ$} \\

\midrule

\multirow{1}{*}{CMT}   & {\multirow{1}{*}{44.3}} & {\multirow{1}{*}{43.8}} & {\multirow{1}{*}{43.1}} & {\multirow{1}{*}{42.2}}  \\
{\textbf{MMLF(CP, Co-DINO)}} 
& {\multirow{1}{*}{53.4}} & {\multirow{1}{*}{52.2}} & {\multirow{1}{*}{50.7}} & {\multirow{1}{*}{50.2}}  \\
\bottomrule
\end{tabular}
}
\caption{\small 
{\bf Impact of calibration noise}. We apply a small amount of additive Gaussian noise to the translation and rotation calibration matrices to measure the impact of imprecise calibration on late fusion. While translation noise has little effect on detection accuracy, rotation noise significantly impacts performance. As expected, combining both translation and rotation noise degrades accuracy further.
}
\label{tab:calibration_noise}
\end{table}
}

\newpage

\section{Per-Class Breakdown Results w.r.t Distance and LiDAR Sparsity} 
\label{sec:more_breakdown}

Table~\ref{tab:multi_class_results} supplements Table~\ref{tab:selected_stratified_results} of the main paper with per-class breakdown results w.r.t LiDAR density, distance, and cardinality.

\begin{table*}[htbp]
    \centering
    \begin{minipage}[t]{0.32\textwidth}
        \centering
        \small\textbf{(a) Car} \\
        \vspace{1.5mm}
        \scalebox{0.7}{
        \begin{tabular}{lcccc}
            \toprule
            Methods & near & far & sparse & dense \\
            \midrule
            CenterPoint & 93.7 & 65.9 & 70.5 & 96.5 \\
            FSHNet      & \textbf{94.1} & 67.8 & 72.5 & \textbf{96.9} \\
            CMT         & 93.6 & 69.8 & 73.6 & 96.2 \\
            CCF         & 93.8 & 70.2 & \textbf{73.8} & 96.4 \\
            MMLF        & 93.5 & \textbf{71.4} & 73.7 & 96.1 \\
            \bottomrule
        \end{tabular}}
    \end{minipage}
    \hfill
    \begin{minipage}[t]{0.32\textwidth}
        \centering
        \small\textbf{(b) Truck} \\
        \vspace{1.5mm}
        \scalebox{0.7}{
        \begin{tabular}{lcccc}
            \toprule
            Methods & near & far & sparse & dense \\
            \midrule
            CenterPoint & 65.6 & 35.6 & 30.1 & 71.8 \\
            FSHNet      & 68.5 & 38.5 & 34.0 & 74.4 \\
            CMT         & 65.8 & 42.5 & 39.1 & 69.9 \\
            CCF         & 67.2 & 42.9 & 39.5 & 71.1 \\
            MMLF        & \textbf{71.2} & \textbf{45.2} & \textbf{40.1} & \textbf{76.1} \\
            \bottomrule
        \end{tabular}}
    \end{minipage}
    \hfill
    \begin{minipage}[t]{0.32\textwidth}
        \centering
        \small\textbf{(c) Trailer} \\
        \vspace{1.5mm}
        \scalebox{0.7}{
        \begin{tabular}{lcccc}
            \toprule
            Methods & near & far & sparse & dense \\
            \midrule
            CenterPoint & 53.2 & 23.0 & 13.7 & 60.8 \\
            FSHNet      & 55.0 & 25.1 & 16.4 & 63.2 \\
            CMT         & 49.6 & 28.9 & 18.7 & 58.8 \\
            CCF         & 52.1 & 28.8 & 18.6 & 61.3 \\
            MMLF        & \textbf{56.9} & \textbf{29.7} & \textbf{19.7} & \textbf{64.4} \\
            \bottomrule
        \end{tabular}}
    \end{minipage}

    \vspace{4mm} 

    \begin{minipage}[t]{0.32\textwidth}
        \centering
        \small\textbf{(d) Bus} \\
        \vspace{1.5mm}
        \scalebox{0.7}{
        \begin{tabular}{lcccc}
            \toprule
            Methods & near & far & sparse & dense \\
            \midrule
            CenterPoint & 81.0 & 54.9 & 52.4 & 82.9 \\
            FSHNet      & 83.2 & 57.8 & 56.1 & 84.9 \\
            CMT         & 81.8 & 61.3 & 58.9 & 83.6 \\
            CCF         & 83.5 & 62.3 & 59.3 & 85.6 \\
            MMLF        & \textbf{86.4} & \textbf{64.2} & \textbf{62.3} & \textbf{87.7} \\
            \bottomrule
        \end{tabular}}
    \end{minipage}
    \hfill
    \begin{minipage}[t]{0.32\textwidth}
        \centering
        \small\textbf{(e) Construction vehicle} \\
        \vspace{1.5mm}
        \scalebox{0.7}{
        \begin{tabular}{lcccc}
            \toprule
            Methods & near & far & sparse & dense \\
            \midrule
            CenterPoint & 34.5 & 8.9 & 9.7 & 34.4 \\
            FSHNet      & 39.0 & 14.1 & 13.8 & 39.2 \\
            CMT         & 41.5 & 19.6 & 18.0 & 43.7 \\
            CCF         & 42.4 & 20.1 & 19.2 & 44.4 \\
            MMLF        & \textbf{44.2} & \textbf{21.5} & \textbf{20.6} & \textbf{46.0} \\
            \bottomrule
        \end{tabular}}
    \end{minipage}
    \hfill
    \begin{minipage}[t]{0.32\textwidth}
        \centering
        \small\textbf{(f) Bicycle} \\
        \vspace{1.5mm}
        \scalebox{0.7}{
        \begin{tabular}{lcccc}
            \toprule
            Methods & near & far & sparse & dense \\
            \midrule
            CenterPoint & 54.7 & 3.4 & 16.4 & 67.6 \\
            FSHNet      & 62.1 & 7.8 & 24.3 & 72.1 \\
            CMT         & 70.0 & 19.6 & 43.2 & 76.7 \\
            CCF         & 72.2 & 19.7 & 44.7 & 78.2 \\
            MMLF        & \textbf{81.3} & \textbf{26.5} & \textbf{58.2} & \textbf{88.4} \\
            \bottomrule
        \end{tabular}}
    \end{minipage}

    \vspace{4mm}

    \begin{minipage}[t]{0.32\textwidth}
        \centering
        \small\textbf{(g) Motorcycle} \\
        \vspace{1.5mm}
        \scalebox{0.7}{
        \begin{tabular}{lcccc}
            \toprule
            Methods & near & far & sparse & dense \\
            \midrule
            CenterPoint & 73.8 & 15.2 & 31.5 & 86.0 \\
            FSHNet      & 78.1 & 23.9 & 40.4 & 88.1 \\
            CMT         & 80.3 & 42.8 & 53.9 & 89.8 \\
            CCF         & 81.4 & 43.1 & 55.2 & 91.3 \\
            MMLF        & \textbf{85.3} & \textbf{49.2} & \textbf{60.3} & \textbf{95.1} \\
            \bottomrule
        \end{tabular}}
    \end{minipage}
    \hfill
    \begin{minipage}[t]{0.32\textwidth}
        \centering
        \small\textbf{(h) Emergency vehicle} \\
        \vspace{1.5mm}
        \scalebox{0.7}{
        \begin{tabular}{lcccc}
            \toprule
            Methods & near & far & sparse & dense \\
            \midrule
            CenterPoint & 18.9 & 0.8 & 1.7 & 21.5 \\
            FSHNet      & 20.2 & 1.1 & 2.2 & 23.2 \\
            CMT         & 5.7 & \textbf{4.5} & \textbf{3.2} & 6.6 \\
            CCF         & 15.3 & 4.4 & \textbf{3.2} & 18.1 \\
            MMLF        & \textbf{43.1} & 0.2 & 0.8 & \textbf{51.2} \\
            \bottomrule
        \end{tabular}}
    \end{minipage}
    \hfill
    \begin{minipage}[t]{0.32\textwidth}
        \centering
        \small\textbf{(i) Adult} \\
        \vspace{1.5mm}
        \scalebox{0.7}{
        \begin{tabular}{lcccc}
            \toprule
            Methods & near & far & sparse & dense \\
            \midrule
            CenterPoint & 87.4 & 57.8 & 74.5 & 89.6 \\
            FSHNet      & 88.3 & 59.3 & 76.1 & 90.8 \\
            CMT         & 88.6 & 59.7 & 76.1 & 91.3 \\
            CCF         & 88.8 & 60.8 & 77.1 & 91.7 \\
            MMLF        & \textbf{90.2} & \textbf{63.9} & \textbf{80.1} & \textbf{93.3} \\
            \bottomrule
        \end{tabular}}
    \end{minipage}

    \vspace{4mm}

    \begin{minipage}[t]{0.32\textwidth}
        \centering
        \small\textbf{(j) Child} \\
        \vspace{1.5mm}
        \scalebox{0.7}{
        \begin{tabular}{lcccc}
            \toprule
            Methods & near & far & sparse & dense \\
            \midrule
            CenterPoint & 1.2 & 0.0 & 0.3 & 1.5 \\
            FSHNet      & 1.4 & 0.0 & 0.4 & 1.8 \\
            CMT         & 5.3 & 0.2 & 0.8 & 8.2 \\
            CCF         & 8.1 & 0.4 & 2.1 & 10.2 \\
            MMLF        & \textbf{37.0} & \textbf{1.5} & \textbf{14.0} & \textbf{43.9} \\
            \bottomrule
        \end{tabular}}
    \end{minipage}
    \hfill
    \begin{minipage}[t]{0.32\textwidth}
        \centering
        \small\textbf{(k) Police officer} \\
        \vspace{1.5mm}
        \scalebox{0.7}{
        \begin{tabular}{lcccc}
            \toprule
            Methods & near & far & sparse & dense \\
            \midrule
            CenterPoint & 0.3 & 0.0 & 0.0 & 0.4 \\
            FSHNet      & 0.5 & 0.0 & 0.0 & 0.8 \\
            CMT         & 8.8 & 0.0 & 0.0 & 15.4 \\
            CCF         & 10.1 & 0.0 & 0.0 & 18.3 \\
            MMLF        & \textbf{18.7} & 0.0 & 0.0 & \textbf{31.9} \\
            \bottomrule
        \end{tabular}}
    \end{minipage}
    \hfill
    \begin{minipage}[t]{0.32\textwidth}
        \centering
        \small\textbf{(l) Construction worker} \\
        \vspace{1.5mm}
        \scalebox{0.7}{
        \begin{tabular}{lcccc}
            \toprule
            Methods & near & far & sparse & dense \\
            \midrule
            CenterPoint & 31.9 & 1.4 & 11.8 & 34.0 \\
            FSHNet      & 36.2 & 2.8 & 15.1 & 38.2 \\
            CMT         & 33.8 & 6.3 & 19.5 & 32.5 \\
            CCF         & 38.1 & 7.6 & 21.8 & 36.3 \\
            MMLF        & \textbf{65.6} & \textbf{15.8} & \textbf{40.9} & \textbf{69.5} \\
            \bottomrule
        \end{tabular}}
    \end{minipage}

    \vspace{4mm}

    \begin{minipage}[t]{0.32\textwidth}
        \centering
        \small\textbf{(m) Stroller} \\
        \vspace{1.5mm}
        \scalebox{0.7}{
        \begin{tabular}{lcccc}
            \toprule
            Methods & near & far & sparse & dense \\
            \midrule
            CenterPoint & 4.3 & 1.3 & 2.3 & 4.2 \\
            FSHNet      & 3.4 & 1.1 & 1.9 & 3.1 \\
            CMT         & 0.8 & 0.3 & 0.4 & 0.6 \\
            CCF         & 8.1 & 1.9 & 3.7 & 9.9 \\
            MMLF        & \textbf{42.0} & \textbf{40.5} & \textbf{32.7} & \textbf{53.4} \\
            \bottomrule
        \end{tabular}}
    \end{minipage}
    \hfill
    \begin{minipage}[t]{0.32\textwidth}
        \centering
        \small\textbf{(n) Personal mobility} \\
        \vspace{1.5mm}
        \scalebox{0.7}{
        \begin{tabular}{lcccc}
            \toprule
            Methods & near & far & sparse & dense \\
            \midrule
            CenterPoint & 0.5 & 0.0 & 0.3 & 0.4 \\
            FSHNet      & 0.6 & 0.0 & 0.3 & 0.5 \\
            CMT         & 4.8 & 0.0 & 3.7 & 1.4 \\
            CCF         & 6.1 & 0.0 & 3.9 & 2.1 \\
            MMLF        & \textbf{48.9} & 0.0 & \textbf{41.3} & \textbf{74.0} \\
            \bottomrule
        \end{tabular}}
    \end{minipage}
    \hfill
    \begin{minipage}[t]{0.32\textwidth}
        \centering
        \small\textbf{(o) Pushable pullable} \\
        \vspace{1.5mm}
        \scalebox{0.7}{
        \begin{tabular}{lcccc}
            \toprule
            Methods & near & far & sparse & dense \\
            \midrule
            CenterPoint & 36.2 & 2.0 & 13.3 & 37.9 \\
            FSHNet      & 40.2 & 2.9 & 16.1 & 42.1 \\
            CMT         & 35.6 & 1.7 & 13.3 & 37.3 \\
            CCF         & 40.2 & 2.2 & 15.2 & 42.3 \\
            MMLF        & \textbf{59.7} & \textbf{4.8} & \textbf{30.6} & \textbf{65.3} \\
            \bottomrule
        \end{tabular}}
    \end{minipage}

    \vspace{4mm}

    \begin{minipage}[t]{0.32\textwidth}
        \centering
        \small\textbf{(p) Debris} \\
        \vspace{1.5mm}
        \scalebox{0.70}{
        \begin{tabular}{lcccc}
            \toprule
            Methods & near & far & sparse & dense \\
            \midrule
            CenterPoint & 6.6 & N/A & 1.7 & 6.9 \\
            FSHNet      & 6.1 & N/A & 1.4 & 6.5 \\
            CMT         & 4.6 & N/A & 1.4 & 4.3 \\
            CCF         & 6.3 & N/A & 2.1 & 5.9 \\
            MMLF        & \textbf{11.4} & N/A & \textbf{5.8} & \textbf{7.6} \\
            \bottomrule
        \end{tabular}}
    \end{minipage}
    \hfill
    \begin{minipage}[t]{0.32\textwidth}
        \centering
        \small\textbf{(q) Traffic cone} \\
        \vspace{1.5mm}
        \scalebox{0.70}{
        \begin{tabular}{lcccc}
            \toprule
            Methods & near & far & sparse & dense \\
            \midrule
            CenterPoint & 75.7 & N/A & 56.3 & 83.7 \\
            FSHNet      & 78.7 & N/A & 61.1 & 85.1 \\
            CMT         & 80.1 & N/A & 63.8 & 84.8 \\
            CCF         & 80.8 & N/A & 64.8 & 85.8 \\
            MMLF        & \textbf{82.3} & N/A & \textbf{66.8} & \textbf{86.7} \\
            \bottomrule
        \end{tabular}}
    \end{minipage}
    \hfill
    \begin{minipage}[t]{0.32\textwidth}
        \centering
        \small\textbf{(r) Barrier} \\
        \vspace{1.5mm}
        \scalebox{0.70}{
        \begin{tabular}{lcccc}
            \toprule
            Methods & near & far & sparse & dense \\
            \midrule
            CenterPoint & 71.3 & N/A & 48.6 & 74.3 \\
            FSHNet      & 74.4 & N/A & 54.1 & 78.2 \\
            CMT         & 77.9 & N/A & 57.2 & 84.0 \\
            CCF         & \textbf{78.6} & N/A & \textbf{58.2} & \textbf{84.6} \\
            MMLF        & 75.6 & N/A & 52.8 & 79.9 \\
            \bottomrule
        \end{tabular}}
    \end{minipage}
    \vspace{1mm}
    \caption{Detailed evaluation results of various methods across 18 obstacle categories.}
    \label{tab:multi_class_results}
\end{table*}

\section{Benchmarking on Argoverse 2}
\label{sec:av2}

{
\setlength{\tabcolsep}{0.3em} 
\begin{table}[t]
\small
\centering
\vspace{1mm}
\scalebox{0.81}{
\begin{tabular}{l c c c c c c c c c c c}
\toprule
\multirow{1}{*}{Method} & \multirow{1}{*}{Modality} & \multicolumn{1}{c}{\tt All}  & \multicolumn{1}{c}{\tt Many} & \multicolumn{1}{c}{\texttt{Medium}} & \multicolumn{1}{c}{\texttt{Few}} \\ 

\midrule
\multirow{1}{*}{FCOS3D}                &     C       &      14.6  &     27.4       &    17.0       &     7.8            \\ 
\multirow{1}{*}{DenseBEV}                &     C       &     26.1  &     35.8       &    28.9      &   19.6            \\ 
\midrule
\multirow{1}{*}{CenterPoint (CP)}         &     L      &     44.0  &    77.4        &       46.9    &   30.2                   \\ 
\multirow{1}{*}{FSHNet}         &     L      &     45.1  &   78.2        &       47.0    &  32.7                 \\ 
\midrule
\multirow{1}{*}{CCF}         &     C + L      &     49.9  &   81.8        &       52.0    &  37.6   \\

\midrule

{MMF(CP, FCOS3D)} 
& {\multirow{1}{*}{C + L} } & {\multirow{1}{*}{48.4}} & {\multirow{1}{*}{79.0}} & {\multirow{1}{*}{51.4}} & {\multirow{1}{*}{35.3}}  \\
{\textbf{MMLF(CP, Co-DINO)}} 
& {\multirow{1}{*}{C + L} } & {\multirow{1}{*}{\textbf{58.6}}} & {\multirow{1}{*}{\textbf{93.9}}} & {\multirow{1}{*}{\textbf{60.7}}} & {\multirow{1}{*}{\textbf{45.3}}}  \\
\bottomrule
\end{tabular}
}
\caption{\small 
{\bf Results on the Argoverse 2 dataset}. 
We present results on AV2 evaluated at 50m. 
The RGB-only 3D detector FCOS3D \citep{wang2021fcos3d} achieves poor performance, likely due to inaccurate depth estimates. In contrast, the LiDAR-only detector CenterPoint (CP)~\citep{yin2021center} achieves strong performance on all classes. Our multi-modal fusion approach significantly improves over CenterPoint, achieving a 14.6 mAP improvement averaged over all classes.
Importantly, late-fusion with 2D detector Co-DINO~\citep{zong2023detrs} performs best by a notable margin.
These results on AV2 are consistent with those on nuScenes, demonstrating the general applicability of our approach.
}
\vspace{-3mm}
\label{tab:av2}
\end{table}
}

\begin{figure}[t]
\centering
\small
\  \hspace{4mm} {\tt \scriptsize  Few} \hspace{10mm} {\tt \scriptsize  Medium} \hspace{10mm} {\tt \scriptsize Many}  \hspace{1mm}  \ \\
\includegraphics[width=0.95\linewidth, clip, trim={0cm 0cm 0cm 0cm}]{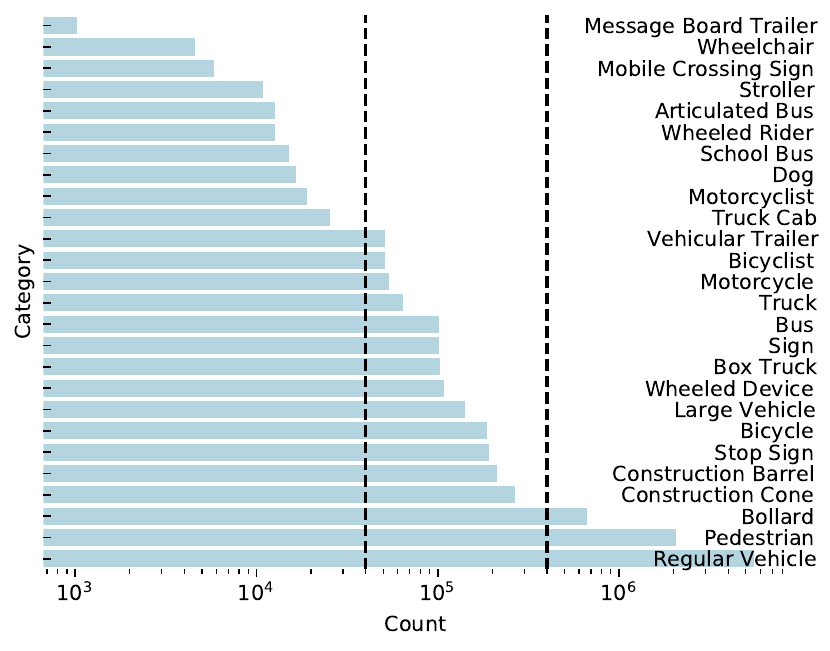} \\
\includegraphics[width=0.95\linewidth, clip, trim={4cm 0cm 3cm 0cm}]{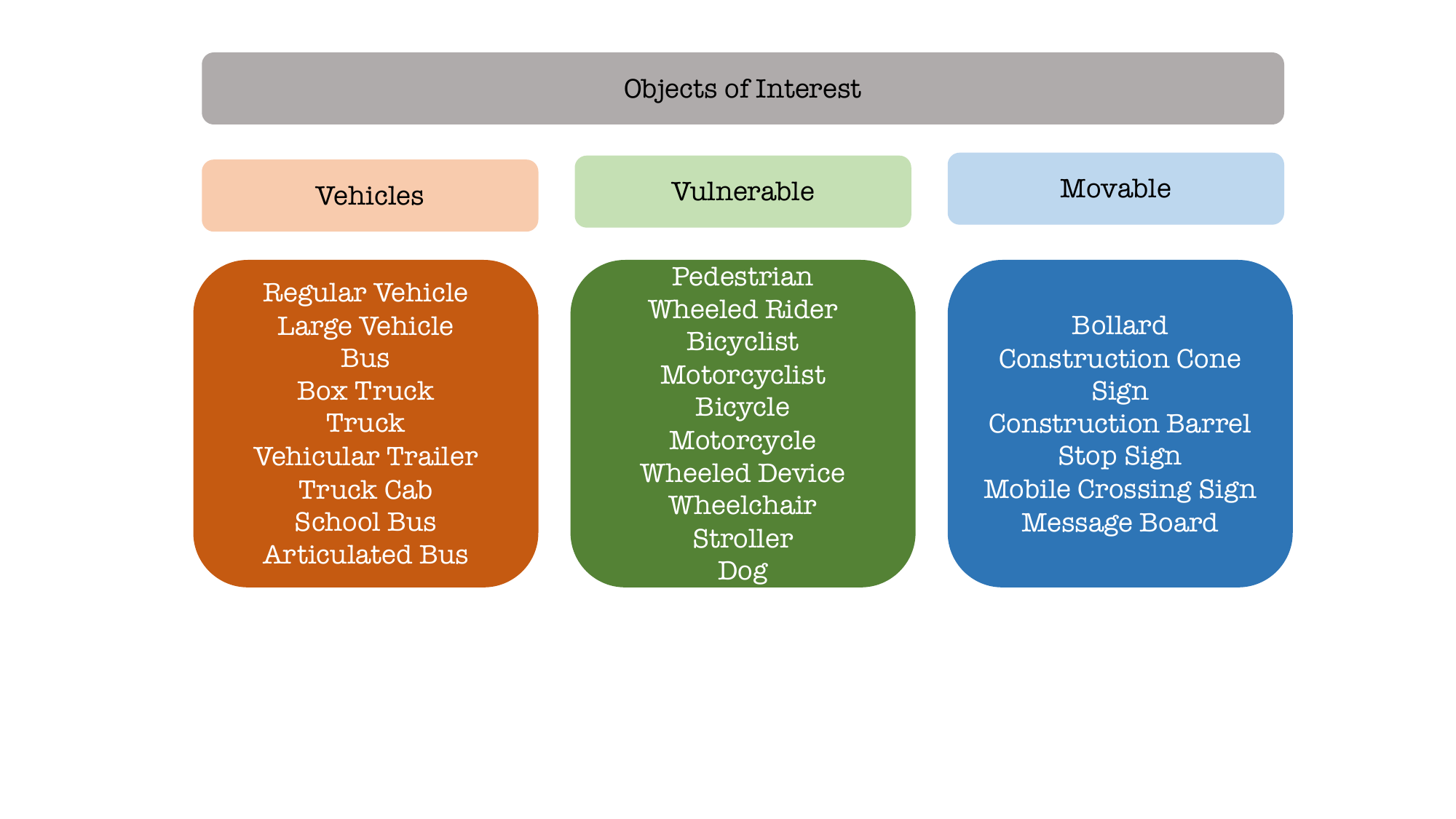} 
\vspace{-6mm}
\caption{\small
{\bf Left:}
According to the histogram of per-class object counts (on the \textbf{left}), 
classes in Argoverse 2.0 (AV2) follow a long tailed distribution. 
Following~\citep{openlongtailrecognition} and nuScenes, we report performance for three groups of classes based on their cardinality (split by dotted lines): {\tt Many}, {\tt Medium}, and {\tt Few}.
As AV2 does not provide a class hierarchy, we construct one  by referring to the nuScenes hierarchy (cf. Fig.~\ref{fig:hierarchy} on the \textbf{right).}
}
\vspace{-1mm}
\label{fig:av2_hierarchy}
\end{figure}

We present results on the large-scale Argoverse 2 (AV2) dataset developed for autonomous vehicle research.  AV2 evaluates on 26 classes, which follow a long-tailed distribution. 
In Fig.~\ref{fig:av2_hierarchy}, we plot the distribution on the left panel and show how we construct a semantic hierarchy on the right panel.
Following prior work \citep{peri2022towards}, 
we train on and evaluate detections up to 50m. As shown in Table~\ref{tab:av2}, our main conclusions from nuScenes still hold for AV2. FCOS3D \citep{wang2021fcos3d} yields poor performance on all classes, likely due to inaccurate depth estimates. CenterPoint performs considerably better, achieving high accuracy on classes with {\tt many} examples. Notably, CenterPoint performs better on AV2's rare classes (30.2 mAP) compared to nuScenes rare classes (3.5 mAP), likely because AV2 has more examples per class in the tail. Lastly, our proposed late-fusion approach improves over CenterPoint by 14.6 mAP and over prior work by 8.7 mAP. These new results on AV2 are consistent with those on nuScenes, demonstrating the general applicability of our approach.

\section{Standard Benchmarking on nuScenes}
\label{sec:standard}

{
\setlength{\tabcolsep}{0.65em}
\begin{table*}[t]
\small
\centering
\resizebox{\textwidth}{!}{
\begin{tabular}{l*{12}{c}}
\toprule
Method
& Car & Truck & Trailer & Bus & CV
& Bicy. & MC & Ped. & TC & Barrier & mAP & NDS \\
\midrule

DeepInteraction \citep{yang2022deepinteraction} (NeurIPS'22)
& 88.5 & 64.4 & 44.6 & 79.2 & 30.1
& 67.8 & 79.0 & 88.9 & 80.0 & 76.4 & 69.9 & 72.6 \\

FSHNet \citep{liu2025fshnet} (CVPR'25)
& 88.7 & 61.4 & 47.8 & 79.3 & 26.3
& 60.5 & 76.7 & 89.3 & 78.6 & 72.3 & 68.1 & 71.7 \\

DenseBEV \citep{dahling2026densebev} (WACV'26)
& 63.1 & 38.6 & 17.1 & 46.2 & 12.9
& 48.2 & 45.6 & 53.2 & 63.5 & 60.6 & 44.9 & 54.9 \\

\midrule

CenterPoint \citep{yin2021center} (CVPR'21)
& 87.7 & 61.6 & 43.7 & 73.4 & 28.4
& 49.5 & 65.9 & 87.1 & 75.7 & 71.3 & 66.6 & 64.4 \\

MMLF(CenterPoint, Co-DINO)
& 88.5 & 68.1 & 49.4 & 79.2 & 40.0
& 77.3 & 80.7 & 89.9 & 82.3 & 75.6 & 73.1 & 74.6 \\
\midrule

CMT \citep{yan2023cross} (ICCV'23)
& 88.6 & 65.1 & 46.5 & 76.4 & 36.9
& 66.7 & 76.2 & 88.1 & 80.1 & 77.9 & 70.3 & 72.9 \\

MMLF(CMT, Co-DINO)
& 88.7 & 68.3 & 49.1 & 79.0 & 37.2
& 77.3 & 80.9 & 89.3 & 82.0 & 78.1 & 73.0 & 74.8 \\
\bottomrule
\end{tabular}
}
\caption{\small
\textbf{3D detection performance on the standard nuScenes benchmark.}
We evaluate the impact of late fusion on the standard 10-class
validation benchmark.
Although our work focuses on improving detection accuracy in the tail,
our proposed approach also improves performance on the standard
benchmark. Notably, fusing CenterPoint \citep{yin2021center} with
Co-DINO \citep{zong2023detrs} improves mAP and NDS by 6.5 and
10.2 percentage points, respectively.
Fusing CMT \citep{yan2023cross}, a multimodal 3D detector, with
Co-DINO yields improvements of 2.7 mAP and 1.9 NDS, improving
all standard classes, with the largest gains on \texttt{bicycle},
\texttt{motorcycle}, \texttt{truck}, \texttt{trailer}, and \texttt{bus}.
We posit that CMT already effectively uses multimodal training data
for the common classes of the standard nuScenes benchmark, yielding
diminishing returns with late fusion.
CV denotes \texttt{construction vehicle}, MC denotes
\texttt{motorcycle}, Ped.\ denotes \texttt{pedestrian}, and
TC denotes \texttt{traffic cone}.
}
\label{tab:standard}
\end{table*}
}

In Table~\ref{tab:standard}, we evaluate our late-fusion method on the standard 10 classes of the nuScenes benchmark \citep{caesar2020nuscenes}. Although our late-fusion approach primarily improves performance for classes with {\tt medium} and {\tt few} examples per class, we find that the late-fusion of CenterPoint and Co-DINO yields a 6.5 mAP and 10.2 NDS performance increase on the standard nuScenes benchmark, respectively. Notably, we see significant performance improvements for {\tt bicycle} and {\tt motorcycle}, likely because our LiDAR-only detector misclassifies these geometrically similar, but semantically different categories. 
Next, we apply our late-fusion to CMT \citep{yan2023cross}, a state-of-the-art multi-modal 3D detector. Fusing CMT with Co-DINO \citep{zong2023detrs} yields improvements of 2.7 mAP and 1.9 NDS, improving all standard classes, with the largest gains on bicycle, motorcycle, truck, trailer, and bus.

\section{More Visualizations}
\label{sec:more-visualizations}

\begin{figure*}[t]
\centering
\includegraphics[trim=0cm 0 0cm 0cm, clip, width=1\linewidth]{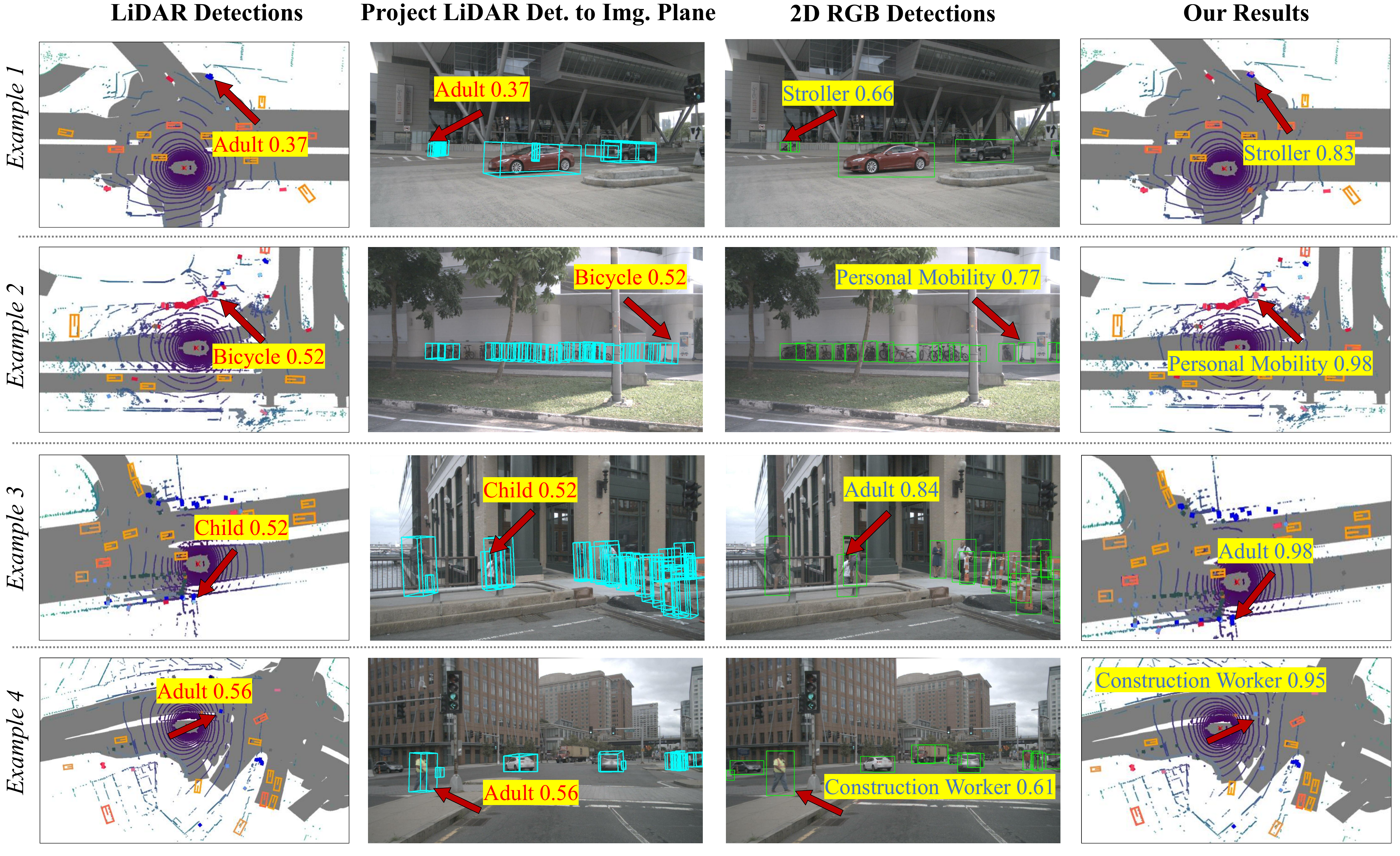}

\vspace{-1mm}
\caption{\small We visualize the output of our late-fusion method, which combines 3D LiDAR detections from CenterPoint and 2D RGB detections from Co-DINO. In all cases, we find that the 2D RGB detector is able to correct classification errors from the 3D LiDAR detector, improving overall performance. Importantly, score calibration and probabilistic ensembling increase the confidence of the final prediction.
}
\label{fig:More visualizations of ours}
\vspace{-1mm}
\end{figure*}

We present additional visualizations of our multi-modal late-fusion approach in Fig.~\ref{fig:More visualizations of ours}.
Our method correctly classifies geometrically similar but semantically different categories like {\tt adult} versus {\tt stroller}, {\tt bicycle} versus {\tt personal mobility}, {\tt child} versus {\tt adult}, and {\tt adult} versus {\tt construction worker}. We provide a video of our results in the supplement.
Our \href{https://github.com/cc50121/lt3d-lf}{project page} contains a video demo.

\begin{figure*}[t]
\centering
\ \hspace{-5.5cm} \bf{MMLF(CenterPoint, Co-DINO)} \hspace{2cm} \bf{CenterPoint} \hspace{10cm} \\    
    \includegraphics[width=0.8\linewidth]{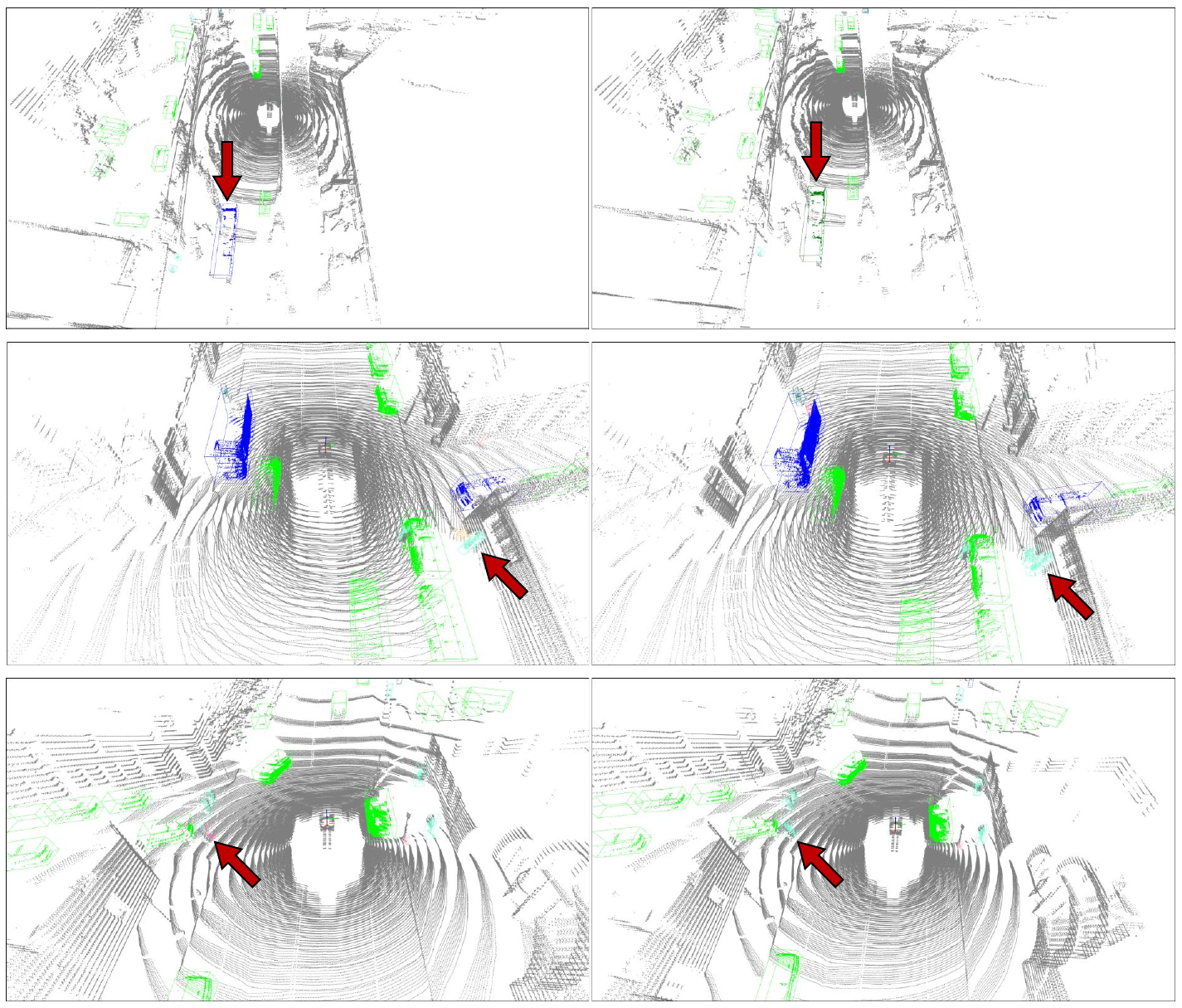}\hfill
    \includegraphics[width=0.18\linewidth]{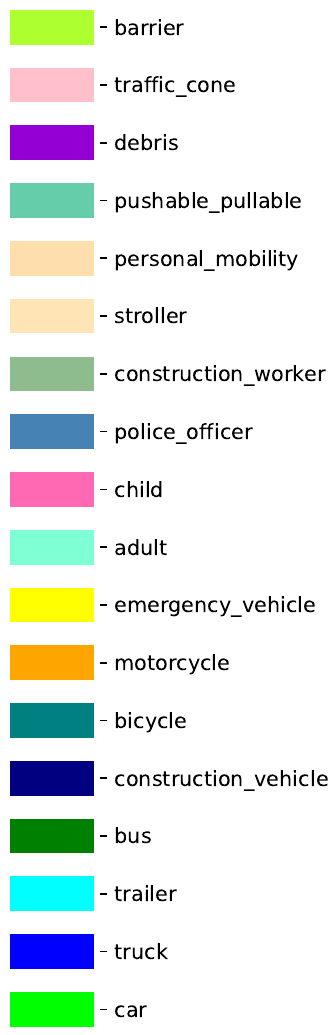}
    \vspace{1mm}
    \caption{\small
    Three examples demonstrate how our multi-modal late-fusion (MMLF) approach improves LT3D by ensembling 2D RGB detections (from Co-DINO) and 3D LiDAR detections (from CenterPoint \citep{yin2021center}). In all examples, MMLF correctly relabels detections which are geometrically similar (w.r.t size and shape) in LiDAR but are visually distinct in RGB, such as {\tt bus}-vs-{\tt truck}, {\tt adult}-vs-{\tt stroller}, and {\tt adult}-vs-{\tt child}.
    This figure complements Fig.~\ref{fig:visual-results} in the main paper.
    }
    \label{fig:fig9_3d}
\end{figure*}

\begin{figure*}[t]
    \centering

\ \hspace{-5.5cm} \bf{MMLF(CenterPoint, Co-DINO)} \hspace{2cm} \bf{TransFusion} \hspace{10cm} \\

    \includegraphics[width=0.80\linewidth]{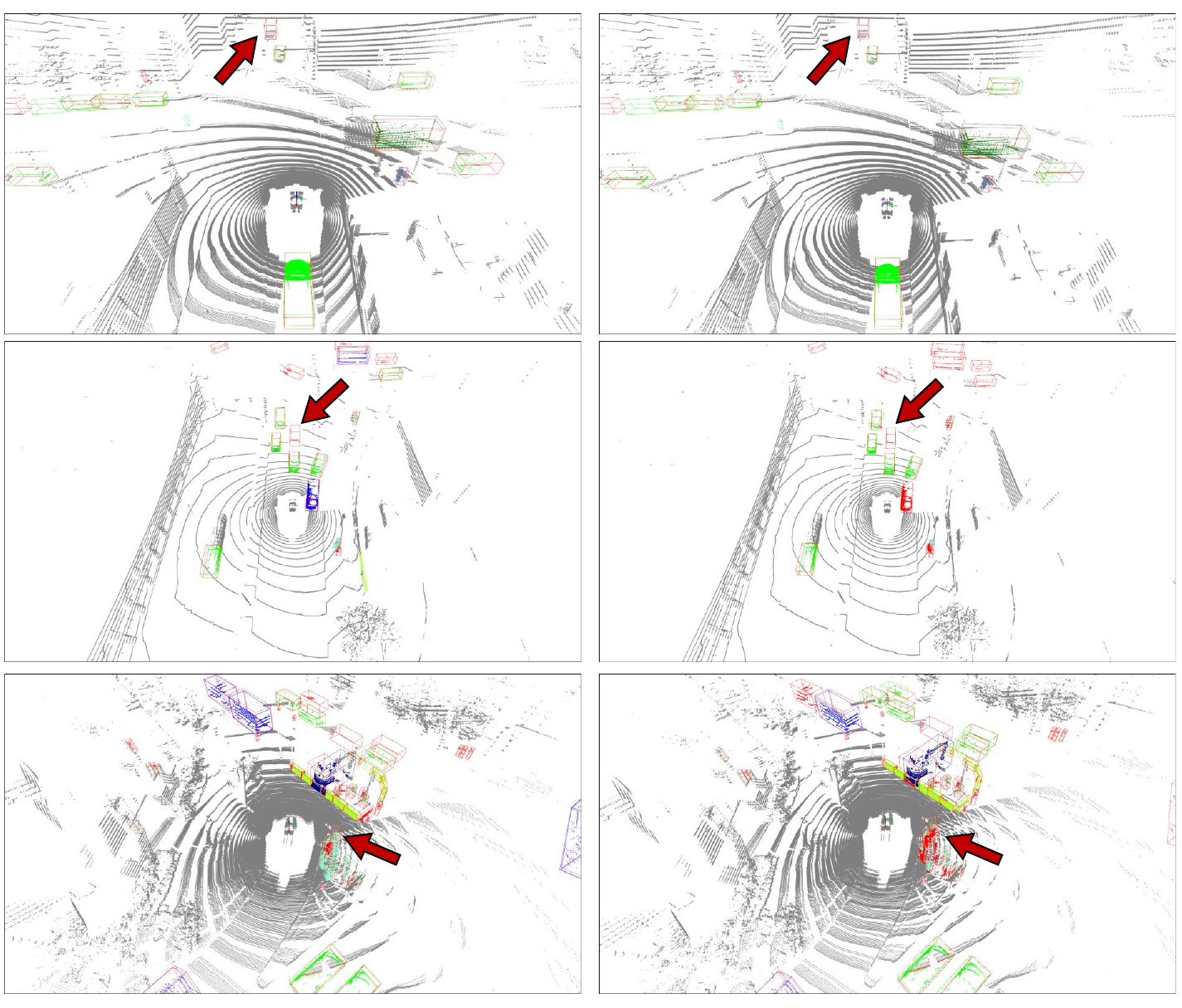}\hfill
    \includegraphics[width=0.18\linewidth]{colors_classes.pdf}
    \vspace{-2mm}
    \caption{\small
    Both our MMLF method (column 1) and TransFusion~\citep{bai2022transfusion} (column 2) share the same failure cases. In the first and second row, the 2D RGB detector Co-DINO detects the heavily occluded cars but 3D LiDAR detector fails to detect them. As a result, our MMLF misses these cars because it throws away unmatched RGB detections which do not have accurate 3D information. In the third row, we see that, although both the LiDAR and RGB detectors fire on the object (whose ground-truth label is {\tt police-officer}), the LiDAR detector classifies it as {\tt adult} and RGB detector classifies it as {\tt construction-worker}. As a result, the final detection is incorrect w.r.t the predicted label.
    TransFusion also misclassifies this object as an {\tt adult}.
    This figure complements Fig.~\ref{fig:failure-cases} in the main paper.
}
    \label{fig:fig10_3d}
\end{figure*}

In Fig.~\ref{fig:fig9_3d} and \ref{fig:fig10_3d},
we visualize the LT3D detection results in a 3D virtual scene. The results are also visualized in the main paper (corresponding to Fig.~\ref{fig:visual-results} and \ref{fig:failure-cases}, respectively) in the BEV scene.

\end{appendices}

\end{document}